%% file: main.tex
\documentclass[10pt,twocolumn,letterpaper]{article}

\usepackage[pagenumbers]{cvpr} % To force page numbers, e.g. for an arXiv version

\input{preamble}

\definecolor{cvprblue}{rgb}{0.21,0.49,0.74}
\usepackage[pagebackref,breaklinks,colorlinks,allcolors=cvprblue]{hyperref}

\def\paperID{16877} % *** Enter the Paper ID here
\def\confName{CVPR}
\def\confYear{2025}

\title{Embodied Scene Understanding for Vision Language Models via MetaVQA}

\author{Weizhen Wang, 
Chenda Duan, 
Zhenghao Peng, 
Yuxin Liu, 
Bolei Zhou\\
University of California, Los Angeles%\\
}

\begin{document}
\maketitle

\input{sec/0_abstract}    
\input{sec/1_intro}

\input{sec/2_related_works}

\input{sec/3_methodology}

\input{sec/4_experiments}

\input{sec/6_conclusions}
{
    \small
    \bibliographystyle{unsrt}
    \bibliography{main}
}
\appendix
\input{sec/X_suppl}

\end{document}

%% file: preamble.tex
\usepackage{amsmath}
\usepackage{amssymb}
\usepackage{booktabs}
\usepackage{multicol,multirow}
\usepackage{setspace}
\usepackage{makecell}
\usepackage{tabularray}
\usepackage{pifont}
\usepackage{wrapfig}
\usepackage{color}
\usepackage{caption, subcaption, multirow, overpic, textpos}
\usepackage{bbm}
\usepackage{titletoc}
\usepackage{graphicx}
\usepackage{grffile}

\usepackage{colortbl}
\usepackage{tcolorbox}

\definecolor{baselinecolor}{gray}{.9}

\definecolor{citecolor}{RGB}{65,105,225}
\newcommand\mypar[1]{\par\vspace{1.0mm}\noindent\textbf{#1}\;\;}

%% file: sec/0_abstract.tex
\begin{abstract}
Vision Language Models (VLMs) demonstrate significant potential as embodied AI agents for various mobility applications. However, a standardized, closed-loop benchmark for evaluating their spatial reasoning and sequential decision-making capabilities is lacking. To address this, we present MetaVQA: a comprehensive benchmark designed to assess and enhance VLMs’ understanding of spatial relationships and scene dynamics through Visual Question Answering (VQA) and closed-loop simulations. MetaVQA leverages Set-of-Mark prompting and top-down view ground-truth annotations from nuScenes and Waymo datasets to automatically generate extensive question-answer pairs based on diverse real-world traffic scenarios, ensuring object-centric and context-rich instructions. Our experiments show that fine-tuning VLMs with the MetaVQA dataset significantly improves their spatial reasoning and embodied scene comprehension in safety-critical simulations, evident not only in improved VQA accuracies but also in emerging safety-aware driving maneuvers. In addition, the learning demonstrates strong transferability from simulation to real-world observation. Code and data will be publicly available at \url{https://metadriverse.github.io/metavqa}.
\end{abstract}

%% file: sec/1_intro.tex
\begin{figure*}[!t]
    \centering
    \includegraphics[width=\linewidth]{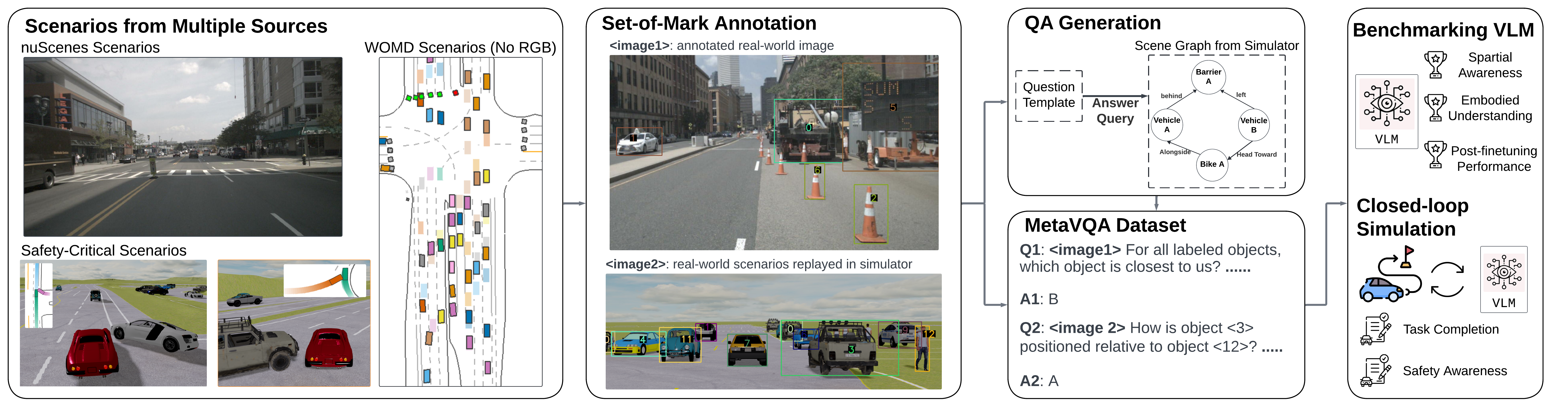}
    \vspace{-2em}
    \caption{\textbf{Constructing MetaVQA benchmark}.
    % VQA generation pipeline. 
    We extract scene graphs from real-world traffic scenarios collected from nuScenes and Waymo datasets(WOMD) and then feed them into question-type-dependent queries to generate ground-truth answers. 
    The real and simulated RGB observations are processed with Set-of-Mark prompting. We evaluate the VLMs on both open-loop VQA tasks and closed-loop navigation tasks in simulation.
    %\bz{remove the color background of each block, as it reduces readability. You can also simplify the text prompts and answers, as it is illustrative purpose, no need to show exact what fed to the model. For example you can remove "from (A) through (D):". Also remove some unnecessary icons, such as the trophy and car-tabels, they are distractions.}
    }
    \label{fig:pipeline}
    % \vspace{-15pt}
        \vspace{-1em}
\end{figure*}
\section{Introduction}
In many real-world robotic applications like autonomous driving and warehouse robots, embodied AI agents have started interacting with physical environments and impacting their surroundings. 
% embodied AI agents have begun interacting with their physical environment, influencing their surroundings.
%safety is at the top of their priorities to prevent damage to the environment (for example, pedestrians) and themselves. 
% To ensure the safety of agent interactions with the environment,
These agents should be sufficiently aware of their surroundings to interact with their environments safely. In this paper, we define this ability as \textit{embodied scene understanding}, which we believe contains two intertwined facets: \textit{spatial awareness} and \textit{embodied understanding}.
Spatial awareness refers to the ability to internalize spatial relationships among observed objects when perceiving the 3D world through the 2D image captured by a monocular camera.
Embodied understanding is the ability to relate observed objects egocentrically, foresee the implication of action, and choose the optimal action to achieve the instructed goal safely.

Recent advances demonstrate the potential of using Vision Language Models~(VLMs) as embodied agents in applications from robot arms control~\cite{rt22023arxiv,kim24openvla} to autonomous driving~\cite{DriveVLM}. These tasks share the common components of following instructions, understanding the environment, and taking the optimal action to achieve specified goals. Benefiting from large-scale pre-training, VLMs retain embodied scene understanding to a certain extent. However, their spatial awareness is limited as most VLMs are pre-trained on offline text and images. Meanwhile, their embodied understanding is also constrained because instruction-following interaction with the environment occupies a very small portion of their training data.

In the task of autonomous driving, many prior works~\cite{DriveVLM, sima2023drivelm, zhou2024embodied, xu2023drivegpt4, deruyttere2019talk2car, autotrust, DriveMLLM} address this training distribution mismatch by fine-tuning VLMs on Visual-Question-Answering (VQA) tasks tailored for driving scenarios with reported improvements on their benchmarks. However, these benchmarks are not commensurable or suitable for zero-shot evaluation on off-the-shelf general-purpose VLMs. 
This is because 
% Each work establishes its tailored metrics. In addition, 
they follow different textual and visual expressions to describe the scene and refer to objects.
% when conveying the semantically same information. 
For example, DriveLM~\cite{sima2023drivelm} refers to objects by tuples composed of the object identifier, the ID of the corresponding camera, and the pixel positions of the 2D bounding box's vertices in the camera's coordinate. In contrast, in ELM~\cite{zhou2024embodied}, objects are grounded by a triple composed of the character ``c'' and the pixel coordinates of the center of the 2D bounding box. Not only do these works disagree in description conventions, but their chosen conventions drastically differ from how humans would intuitively refer to an object. A person would point to the object or ground the object by its features (for example, color or shape). This mismatch can weaken the diagnosing power of the VQA datasets: an unsatisfactory performance of a VLM may be caused by its inability to interpret the question expressions rather than its lack of scene understanding capability. %since they are very ``niched" and distant from the human-generated content.

In addition, existing works mainly evaluate embodied scene understanding of VLMs on the VQA task in the open-loop setting. Nevertheless, embodied understanding capability should be examined more thoroughly in the closed-loop setting, where the VLMs must act in an interactive environment for safe sequential decision-making. DriveVLM~\cite{DriveVLM} attempts the closed-loop evaluation by prompting VLMs to describe the surroundings of a vehicle driven by a human driver. However, involving manual effort in closed-loop evaluations is not scalable, and this setting can not evaluate the embodied understanding of the VLMs since their actions have no consequences for the proceeding environment.
Vista~\cite{gao2024vista} uses a trained world model to simulate real-world driving scenarios. However, the observations from the trained world model suffer deteriorated distortion from reality in long-term trajectories. Finally, safety-critical scenarios are absent in existing works since they utilize only real-world data (and therefore, the collection of dangerous situations is costly and unlikely). However, we need a stress test for an embodied agent to verify the safe autonomy of the VLMs. Thus, it is not sufficient to only evaluate the VLMs as embodied agents against normal scenarios. 

To address these issues, we present MetaVQA, a benchmark designed for performing the zero-shot evaluation of \textbf{\textit{general-purpose VLMs}} and further improving their embodied scene understanding through finetuning, as shown in ~\cref{fig:pipeline}. For generalizable learning and evaluation, MetaVQA contains a large-scale and high-quality VQA corpus in common wordings, and its observations--including both real and simulated images from diverse scenarios--are annotated following the Set-of-Mark prompting\cite{yang2023setofmark} for clear reference and visual grounding. 
Furthermore, MetaVQA utilizes the simulation environment MetaDrive~\cite{li2022metadrive} for scalable closed-loop evaluation of the VLMs as embodied agents in real-world traffic scenarios collected from nuScenes~\cite{caesar2019nuscenes} and Waymo Open Motion Dataset~\cite{sun2020scalability}. Models go through stress tests by driving in safety-critical scenarios for more thorough safety evaluations. We have established the baseline performance of many representative VLMs ~\cite{liu2023improved, li2024llava, Qwen2VL, OpenAI_ChatGPT4_2024, dubey2024llama3herdmodels, chen2023internvl}, and we observe significant improvement in embodied scene understanding after fine-tuning, shown by the VQA accuracy and the performance of the closed-loop driving task. Additional experiments also verify that \textit{learning on simulated data can substantially improve the embodied understanding in the held-out real-world data}. More specifically, the VLM model fine-tuned on our synthetic VQA data becomes better in nuScene real-world VQA task.
We summarize our contributions as follows:

\begin{enumerate} 
    \item We propose the MetaVQA dataset to evaluate and enhance the embodied scene understanding capabilities of the VLMs in a plug-and-go way and bring significant improvement when fine-tuned on our data.

    \item  We show that the embodied scene understanding of the VLMs is generalizable and transferable from simulated to real-world data. The VLM trained only on simulated data shows remarkable zero-shot performance on real-world visual question answering. 

     \item We conduct the closed-loop simulation evaluation of the VLMs as embodied agents in safety-critical driving scenarios. The fine-tuned VLMs exhibit reasonable sequential decision-making skills when charged with the unseen task of driving, validating the generalizability of learned embodied knowledge.
\end{enumerate}
% \noindent 

%% file: sec/2_related_works.tex
\section{Related Work}

\mypar{Driving scene understanding datasets.}
Many datasets have been proposed for driving scene understanding~\cite{wu2023language,xu2020explainable,wu2023referring,li2024opensourceddataecosystemautonomous}.
For explaining driving behaviors, some works \cite{kim2019CVPR,keysan2023text,malla2023drama,echterhoff2023driving,ding2023hilmd,chen2023drivingwithllms,jin2023adapt, kim2018textual, deruyttere2019talk2car, sachdeva2024rank2tell, qian2023nuscenes, sima2023drivelm, autotrust, DriveMLLM} provide annotations for scene descriptions, traffic elements, and high-level instructions.
For example, BDD-X \cite{kim2018textual} and Talk2Car \cite{deruyttere2019talk2car} supply succinct descriptions of driving scenarios and directions; Rank2Tell \cite{sachdeva2024rank2tell} and nuScenes-QA \cite{qian2023nuscenes} focus on ranking object importance and annotating road element attributes, respectively; DriveLM-nuScenes/CARLA \cite{sima2023drivelm} utilizes a graph-based visual-question-answering for driving descriptions. Concurrent with this work, DriveMLLM ~\cite{DriveMLLM} inspects spatial understanding in absolute (detailed to meters) and relative descriptions (for example, left or right).
%
 %concurrent
Additionally, AutoTrust~\cite{autotrust}, another concurrent work, focuses on the trustworthiness and safety-awareness of VLMs in driving. 
However, these datasets often rely on limited traffic data (primarily the nuScenes ~\cite{caesar2019nuscenes} dataset) and lack diversity in data/question types, limiting their potential for broader, generalizable applications. We aim to address these limitations by enhancing the data scale and providing diverse question types for improved scene understanding.

\mypar{Vision language models as driving agents.}
Recent attempts leverage language models to enhance traditional autonomous driving tasks~\cite{elhafsi2023semantic,seff2023motionlm,sha2023languagempc}.
DriveGPT4~\cite{xu2023drivegpt4} delivers an interpretable system, while Lingo-1~\cite{wayve2023lingo1} integrates vision, language, and action for model training and interpretations. GPT-Driver~\cite{mao2023gptdriver} leverages large language models for trajectory prediction, and ELM~\cite{zhou2024embodied} and DriveVLM~\cite{tian2024drivevlm} expand capabilities in long-horizon space and handling corner cases. However, these models are often evaluated offline, lacking a closed-loop evaluation framework that captures the interactive nature of driving. Thus, closed-loop evaluation is crucial for assessing the robustness and reliability of vision-language models in real-world driving scenarios.

\mypar{Closed-loop evaluation of end-to-end driving agents.}
Driving simulators are crucial for autonomous driving, and they facilitate closed-loop testing before real-world trials. Notable examples include LimSim++~\cite{fu2024limsim++}, MetaDrive~\cite{li2022metadrive}, CARLA~\cite{dosovitskiy2017carla}, and nuPlan~\cite{caesar2021nuplan} which use rule-based engines in evaluating driving behavior. However, their evaluation metrics typically focus on numerical values such as success rate and route completion rate, which lack interpretability. To address this limitation, we aim to incorporate scene understanding and the reasonableness of behaviors into the evaluation process by using more diverse and meaningful metrics.

%% file: sec/3_methodology.tex
\section{Constructing MetaVQA Dataset}
\subsection{Our Design Principles}
We aim to develop a VQA generation pipeline for benchmarking general-purpose VLMs on embodied scene understanding and for fine-tuning them to serve as an embodied agent in the driving task. We consider the following two key questions when designing our dataset:
% \begin{itemize}
\mypar{How to effectively communicate with general-purpose VLMs?}
An analogy can be drawn for the VLMs performing the VQA task as students taking standardized tests. To fairly evaluate all students' learning outcomes, the instructor should create a problem set where the question and answering instructions are clear and intuitive. We notice that when evaluating embodied scene understanding, existing works have a variety of heterogeneous prompting conventions and expected answer forms. For example, DriveLM~\cite{sima2023drivelm}, ELM~\cite{zhou2024embodied}, and DriveVLM~\cite{DriveVLM} refer to objects by pixel positions of 2D bounding box's vertices in the camera's coordinate. These works expect the models to associate pixel coordinates with regions on the images. However, such association convention is rare in the pre-training data corpus collected from the Web. In addition, ELM prompts models to convey spatial and dynamics (ego speed, heading, \textit{etc.}) information in continuous values (world coordinates, top-down yaw angle, meters per second, \textit{etc.}), contrasting to DriveLM, which discretizes this information. Nevertheless, both conventions are unfamiliar to general-purpose VLMs pre-trained on human-created Internet data.
 % as it is dominated by human-created content in which such convention is counterintuitive
Therefore, if general-purpose VLMs are zero-shot evaluated on these tasks and protocols, it will be difficult to attribute the cause of deficient performance to a lack of embodied scene understanding or an inability to follow the question-answering convention. % and the direct diagnosing power of these works is questionable. 
In contrast, CLEVR~\cite{johnson2017clevr} uses template-generated English referrals. However, natural language referral can be ambiguous when the scene becomes complicated in layout and appearance, making the questions disputable and hindering fair evaluation.

\mypar{How to thoroughly evaluate embodied scene understanding?}
As mentioned in the introduction section, we interpret the embodied scene understanding capabilities from two aspects: understanding spatial relationships and understanding an action's consequence. 

%concurrent

Existing works~\cite{DriveVLM, sima2023drivelm, tian2024drivevlm, zhou2024embodied, DriveMLLM, autotrust} evaluate this capability on the VQA tasks by creating a sufficiently encompassing question suite. However, the metrics for evaluations are incommensurable. ELM introduced ``Pr@k" metrics to evaluate the performance on localization tasks, while DriveLM uses classification error as the metrics since they discretize space. An extreme case can be found in DriveMLLM~\cite{DriveMLLM}, in which distinct accuracy metrics are defined for each type of question. In addition, in the driving field, closed-loop egocentric evaluation through interactive simulation is rarely explored, leaving the claim of learned embodied scene understanding untested in situations of actual embodiments.

To address the two questions above, we start with the Set-of-Mark (SoM) prompting technique~\cite{yang2023setofmark}. SoM elevates the visual grounding capabilities of VLMs and provides an intuitive and unambiguous referral scheme with labeling. The suitability of using SoM prompting is further discussed in \cref{sec:human} and \cref{sec:groundable}. We formulate each question in the multiple-choice setting with only one unique correct option to make fair evaluations. To support zero-shot evaluations of general-purpose VLMs, spatial and dynamics information is discretized into common phrases (for example, ``front") with fine-grained numerical descriptions added as additional explanations. ~\cref{sec:human} showcases the intuitiveness and descriptiveness of this setup through human study, and the zero-shot performances in the benchmark from ~\cref{sec:benchmark} empirically validate the suitability. To thoroughly examine the embodied scene understanding of VLMs, we construct 30 question types covering all aspects of spatial reasoning and embodied understanding of VLMs with diverse scenarios using both real-world and simulated observations, and we further evaluate VLMs with closed-loop  simulation.

\subsection{VQA Generation Pipeline}
To generate diverse questions from large-scale driving scenarios in scale, we employ a search-based method to programmatically extract answers for template-generated questions from scene graphs and quality-check the dataset with human evaluators as discussed in \cref{sec:human}.
\cref{fig:pipeline} illustrates the overall VQA generation pipeline of the MetaVQA. There are three essential stages: scenario aggregation, Set-of-Mark annotations, and QA generation. We describe each stage as follows.

\subsubsection{Scenario Aggregation from Multiple Sources.}
MetaVQA utilizes real-world traffic from Waymo Open Motion Dataset (WOMD)~\cite{ettinger2021large} and nuScenes dataset~\cite{caesar2019nuscenes} to evaluate the safety and robustness of the VLMs against diverse situations. The WOMD dataset is much larger than the nuScenes dataset, which contains only around 800 20-second scenarios. However, the WOMD dataset lacks RGB observations, making its diverse scenarios inaccessible directly. Amending the missing observations, we use a lightweight simulator, MetaDrive\cite{li2022metadrive}, to reconstruct WOMD traffic scenarios for scalable and efficient simulation.  We also create the digital twins of the nuScene dataset in simulation to augment the appearance diversity of collected scenarios. 
\begin{figure}[!t]
    \centering
\includegraphics[width=\linewidth]{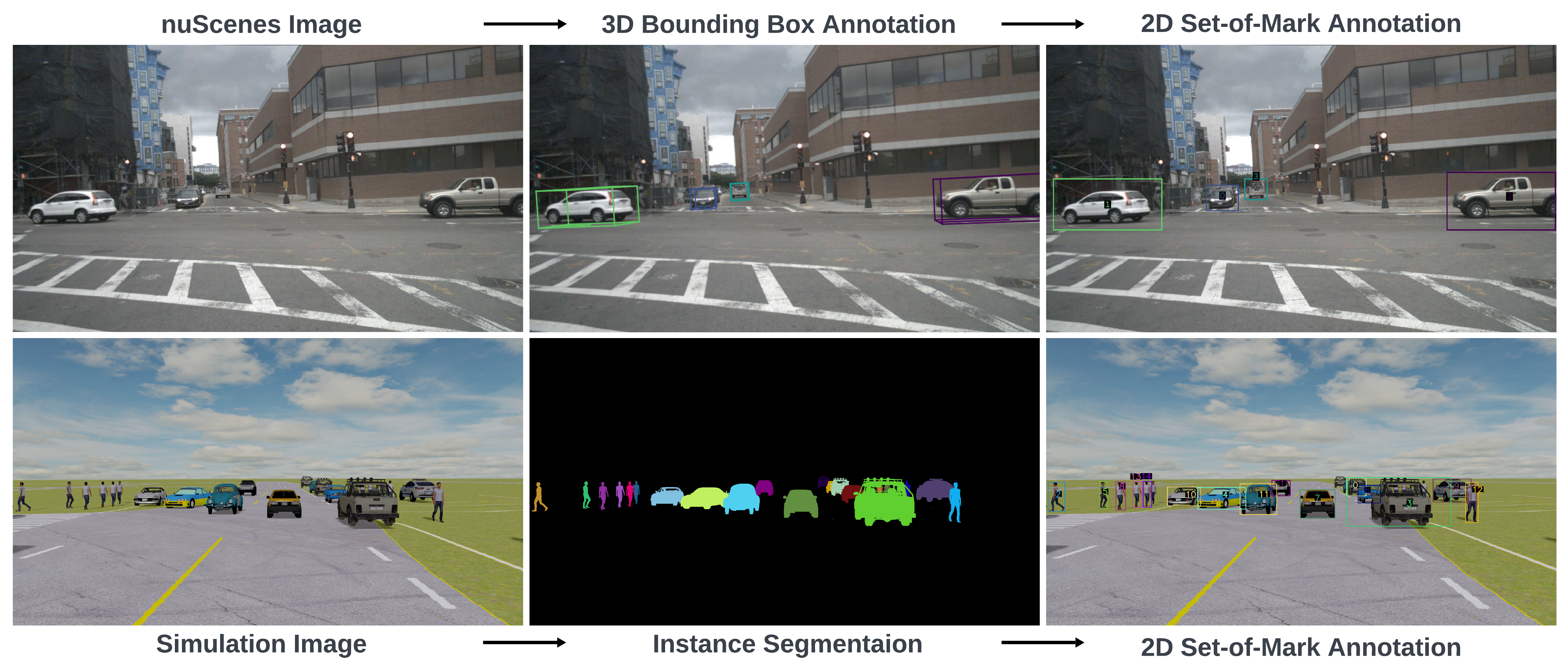}
\vspace{-1em}
    \caption{\textbf{Set-of-Mark annotation process}.
    For real-world images from the nuScenes (upper row) dataset, we cast the corresponding 3D bounding boxes into 2D space.
    For simulated images (lower row) rendered in MetaDrive, we extract 2D bounding boxes from the simulator's instance segmentation. %\bz{put some white margins between images, also good to use vectorized format for a figure (pdf/eps, instead of jpg/png. also it will be great to make the arrow style the same as the arrows in other figures.}
    }
    \label{fig:som}
    \vspace{-1.5em}
\end{figure}

\mypar{Scene collection of nuScenes dataset.} The well-annotated nuScenes dataset provides a devkit to visualize driving scenarios, and we use it to extract 3D scene graphs from each annotated keyframe. Each node in the graph contains spatial and pose information of objects, and labeled edges--representing spatial relationships--are connected algorithmically. Objects are also filtered by relevancy and visibility before being loaded in as object nodes in the scene graph, and implementation details can be found in the supplementary materials. 

\mypar{Scene reconstruction with simulator.} We also harness ScenarioNet~\cite{li2023scenarionet} to aggregate the WOMD and the nuScenes datasets into a unified scene record. The MetaDrive simulator loads those scenarios and renders top-down layouts into egocentric RGB images. Following the nuScenes convention, keyframes are selected at 2Hz frequency, and we follow the sensor setup in the nuScenes dataset for cross-domain perspective consistency of observations. We additionally use nuScenes traffic to create digital twins of the original image, making the evaluation of embodied scene understanding less sensitive to variation in lighting and object appearances.

\subsubsection{Set-of-Mark Prompting}
MetaVQA aims to provide a generalizable evaluation of embodied scene understanding for off-the-shelf VLMs, and Set-of-Mark~\cite{yang2023setofmark} (SoM) prompting provides a clear avenue for humans to issue instructions and VLMs to visually ground objects of interest. We devoted this work on the reasoning ability of VLMs as embodied agents. Therefore, we assume the perception task as a mostly solved problem, and directly highlighting relevant traffic objects is not an overdue hint. This presumption is supported since various detection methods have been established and widely applied~\cite{yolov3, kirillov2023segany, ren15fasterrcnn}. As illustrated in \cref{fig:som}, objects of interest are enclosed by 2D bounding boxes. We extract maximal 2D bounding based on 3D bounding box's projections for the nuScenes images, and we use a shader-based instance segmentation camera to create annotated simulated images. We utilize the labeling algorithm suggested in the original Set-of-Mark paper to ensure maximal labeling consistency and visibility. Further discussions on the impact of marking style can be found in the supplementary materials.

\subsubsection{Question-Answer Generation}
\begin{figure}[!t]
    \centering
    \includegraphics[width=\linewidth]{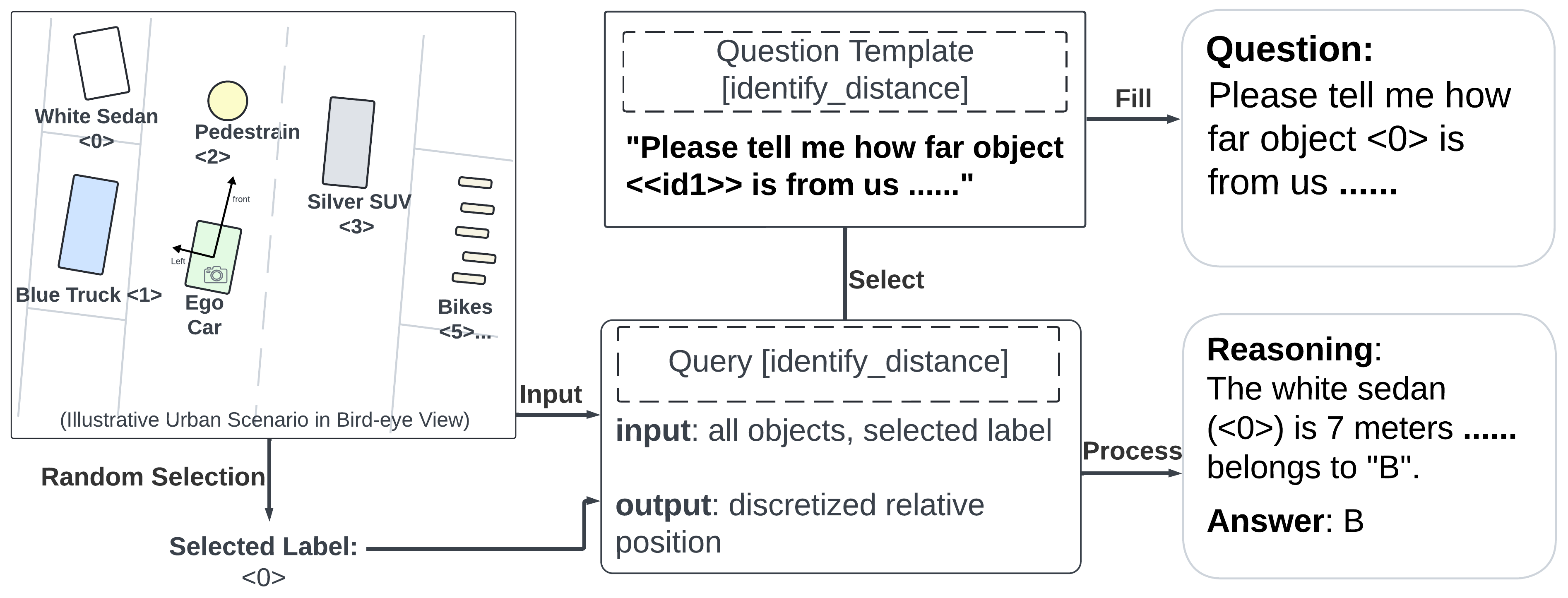}
    \vspace{-1em}
    \caption{\textbf{Question-answer generation pipeline}. An illustrative example for generating the \texttt{identify\_distance} question. Note that an additional ``reasoning" field is generated along with the answer to improve VLM training. This field is not used in evaluation.
    %\bz{Simply the question-answer pairs here and make it more concise and shorter. It is for illustrative purpose.}
    }
    \label{fig:qa_gen}
    \vspace{-1.5em}
\end{figure}

Each question is a multiple-choice question. Since we want a generalizable evaluation without domain-specific fine-tuning of VLMs, this simple setup is straightforward and also complete to cover answer space as we abstract numerical spatial information into discrete options and thus can be fully covered as multiple-choice options. As illustrated in \cref{fig:qa_gen}, during the question-answer generation process, a random object node in the scene graph is selected to be examined, and the ground truth answer--along with other multiple-choice candidates--is generated based on the scene graph using question-type-bounded queries. To prevent VLM collapses during fine-tuning, we additionally incorporate an ``explanation" field in the generated VQAs to explain the selection of answers with dense captions. During training, this field will be part of the ground truth answers VLMs are tasked to generate to deepen their scene understanding. The implementation details can be found in the supplementary materials. 

%% file: sec/4_experiments.tex
\begin{figure*}[!t]
\centering
\includegraphics[width=\linewidth]{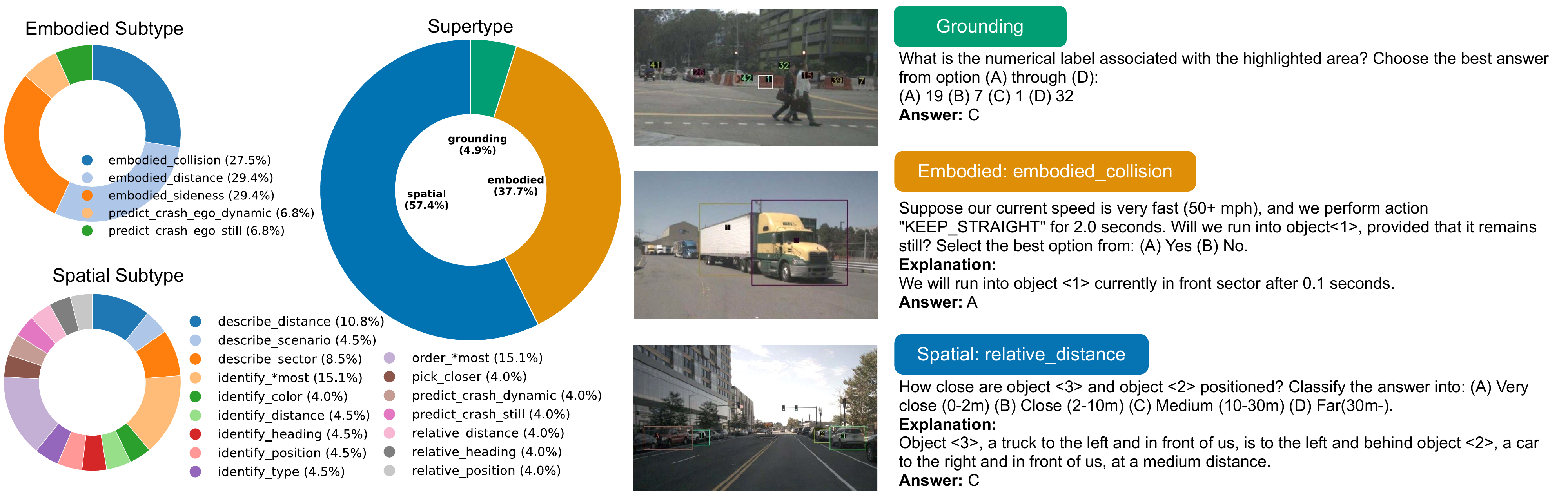}
\vspace{-1em}
\caption{
\textbf{Left:} Distribution of the question types. \textbf{Right}: Example for each question supertype.
%\pzh{Message figure 4 and 6.}
}
\label{fig:vqa_sample}
\vspace{-4pt}
\end{figure*}

\section{MetaVQA Dataset}
We present the MetaVQA Dataset. In addition to the dataset composition, we verify the dataset quality and its helpfulness for evaluating embodied scene understanding. We validate that (1) MetaVQA Dataset's Set-of-Mark annotated questions are suitable for zero-shot evaluation of general-purpose VLMs. (2) The embodied scene understanding capability that emerges from the finetuning is generalizable and transferable from the simulated world to the real world, supporting the utility of simulator-conditioned synthetic data. (3) The learning scales with the training set size using MetaVQA Dataset, warranting the need for a large-scale question-answer corpus.

\subsection{Dataset Composition}
\label{sec:composition}
% \pzh{The term ``digital twin'' suddenly appear without any descrition.}
% As shown in \cref{tab:comparisons}, the
%\bz{instead of using ``"" to denote question type, probably you can use one of the style like \textit{} or \textbf{} or \texttt{}. Please update it in all the paragraphs and figures}. 
MetaVQA Dataset consists of a large corpus of multiple-choice questions, which contains 4,305,450 questions using 442,102 annotated frames extracted from 400 nuScenes scenarios and 6,900 Waymo scenarios covering 59,682 seconds (16.5 hours) of driving log. The questions can be categorized into three supercategories: \texttt{spatial} questions, \texttt{embodied} questions, and \texttt{grounding} questions. The former two supercategories cover the two facets of embodied scene understanding: spatial awareness and embodied understanding,  and the latter one diagnoses VLMs' capabilities to associate marked objects in the observation with textual referral. Detailed formulation for each subcategory is discussed in the supplementary materials, and we lay out the compositions and exemplar VQA pairs in \cref{fig:vqa_sample}.
Noticeably, we design a special training-only \texttt{describe\_scenario} question that asks models to describe all labeled objects in the observation to accentuate embodied scene understanding. For expedited experimentation, we curate a representative training set of 150,000 questions, with 50,000 coming from Waymo scenarios, 50,000 coming from nuScenes scenarios, and 50,000 coming from the simulated nuScenes scenarios in the simulator. In addition, we curate a withheld test set of 9,725 questions with 2,524 annotated frames from 212 traffic scenarios. Approximately half of the questions use observations reconstructed from simulation, and the other half use real-world images.

\subsection{Zero-shot Answerability with Set-of-Mark Prompting}
\label{sec:human}
\mypar{Human evaluation.} We create a questionnaire comprising 35 sampled questions and distribute it to six novice human participants, who are instructed to answer the questions without assistance.
Since VLMs are trained on human-generated texts, good human amateur performance suggests the intuitiveness of the wordings in the question bodies and the clarity in permitted answers. Upon reporting, novice participants achieve an average of 88\% accuracy despite acknowledging question difficulty. This finding showcases the answerability and directness of the MetaVQA Dataset. Details are included in the supplementary materials.

\mypar{Set-of-Mark prompting.} Previous work verifies that using Set-of-Mark prompting on real pictures improves the visual grounding abilities of VLMs~\cite{yang2023setofmark}. To further validate that VLMs can associate referred labels in text with marked regions in both simulated and real images, we introduce \texttt{grounding} questions. Objects are labeled randomly for each such question, and one of the labels is enclosed with a white box. As illustrated in \cref{fig:vqa_sample}, VLMs are prompted to identify the singular correct label located inside the box. We benchmark the zero-shot grounding performance on multiple VLMs on the withheld test set from \cref{sec:composition}, as shown in \cref{tab:grounding}, VLMs achieve, on average, 69.6\% zero-shot accuracies on the 467 grounding questions, with the best-performing VLM achieving 87.4\% accuracies. Noticeably, LLaVA-NeXT (llava-1.6-vicuna-7b)~\cite{liu2023improved}  underperforms significantly, and this abnormality is caused by the VLM failing to generate tokens in the expected convention, which will be discussed in ~\cref{sec:benchmark}. 
This finding suggests that most VLMs can accurately associate labels referred in text with objects. Therefore, using labels for object referral in the question body is an unambiguous and effective wording convention, and we can be confident in associating bad testing performance on the MetaVQA Dataset with a deficiency in embodied scene understanding.

\begin{table}[!t]
\centering
\resizebox{1.0\linewidth}{!}{
\begin{scriptsize}
  \begin{tabular}{lccc}
    \toprule
    Model                    & Overall & Sim-Grounding & Real-Grounding \\
    \midrule
    \textbf{LLaVA}-NeXT  ~\cite{liu2023improved}      & 0.248   & 0.271         & 0.229          \\ %\textbf{llava}-1.6-vicuna-7b
    \textbf{LLaVA}-OneVision ~\cite{li2024llava}   & 0.728   & 0.615         & 0.827          \\ %\textbf{llava}-onevision-7b-ov 
    \textbf{GPT}-4o ~\cite{OpenAI_ChatGPT4_2024}                   & 0.831   & 0.766         & \textbf{0.888} \\
    \textbf{Qwen2} ~\cite{Qwen2VL}     & \textbf{0.874}   & \textbf{0.890} & 0.859          \\ %\textbf{qwen2}-vl-7b-instruct
    \textbf{Llama3.2} ~\cite{dubey2024llama3herdmodels}     & 0.790   & 0.716         & 0.855          \\ %\textbf{llama-3.2}-11B-Vision-Instruct
    \textbf{InternVL2}-8B ~\cite{chen2024internvl}             & 0.702   & 0.610         & 0.783          \\
    \midrule
    Average                  & 0.696   & 0.644         & 0.740          \\
    \bottomrule
  \end{tabular}
\end{scriptsize}
}
\vspace{-0.5em}
    \caption{\textbf{Zero-shot grounding accuracy}. Most VLMs successfully associate label referrals in the question body with marked regions, showing that Set-of-Mark prompting effectively conveys instruction.}
    \vspace{-1.5em}
    %\bz{Bold the first words for all the methods, do the same for other tables.}}
\label{tab:grounding}
\end{table}

\subsection{Transfer learning with simulated observations}
\label{sec:groundable}
A concern can be raised against the use of simulated images. The MetaDrive simulator~\cite{li2022metadrive}, while capable of importing real-world traffic scenarios, falls short in visual fidelity compared to real-world photos. We conduct four trials on the withheld test set from \cref{sec:composition} using InternVL2-8B ~\cite{chen2024internvl} as the learning VLM to address this concern. In the first trial, InternVL2-8B is evaluated without any fine-tuning. In the next three trials, InternVL2-8B is fine-tuned on (1) 50,000 questions with only simulated observations, (2) 50,000 questions with only real images, and (3) 150,000 questions with both simulated and real observations. The third training set comprises the second set and a superset of the first training set.

\begin{table}
  \begin{minipage}[t]{0.47\linewidth}
  \vspace{0pt} % Ensure top alignment
    \centering
    \resizebox{1\linewidth}{!}{
    \begin{scriptsize}
      \begin{tabular}{lccc} %@{}@{}
        \toprule
        Training Set  & Overall      & Sim & Real \\
        \midrule
        Zeroshot   & 0.592        & 0.552      & 0.632 \\
        Sim only   & 0.807        & 0.795      & 0.819 \\
        Real only  & 0.825        & 0.792      & 0.858 \\
        Sim+Real   & \textbf{0.869}        & \textbf{0.853}      & \textbf{0.884} \\
        \bottomrule
      \end{tabular}
      \end{scriptsize}
    }
    %\vspace{-0.5em}
    \caption{\textbf{Sim-to-Real transferability}. InternVL2-8B achieves the best test accuracy when trained using both simulated and real images, indicating the transferability of embodied knowledge learned from the MetaVQA Dataset.} 
    \label{tab:sim2real}
  \end{minipage}
  \hfill
  \begin{minipage}[t]{0.47\linewidth}
  \vspace{0pt} % Ensure top alignment
    \centering
    \resizebox{1\linewidth}{!}{
    \begin{scriptsize}
    \begin{tabular}{lccc}
    \toprule
    \makecell{Training Size} & Overall & Sim        & Real \\
    \midrule
    9,375         & 0.794  & 0.764       & 0.824        \\
    37,500        & 0.845  & 0.825       & 0.865        \\
    150,000       & \textbf{0.869} & \textbf{0.853} & \textbf{0.884} \\
    \bottomrule
  \end{tabular}
  \end{scriptsize}
  }
  \vspace{5pt}
  \caption{\textbf{Data scalability}. We observe a positive correlation between InternVL2-8B's test performance and training data size. This finding warrants the scalability of learning using the MetaVQA Dataset.}
  \label{tab:scaling}
  \end{minipage}
  \vspace{-1.5em}
\end{table}

Referring to \cref{tab:sim2real}, InternVL2-8B achieves the best test accuracy when learning from both simulated and real observations. Noticeably, training on simulated observations alone significantly increases real-world image test accuracy, and training on real observations benefits simulated image test performance reciprocally. Therefore, the learned embodied scene understanding is transferrable between simulation and the real world, and this finding warrants the benefits of adding simulated observation into the training data.

\subsection{Data Scalability of Learning}
We conduct this experiment to demonstrate the positive correlation between learned embodied understanding and the size of the training set to support the need for a large VQA corpus. Using the training set specified in \cref{sec:composition}, we gradually down-sample the questions with a factor of 4, generating training sets of 150,000 questions, 37,500 questions, and 9,375 questions correspondingly. We fine-tune InternVL2-8B on each of the three sets and report the test accuracy on the withheld test set specified in \cref{sec:composition}. As illustrated in \cref{tab:scaling}, InternVL2-8B's test performance scales with the training set size, warranting the encompassing size of the MetaVQA dataset.
%\begin{table}
%\centering
%\begin{scriptsize}
%  \begin{tabular}{lccc}
%    \toprule
%    Training Set Size & Overall & Sim        & Real \\
%    \midrule
%    9,375         & 0.794  & 0.764       & 0.824        \\
%    37,500        & 0.845  & 0.825       & 0.865        \\
%    150,000       & \textbf{0.869} & \textbf{0.853} & \textbf{0.884} \\
%    \bottomrule
%  \end{tabular}
%\end{scriptsize}
%\vspace{-0.5em}
%  \caption{\textbf{Data Scalability Analysis.} Evaluation of InternVL2 with increasing training set size. We observe that %the test performance scales with the training set size.}\label{tab:scaling}
 %\vspace{-5pt}
%\end{table}

%\subsection{Contribution of Scenario Diversity}
%We fine-tune models on three equally sized sets with simulated observations only. The first set has %100\% of data generated on Waymo scenarios; the second set has 100\% of data generated on nuScenes %scenarios; and the third set has 50\% of data coming from the first set and the other 50\% coming from %the second set. We examine the models' performance on the withheld test set. 

%\noindent \textbf{Results.}
%Referring to \href{https://docs.google.com/spreadsheets/d/1ZEJIFuGpeVLGpf5EFd9zQkSKQaMoA6o54NDfp1eNc-M/edit?gid=120191788#gid=120191788}{table}, models fine-tuned on combined training data(set 3) achieves the best overall performance. This finding warrants the benefits of simulators introducing previously unusable traffic data into the training corpus for elevated performance. 

%\subsection{Cross-Dataset Improvements}
%We co-fine-tune models with other VQA datasets. The evaluation performances are reported.

%\noindent \textbf{Results.}
%Referring to %\href{https://docs.google.com/spreadsheets/d/1ZEJIFuGpeVLGpf5EFd9zQkSKQaMoA6o%54NDfp1eNc-M/edit?gid=689267828#gid=689267828}{table} \td{report findings.}

\begin{figure*}[!t]
    \centering
    \includegraphics[width=\linewidth]{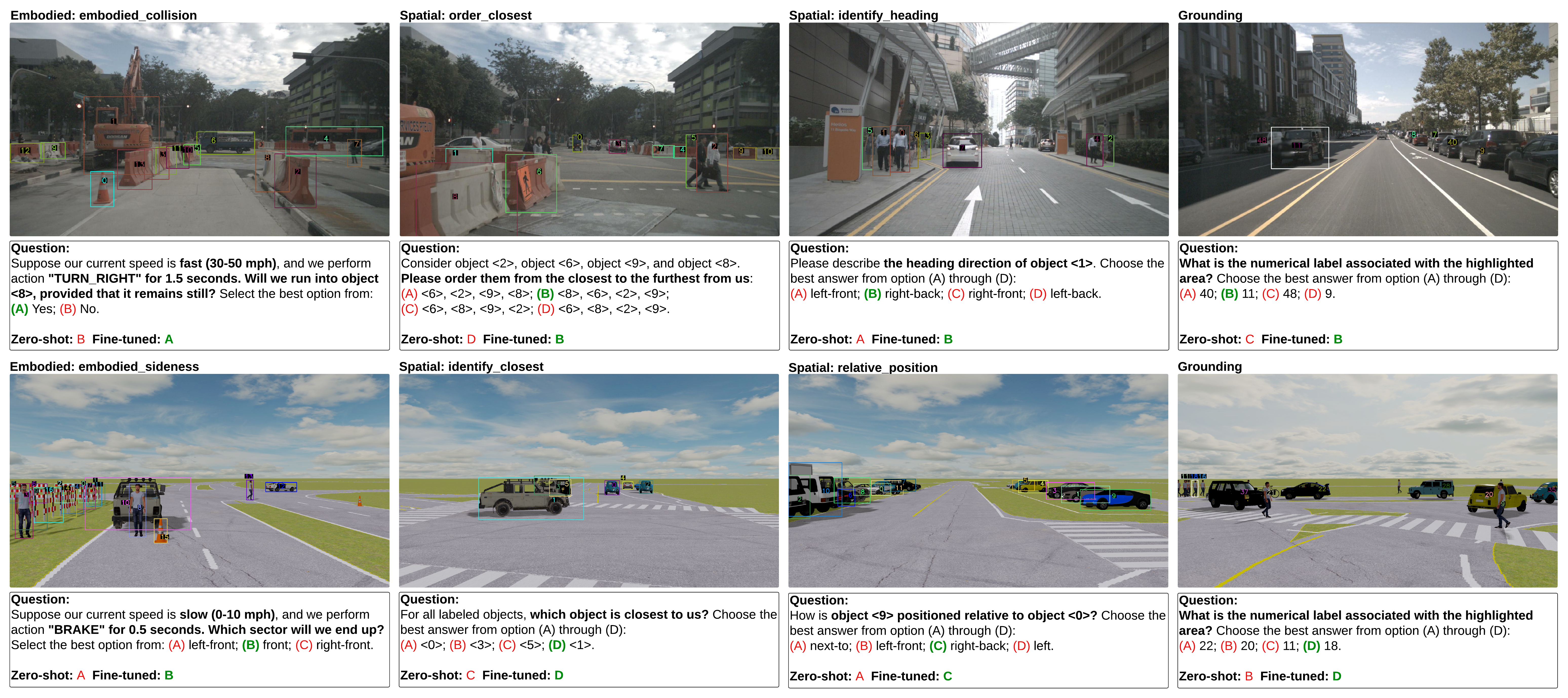}
    \vspace{-2em}
    \caption{\textbf{Improved embodied scene understanding after fine-tuning} of InternVL2-8B on the withheld training set from ~\cref{sec:composition}. The VLM demonstrates improved spatial understanding and embodied knowledge after learning the MetaVQA Dataset. In addition, the model attains better grounding capability.
    }
    \label{fig:vqa_demo}
    \vspace{-1.0em}
\end{figure*}

\begin{figure*}[!t]
    \centering
    \includegraphics[width=\linewidth]{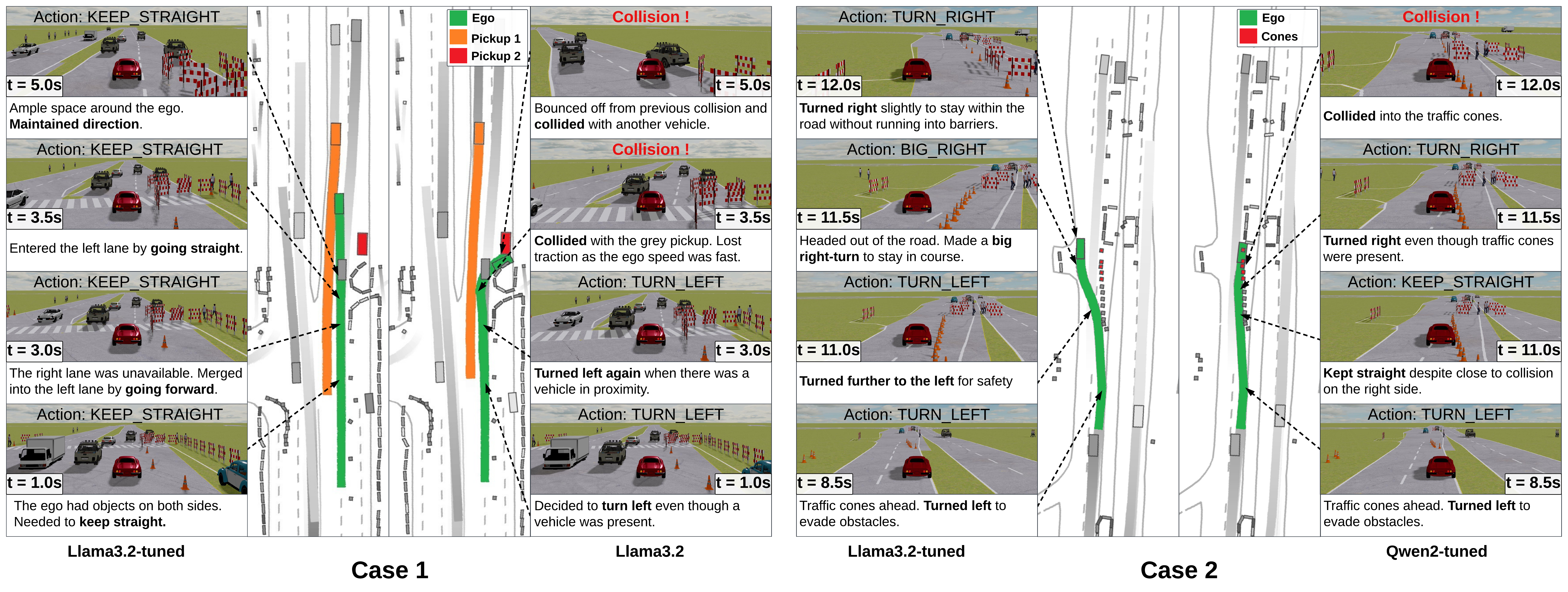}
    \vspace{-2em}
    \caption{\textbf{Qualitative result of closed-loop evaluation}. Case 1 compares the performance of fine-tuned Llama3.2 (left) versus its zero-shot counterpart (right) in the same scenario. Case 2 compares the performance of fine-tuned Llama3.2 (left) versus fine-tuned Qwen2 (right). As shown, fine-tuned Llama3.2 gains elevated situational awareness and can avoid collision. It also demonstrates superior safety capability compared to its trained peers.}
    \label{fig:closed_loop_demo}
    \vspace{-1.5em}
\end{figure*}

\section{Benchmark Results}
\subsection{Visual Question Answering}

\paragraph{Task formulation \& Metrics.} Under this task, VLMs receive multiple-choice questions and corresponding images annotated with Set-of-Marks prompting. The model is trained to select the best matching options provided in the question bodies. As shown in ~\cref{fig:vqa_sample}, the model is instructed to answer in single capitalized characters. We evaluate the model based on its failure rate (of outputting valid responses) and the accuracy of the selected options. Implementation for the answer parsing will be discussed in the supplementary materials.
% \paragraph{Experiments setup.}
We benchmark the zero-shot performance of state-of-the-art VLMs on the curated test set mentioned in \cref{sec:composition}. In addition, we establish the performance of VLMs fine-tuned on the training set mentioned in \cref{sec:composition}.

\begin{table}[!t]
  \centering
\begin{scriptsize}
  \begin{tabular}{lcccc}
    \toprule
    Model                  & Overall & Sim   & Real  & Parse Fail$\downarrow$  \\
    \midrule
    \textbf{Random}               & 0.329   & 0.326 & 0.332 & 0.0000     \\
    \textbf{LLaVA}-NeXT      & 0.295   & 0.287 & 0.302 & 0.2750     \\
    \textbf{LLaVA}-OneVision  & 0.581   & 0.550 & 0.613 & 0.0000     \\
    \textbf{GPT}-4o                 & 0.628   & 0.602 & 0.655 & 0.0004     \\
    \midrule
    \textbf{Qwen2}    & 0.539   & 0.527 & 0.552 & 0.0000     \\
    \textbf{Qwen2}-finetuned    & 0.844   & 0.839 & 0.848 & 0.0000     \\
    \midrule
    \textbf{Llama3.2} & 0.500   & 0.478 & 0.523 & 0.0080     \\
    \textbf{Llama3.2}-finetuned & 0.774   & 0.744 & 0.803 & 0.0672    \\
    \midrule
    \textbf{InternVL2}-8B            & 0.592   & 0.552 & 0.632 & 0.0000     \\
    \textbf{InternVL2}-8B-finetuned & \textbf{0.869}   & \textbf{0.853} & \textbf{0.884} & 0.0000    \\ 
    \bottomrule
  \end{tabular}
  \end{scriptsize}
      \caption{\textbf{Visual question answering benchmark}. Performance comparison of different models on overall, simulation-only-part, and real-only-part of the withheld test sets. The parsing failure rate is also provided. Models report consistent improvements after fine-tuning, with InternVL2-8B achieving the best performance.
  }
  \vspace{-5pt}
  \label{tab:model_performance}
\end{table}

\paragraph{Benchmarks.}
\begin{figure}[!t]
    \centering
    \includegraphics[width=\linewidth]{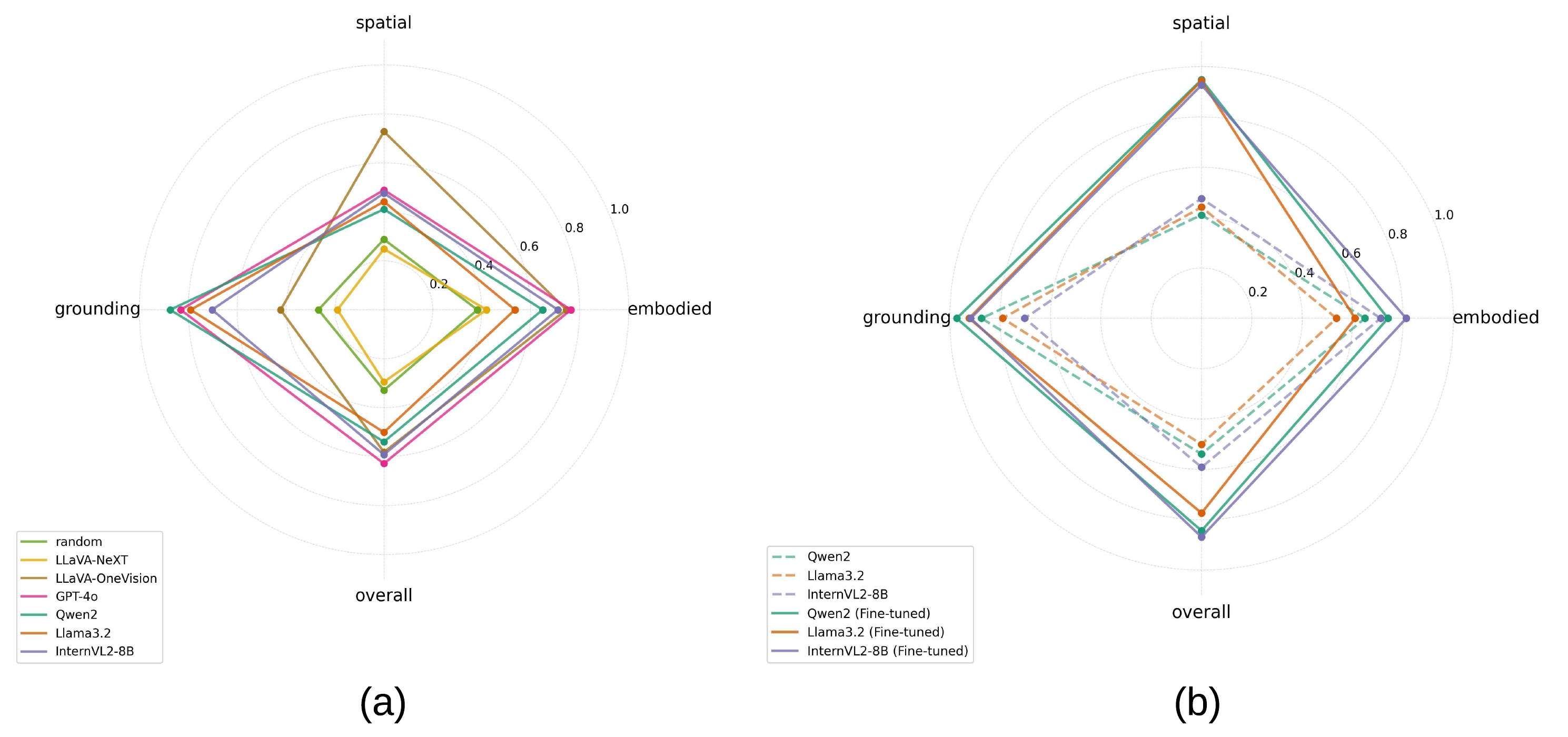}
    \vspace{-2em}
    \caption{\textbf{Performance contours of VLMs}. Sub-figure (a) draws the zero-shot performance comparisons, and sub-figure (b) illustrates the fine-tuning improvments. As illustrated, fine-tuned models perform consistently better on all supertypes of questions, and the improves in spatial understanding are especially pronounced. In addition, models report comparable increases across question supertypes, suggesting that the MetaVQA Dataset is generally learnable.
    %\pzh{Where is the results for baselines? Should have another diamond figure comparing un-finetuned baselines.}
    }
    \label{fig:radars}
    \vspace{-1.5em}
\end{figure}

\label{sec:benchmark}
~\cref{tab:model_performance} lists the performance of various VLMs on the withheld test set, and ~\cref{fig:radars} visualize the performance across question supertypes (along with the total test accuracy in the ``overall" dimension).  
As shown, most models successfully generate valid token sequences, and their zero-shot performances are significantly better than random guessing, indicating that some embodied scene understanding capabilities are already present. GPT-4o achieves the best zero-shot performance as it contains the largest parameters with an unprecedented scale of pre-training data. In addition, the low parse fail rate in most baselines validates the answerability of questions. Noticeably, LLaVA-NeXT ~\cite{liu2023improved} is the only outlier, with a zero-shot accuracy lower than random guessing. The shocking underperformance of LLaVA-NeXT is attributed to two factors. To begin with, the VLM fails to generate legal answer tokens consistently (failure rate of 27\%). In addition, the model, quite frequently, refuses to answer the asked questions. For example, when asked with \texttt{relative\_distance} questions, LLaVA-NeXT typically responds with ``...I cannot provide a definitive answer to your question without more information about the simulation or the specific positions of the objects within it." Consequently, LLaVA-NeXT reports surprisingly bad metrics.

The subfigure (b) of ~\cref{fig:radars} emphasizes the improvements of the benchmarked VLM after fine-tuning. The dotted contours draw out the zero-shot performance of VLMs, and the solid contours describe their fine-tuned counterparts. As shown, fine-tuning on the MetaVQA Dataset results in elevated embodied scene understanding in general, with the most pronounced improvements observed in spatial questions. In addition, comparable gains in accuracy (along each question supertype) are reported across models, suggesting the generalizability of learning for the dataset. We take InternVL2-8B, the best performing VLM after fine-tuning, for a case study: ~\cref{fig:vqa_demo} showcases questions successfully answered by the model after fine-tuning. The model not only gains spatial awareness with a better understanding of pedestrians' projected heading but also shows improved attentive power by reasoning about multiple observed objects. In addition, InternVL2-8B shows improved embodied knowledge: it forecasts the potential hazard of running into traffic barriers and predicts the consequence of its action. Collectively, the improvements in both spatial reasoning and embodied knowledge of VLMs demonstrates the contribution of the MetaVQA Dataset, and we will further validates its effectiveness in ~\cref{closed_loop_formulation}.

\subsection{Closed-loop Evaluation}
\label{closed_loop_formulation}
\begin{figure}[!t]
    \centering
\includegraphics[width=\linewidth]{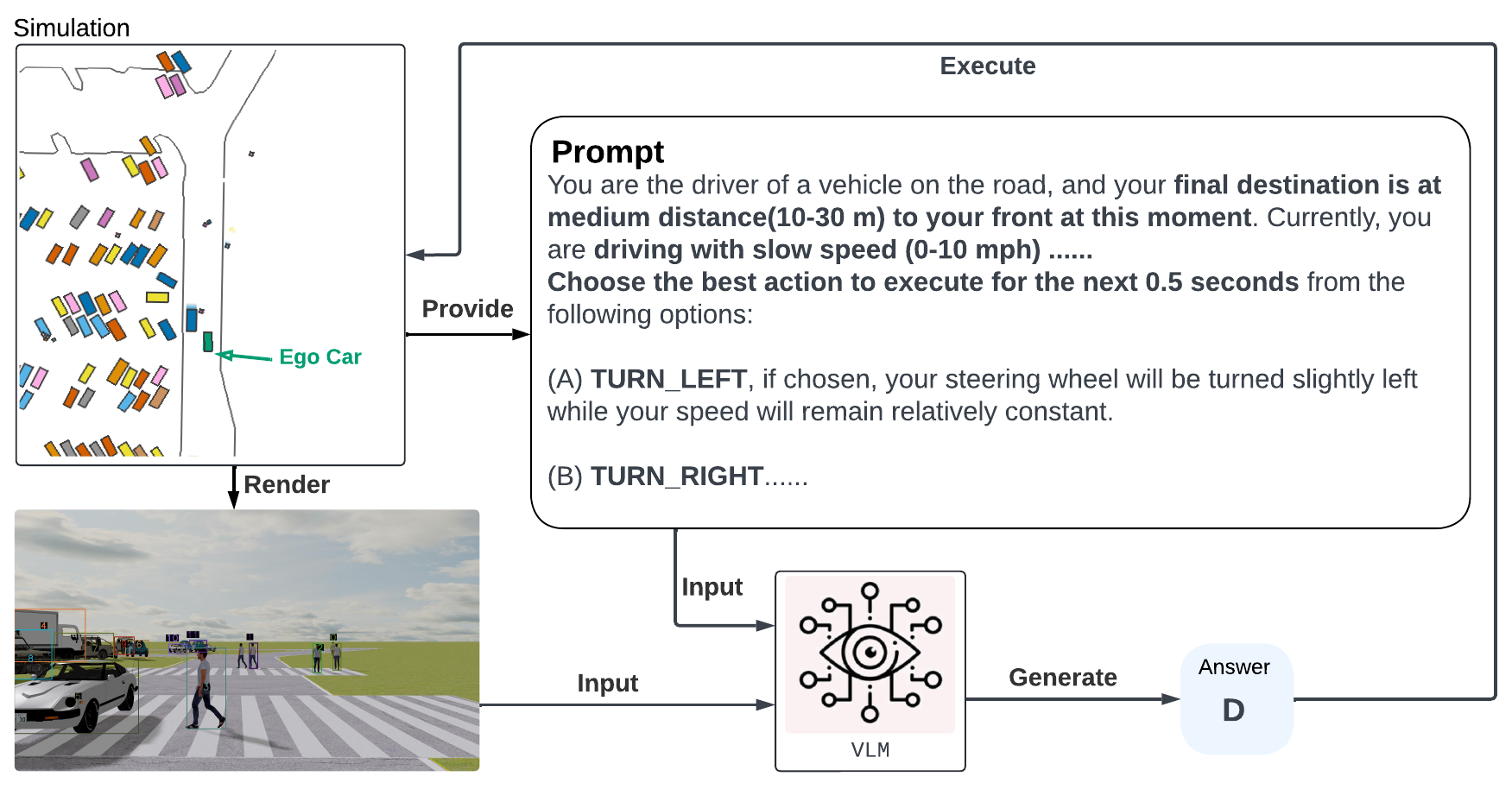}    
\vspace{-1em}
\caption{\textbf{Formulation of closed-loop evaluation}. At every five simulation steps (0.5 seconds wall time), the evaluated VLM is provided with annotated observations and current navigation command. The chosen action will be fed into the simulation.}
\label{fig:closed_loop_paradigm}
\vspace{-1.5em}
\end{figure}

\paragraph{Task formulation.} We use the MetaDrive~\cite{li2022metadrive} simulator and load scenarios from real-world datasets through ScenarioNet~\cite{li2023scenarionet} to create interactive environments to test VLMs as embodied agents. In this task, VLMs are tasked as the self-driving planners of vehicles. \cref{fig:closed_loop_paradigm} illustrates the interaction paradigm.
At every five simulation steps (0.5s in the wall time), the tested VLM receives a first-person view image annotated in the Set-of-Mark convention. The destination, current speed, and allowed actions are provided as textual prompts. Each action is mapped to a fixed value pair of steering and acceleration, which will be fed into the simulator to for vehicle dynamic simulation. Further details are discussed in the supplementary materials.
For generalizable evaluations across varied situations, we augment real-world scenarios with safety-critical scenarios generated using CAT~\cite{zhang2023cat}, in which an adversarial agent will intercept the trajectory of the ego vehicle. 
During evaluation, the trial is considered a failure if the tested VLM drives off the roads, upon which the trial terminates. Details on implementation can be found in the supplementary materials. 

\paragraph{Experiments Setup.} We curate 120 driving scenarios, with 60 scenarios selected from the nuScenes dataset and 60 safety-critical scenarios generated using CAT from WOMD's ~\cite{ettinger2021large} training split. All scenarios are handpicked to cover diverse behaviors such as stop-and-go, U-turn, and crossing intersections. We evaluate learned knowledge gained from the open-loop VQA task using the InternVL2 model family, Llama3.2, and QwenVL2, each trained on a shared condensed training dataset from the larger training set from ~\cref{sec:composition}. % curated with 50,000 questions using Waymo and nuScenes simulated observations. 

Noticeably, the prompt format used in the closed-loop evaluation is absent form the MetaVQA Dataset. However, the skill sets required for driving--spatial awareness and embodied knowledge--are practiced by fine-tuning with the MetaVQA Dataset. Therefore, we expect fine-tuned VLMs should learn generalizable embodied scene understanding for the unseen driving task, and models are tasked as self-driving planners in a closed-loop setting.

\paragraph{Metrics.} For each evaluated VLM, we record the collision rate, defined as the ratio of scenarios in which the VLMs collided over the total number of scenarios, off-road rate, average displacement error (ADE), 
final displacement error (FDE), and route completion ratio. Furthermore, we append its accuracy on the test set mentioned in ~\cref{sec:composition} to examine the correlation between open-loop performance and closed-loop performance of VLMs.
Details on metrics implementation can be found in the supplementary materials.

\begin{table}[!t]
  \setlength{\tabcolsep}{3pt}
  \centering
    \resizebox{1.0\linewidth}{!}{
    \begin{scriptsize}
        \begin{tabular}{lcccccc}
            \toprule
            Model & \makecell{Test\\Accuracy$\uparrow$} & \makecell{Route\\Completion$\uparrow$} & \makecell{Off-Road\\Rate$\downarrow$} & \makecell{Collision\\Rate$\downarrow$} & \makecell{ADE$\downarrow$} & \makecell{FDE$\downarrow$} \\
            \midrule
            \textbf{random}            & \_  & 0.376            & 0.658          & 0.375         & 20.937    & 15.207   \\
            \textbf{brake}       & \_  & 0.188            & 0.167          & 0.533              & 23.225     & 44.659   \\
            \textbf{straight}          & \_  & 0.592            & 0.583          & 0.358        & 13.272      & 12.678  \\
            \midrule
            \textbf{Qwen2}       &  0.539  & 0.615        & 0.583         & 0.367         & \textbf{13.075}     &30.214  \\
            \textbf{Qwen2}-tuned &  \textbf{0.792}  & \textbf{0.667}      & \textbf{0.442}         & \textbf{0.300}      &14.873     & \textbf{27.973} \\
            \midrule
            \textbf{Llama3.2}    &  0.500  & 0.529        & 0.658         & 0.483         & 18.335     &40.665  \\
            \textbf{Llama3.2}-tuned &  \textbf{0.757}  & \textbf{0.632}      & \textbf{0.558}         & \textbf{0.267}      &\textbf{15.118}     & \textbf{31.811} \\
            \midrule
            \textbf{InternVL2}-4B~\cite{chen2024internvl}      &  0.507  & 0.259        & 0.850          & \textbf{0.225}         & 24.824     & 53.020  \\
            \textbf{InternVL2}-4B-tuned &  \textbf{0.719}  & \textbf{0.614}      & \textbf{0.500}         & 0.317      & \textbf{14.368}    & \textbf{29.943}  \\
            \midrule
            \textbf{InternVL2}-8B       &  0.592  & 0.637        & 0.583         & \textbf{0.325}         & 13.361     &30.520  \\
            \textbf{InternVL2}-8B-tuned &  \textbf{0.820}  & \textbf{0.657}      & \textbf{0.517}         & 0.367      &\textbf{13.109}     & \textbf{27.873} \\
            \bottomrule
        \end{tabular}
    % }
    \end{scriptsize}
    }
    \vspace{-0.5em}
    \caption{\textbf{Quantitative result of closed-loop evaluation}. Despite not being directly trained on the driving task, VLMs report improvements in closed-loop metrics after learning the MetaVQA Dataset, in addition to better VQA accuracy. This correlation suggests that the MetaVQA Dataset contains generalizable embodied knowledge that could be easily learned and transferred to the downstream application domain (in this case, self-driving). Note that the fine-tuning set is a condensed version of the training set specified in ~\cref{sec:composition} to expedite experiments.
    }
    \vspace{-1.5em}
    \label{tab:close-loop}
\end{table}

\paragraph{Discussions.} 
As presented in ~\cref{tab:close-loop}, all fine-tuned VLMs exhibit significantly better closed-loop performance in route completion, off-road ratio, and FDE, in addition to improved VQA accuracy. With a few exceptions, models also demonstrate improved ADEs and collision rates in general. We speculate these inconsistencies are a by-product of varied pre-training strategies adopted by each VLM, leading to differed learning of situational awareness. The improved zero-shot closed-loop performance of fine-tuned VLMs confirms our expectation: general-purposed VLMs become stronger adaptive embodied agents after learning the MetaVQA Dataset.

~\cref{fig:closed_loop_demo} provides case comparisons of VLMs in closed-loop evaluation, with the ego trajectory drawn in green while those of interested objects in orange or red. The decision-making of fine-tuned Llama3.2~\cite{dubey2024llama3herdmodels}, the safest driver after fine-tuning, is plotted on the left side in both cases. In case 1, Llama3.2's behavior before fine-tuning is plotted against on the right. Without the open-loop VQA pre-training, naive Llama3.2 fails to realize the danger of turning-left when a pickup was present on that side, resulting in two consecutive collisions. In comparison, fine-tuned Llama3.2 becomes more aware of its surroundings. It deduces that the safest action, when surrounded by obstacles on both sides, is to main direction, and it successfully merges into the unblocked left lane by keeping straight consistently. In case 2, the right side illustrates the trial for fine-tuned Qwen2 ~\cite{Qwen2VL}. Both fine-tuned VLMs demonstrate situational awareness by merging into the left lane at wall time 8.5 second, when traffic cones are present ahead. However, they improve to varied degrees: Llama3.2 keeps going left until there it is sufficiently distant from the obstacles, upon when it makes a right turn to stay within the road. In comparison, Qwen2 becomes oblivious to the traffic cones to its right, and it decides to turn right when there is no space, leading collisions. In both comparisons, consisting of four trials, the VLMs receive the same navigation command ``forward" throughout simulation (complying with the ground-truth trajectories), meaning that the evasions are made on-the-go without behavioral hints from the navigation command. We further supports from improved closed-loop evaluations, we are confident to claim that the MetaVQA Dataset contributes to general-purpose VLMs' embodied scene understanding.

%% file: sec/6_conclusions.tex
\section{Conclusion}
We present MetaVQA, a large-scale benchmark for evaluating and improving the embodied scene understanding of vision language models. Besides establishing the baseline performance of representative VLMs, we showcase the transferable knowledge learned from the MetaVQA Dataset across observation domains through sim-to-real VQA evaluations. Finally, we further evaluate the fine-tuned VLMs in the untrained task of closed-loop driving in simulation, and the significantly improved driving capabilities confirm the generalizability and robustness of learned embodied knowledge from the MetaVQA Dataset.

\mypar{Limitations.} Currently, MetaVQA contains only images as observations, but embodied agents might need multi-step historical information to make the optimal decision in complex situations. In addition, the MetaVQA Dataset contains only single-perspective observations captured from fixed angles, while multi-camera observations could provide more contextual information for better decision-making.

%% file: sec/X_suppl.tex
\clearpage
% \setcounter{page}{1}
% \maketitlesupplementary

\section{VQA Generation Pipeline}
\subsection{Scenario Aggregation from Multiple Sources}
\label{sec:scenes}
\paragraph{Scene collection of nuScenes dataset.}
To collect nuScenes scenarios with the original observations (nuScenes-real), we use the Python implementation of nuscenes-devkit ~\cite{caesar2019nuscenes} to explore traffic scenarios. Following the naming paradigm provided in the official nuScenes documentation, for each \texttt{sample} in a \texttt{scene}, we extract its \texttt{CAM\_FRONT sampled\_data}. At this point, we can associate a recorded keyframe with its image observation. In the first filtering process, if an object is annotated with than a ``3" level of visibility or is scanned by less than five rays of Lidar, then it is considered "invisible," and its information will not be recorded. However, this first pass doesn't consider visual occlusion since the visibility of objects is annotated on the scene level instead of the frame level in nuScenes. Therefore, a second filtering pass is instigated. The nuscenes-devkit provides API to project 3D bounding boxes(8 vertices) of objects onto the 2D image observations, and we create the maximum enclosing 2D filled bounding boxes of the 8 vertices. Then, these filled rectangles are painted onto a black image following the distance order of objects to mimic the process of z-buffering, and boxes of distant objects will be overlayed by closer objects. Finally, we filter out objects with less than 50\% of their 2D boxes visible in the composed image, completing the second filtering pass. The third and last filtering pass removes miscellaneous objects such as \texttt{debris} and \texttt{vegetation} from objects of interest. After these three filtering stages, the interested objects set will have the information mentioned in the color box to the right recorded. Notably, the nuScenes dataset doesn't have the ``color" annotation, and we leave this field empty while collecting the scenarios.

\paragraph{Scene reconstruction with simulator.}
Leveraging the MetaDrive~\cite{li2022metadrive} simulator and ScenarioNet~\cite{li2023scenarionet} data platforms, we aggregate nuScenes~\cite{caesar2019nuscenes} and Waymo~\cite{ettinger2021large}. For simulator-reconstructed traffic scenarios, we record frames every five steps (0.5 seconds wall time) until the end. We set a camera with a 60-degree field-of-view and 1920 $\times$ 1080 resolution to extract rendering. At each simulation step, we record the following information about the ego and objects within 75 meters of the ego: 

{
\captionsetup{type=table}
\begin{tcolorbox}[colback=gray!10,%gray background
                  colframe=black,% black frame color
                  width=\linewidth,
                  arc=1mm, auto outer arc,
                  boxrule=0.5pt,
                 ]
{\color{gray}
\texttt{\textcolor{blue}{Information Recorded per Frame:}}

\texttt{} \textbf{id}, assigned by the simulator.\texttt{}

\texttt{} \textbf{color}, bound to the 3D asset.\texttt{}

\texttt{} \textbf{height}, bound to the 3D asset.\texttt{} 

\texttt{} \textbf{type}, bound to the 3D asset.\texttt{} 

\texttt{} \textbf{bounding box} in world coordinates.\texttt{} 

\texttt{} \textbf{heading vector} in world coordinates.\texttt{} 

\texttt{} \textbf{speed} of the object in meters per second (m/s).

\texttt{} \textbf{position} of the center point in world coordinates.

\texttt{} \textbf{ego camera that observes the vehicle} (if any).

\texttt{} \textbf{visibility} of the object to the ego vehicle. 

\texttt{} \textbf{collided objects} (if any) at this moment.
}
\end{tcolorbox}
\vspace{-0.5em}
}

Note that if an object is "visible," the camera must capture at least 1,200 pixels of its body. This is implemented by assigning an ID color to each active object in the simulation, and we use a special instance segmentation camera (the same intrinsic and placement as the capturing camera) to capture the ID-color-based rendering. The traffic collected has the following statistics. 

% \begin{table}[!h]
%     \vspace{-10pt}
%     \centering
%     \resizebox{\linewidth}{!}{
%     \begin{tabular}{l cccc}
%         \toprule
%         Dataset & \makecell{\# Frames} &\makecell{\# Images} &\makecell{\# Scenarios}\\ 
%         \midrule
%         Waymo           & 199,133 (47.0\%)  & N &     6,900   (51.8\%) \\
%         nuScenes-sim    & 90,587  (21.4\%)  & N &     400    (6.9\%)\\
%         nuScenes-real   & 133,598 (31.6\%)  & N &     400   (41.3\%)\\
%         \midrule
%         Total           & 423,318     & N &           7,700   \\
%         \bottomrule
%     \end{tabular}
%     \caption{\textbf{Distributions of aggregated scenarios.}}
%     \label{tab:collected_scenarios}
%     }
%     \vspace{-15pt}
% \end{table}
%\begin{table}[!h]
%    \centering
%    \scriptsize % Adjust font size
%    \begin{tabular}{lccc} % Fix the number of columns
%        \toprule
%        Dataset & \# Frames & \# Images & \# Scenarios \\ 
%        \midrule
%        Waymo & 199,133 (47.0\%) & -- & 6,900 (51.8\%) \\
%        nuScenes-sim & 90,587 (21.4\%) & -- & 400 (6.9\%) \\
%        nuScenes-real & 133,598 (31.6\%) & -- & 400 (41.3\%) \\
%        \midrule
%        Total & 423,318 & -- & 7,700 \\
%        \bottomrule
%    \end{tabular}
%    \caption{\textbf{Distributions of aggregated scenarios.} \td{collect stats}}
%    \label{tab:collected_scenarios}
%\end{table}

\begin{figure}[!h]
    \centering
    \includegraphics[width=\linewidth]{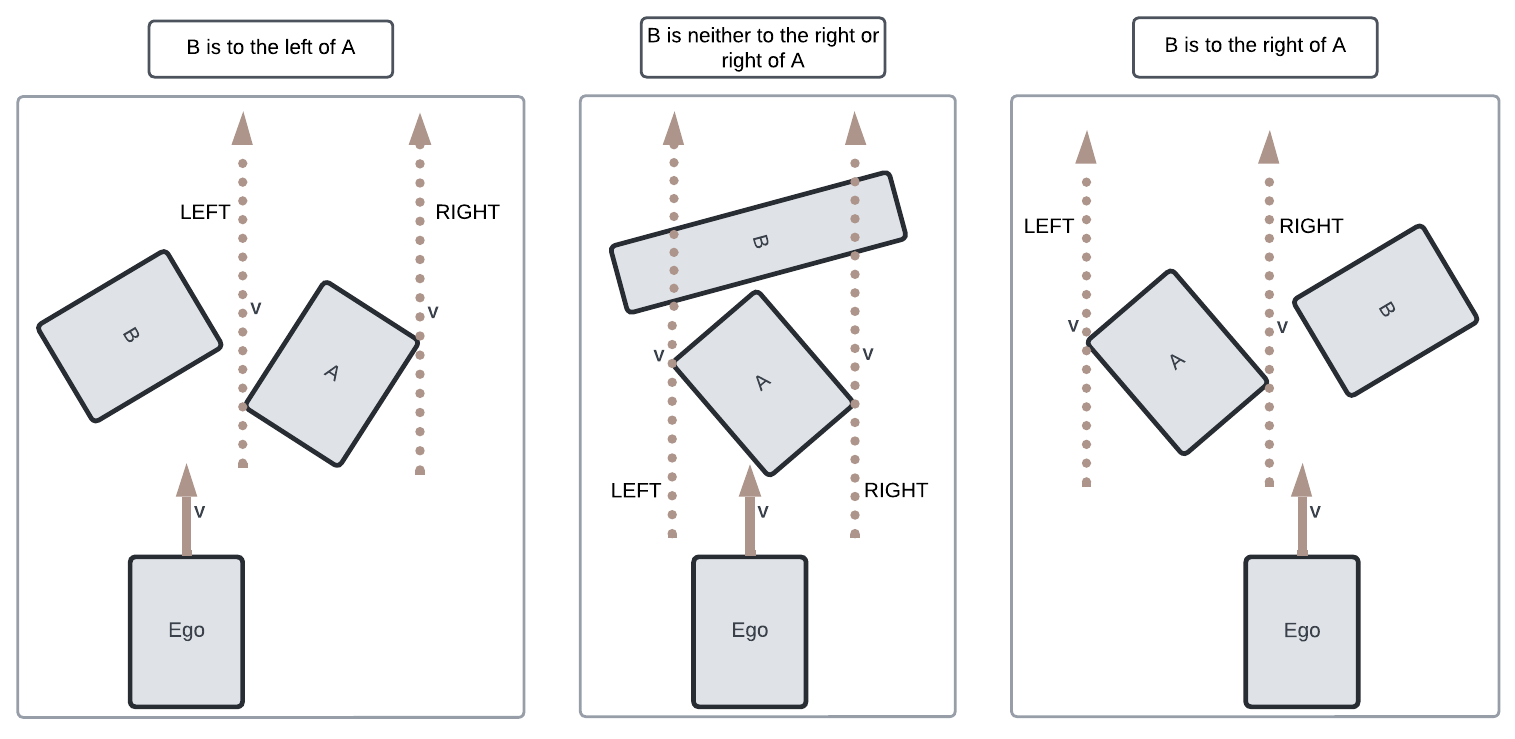}
    \caption{\textbf{Top-down illustration of sidedness.} We demand all vertices of box B reside in the ``LEFT'' region of A for B to be considered ``to the left of'' (and similarly for ``to the right of'') A. }
    \label{fig:sidedness}
    \vspace{-5pt}
\end{figure}

\paragraph{Constuction of 3D scene graphs.}
Each scene graph comprises nodes connected by directed edges representing relative spatial relationships. Each node corresponds to a visible object from the frame information recorded from the previous step, and intrinsic properties (\textit{e.g.}color, height) are contained in the node. Given a reference vector \texttt{V}, we determine the relative spatial relationships between current node \texttt{A} and node \texttt{B} by:
{
\captionsetup{type=table}
\begin{tcolorbox}[colback=gray!10,%gray background
                  colframe=black,% black frame color
                  width=\linewidth,
                  arc=1mm, auto outer arc,
                  boxrule=0.5pt,
                 ]
{\color{gray}
\texttt{\textcolor{blue}{Relative Spatial Relationships(box A,B;front vector V):}}

\texttt{}

\texttt{} \textbf{left or right}. Refer to~\cref{fig:sidedness}, and we determine the leftmost and rightmost vertices of bounding box A using the reference vector V as the front direction. Then, if all vertices of bounding box B are to the left of the leftmost vertex of A, then we consider B's sidedness to be "left" (and similarly for sidedness to be "right"). If bounding box B satisfies neither of the two conditions, then we consider B's ``sidedness'' to be ``none''.

\texttt{} 

\texttt{} \textbf{front or back}. We determine this relationship similarly to determining ``left'' or ``right'', with the modification that V is the left direction. \texttt{} 
}
\end{tcolorbox}
}

This reference vector V is the heading of the ego vehicle when determining ``left or right'', and it's rotated 90 degrees counterclockwise with respect to the yaw axis when determining ``front or back''. Once we have the two values for ``left or right'' and ``front or back'', we draw the corresponding directed edge from A to B from the following:
{
\captionsetup{type=table}
\begin{tcolorbox}[colback=gray!10,%gray background
                  colframe=black,% black frame color
                  width=\linewidth,
                  arc=1mm, auto outer arc,
                  boxrule=0.5pt,
                 ]
{\color{gray}
\texttt{\textcolor{blue}{Named Spatial Edges:}}

\texttt{} \textbf{l}, corresponding to ``to the left of.''

\texttt{} \textbf{lb}, corresponding to ``to the left and behind.'' 

\texttt{} \textbf{lf}, corresponding to ``to the left and in front of.'' 

\texttt{} \textbf{b}, corresponding to ``behind.'' 

\texttt{} \textbf{f}, corresponding to ``in front of.'' 

\texttt{} \textbf{r}, corresponding to ``to the right.'' 

\texttt{} \textbf{rb}, corresponding to ``to the right and behind.'' 

\texttt{} \textbf{rf}, corresponding to ``to the right and in front of.''
}
\end{tcolorbox}
}   
For example, if ``l'' edge is chosen, this means ``B is to the left of A.''

\subsection{Set-of-Mark Prompting}
\begin{figure}[!h]
    \centering
    \includegraphics[width=\linewidth]{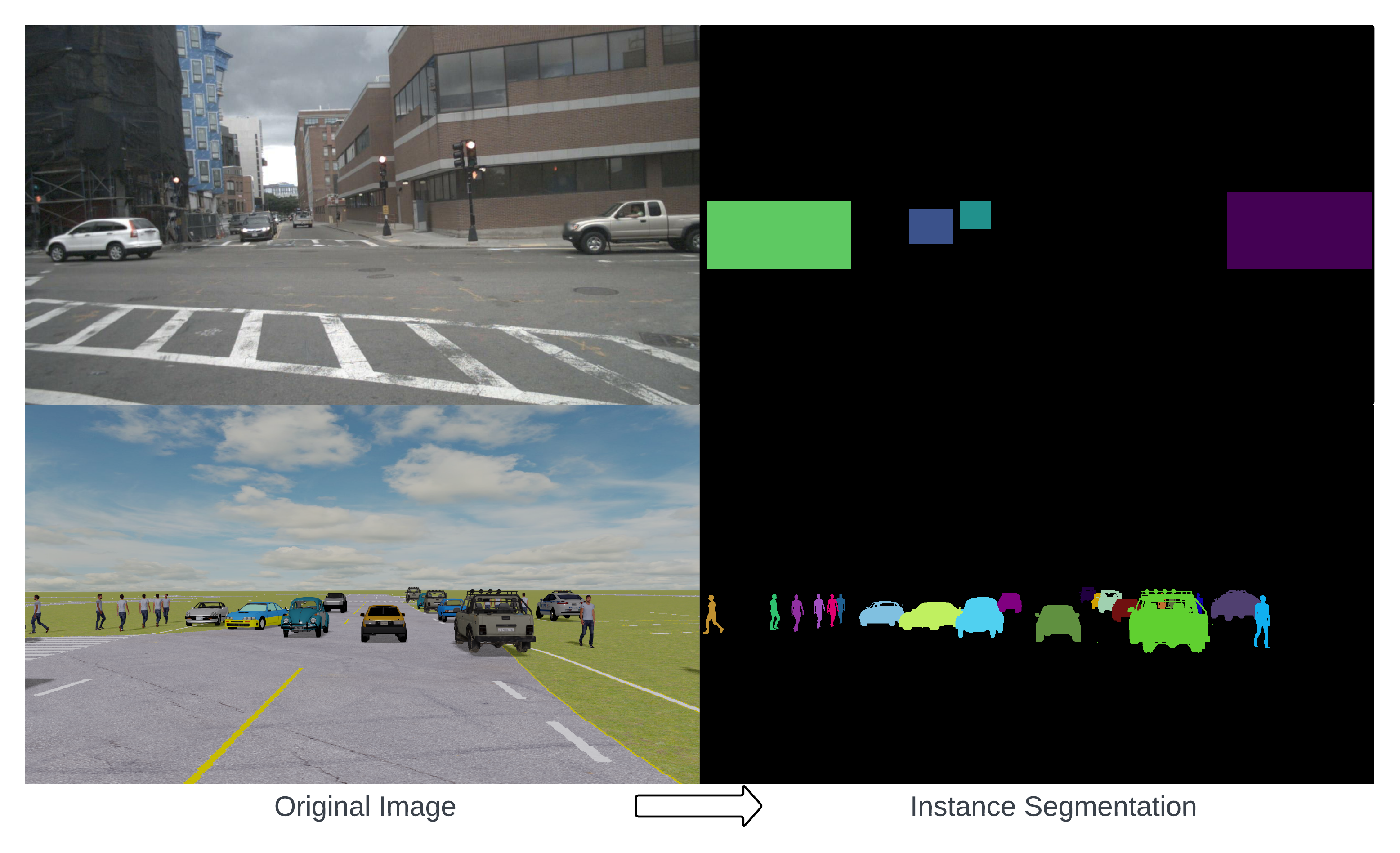}
    \caption{\textbf{Instance segmentation masks.} Approximated instance segmentation is generated for real images from the nuScenes dataset. Simulated images are paired with precise instance segmentation.}
    \label{fig:instances}
    \vspace{-5pt}
\end{figure}
From \cref{sec:scenes}, we have collected image observations and the corresponding instance segmentation in approximated boxes(nuScenes images) or shape-precise masks(simulated images), as shown in \cref{fig:instances}. Then, we run the algorithm illustrated in \cref{fig:som_algo} adopted from the original Set-of-Mark paper ~\cite{yang2023setofmark} to determine the appropriate position for object labels:
\begin{figure}[!h]
    \centering
    \includegraphics[width=\linewidth]{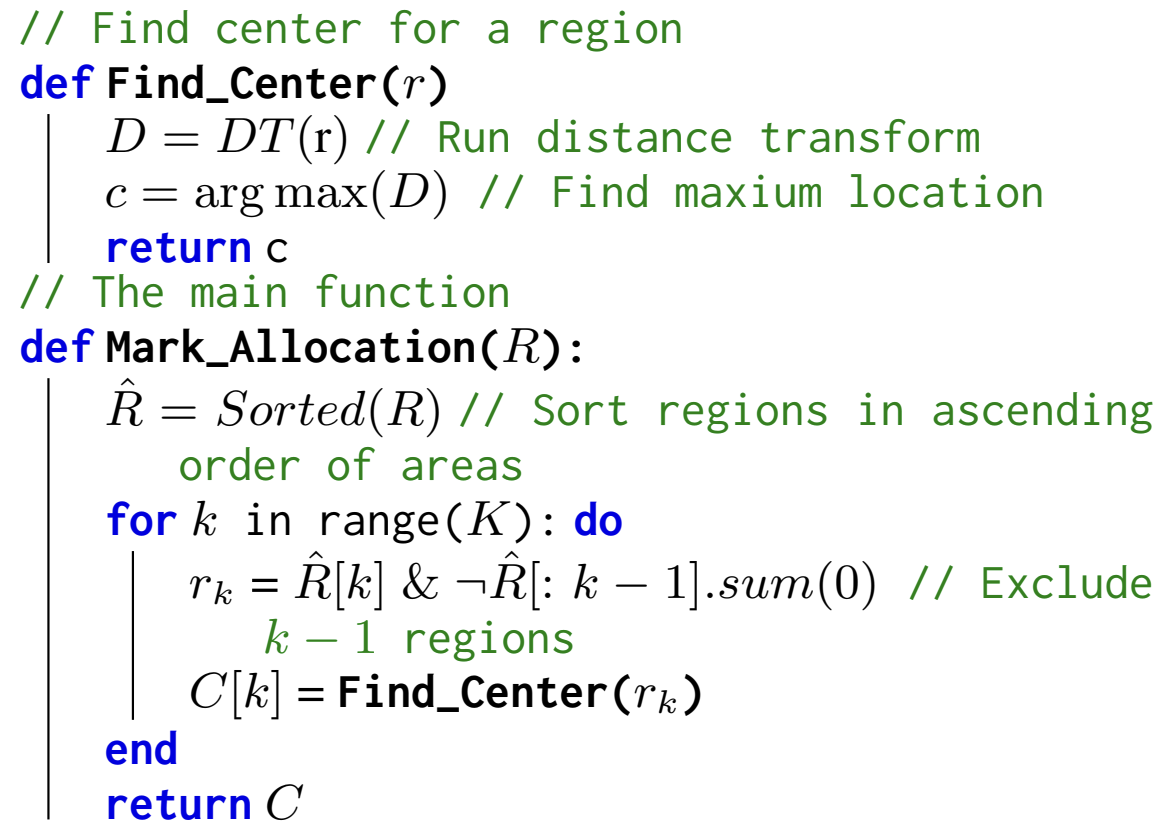}
    \caption{\textbf{Labeling algorithm adopted from the Set-of-Mark paper.} Credit to the Set-of-Mark authors.}
    \label{fig:som_algo}
    \vspace{-5pt}
\end{figure}

The Set-of-Mark paper suggested various schemes to perform the visual prompting. For example, using instance segmentation masks and contours are both valid schemes to improve the visual grounding capabilities of vision-language models (VLMs). As mentioned in the main paper, we conducted an ablation study on different prompting schemes to determine the optimal scheme for referral clarity using labels. Using Qwen2~\cite{Qwen2VL} as the zero-shot evaluating model, we fix the prompting scheme with bounding-box annotations, black text background color, and a text size of 1.00 (to reduce label occlusions). The bounding boxes and texts use colors identical to that of the instance segmentation masks of corresponding objects. We use \texttt{cv2.rectangle} to draw the bounding boxes onto original images with \texttt{thickness = 2}, and we use \texttt{cv2.putText} with \texttt{font\_size = 1} and \texttt{thickness = 2}. In addition, we slightly relocate the labels if their corresponding 2D bounding boxes enclose regions less than 1,600 pixels. This is to ensure the visibility of highlighted objects after the visual prompting. The concrete code implementations will be released.

\subsection{Question-Answer Generation}
\label{sec:qa_generation}

\subsubsection{Question Generation}
MetaVQA adopts a template-based question generation process. Each type of question is bonded to a single template with varying numbers and types of parameters to be replaced with concrete values. We categorize questions into ``non-parameterized" and ``parameterized " based on the number of parameter types in the template.

\begin{figure}[!h]
    \centering
    \includegraphics[width=\linewidth]{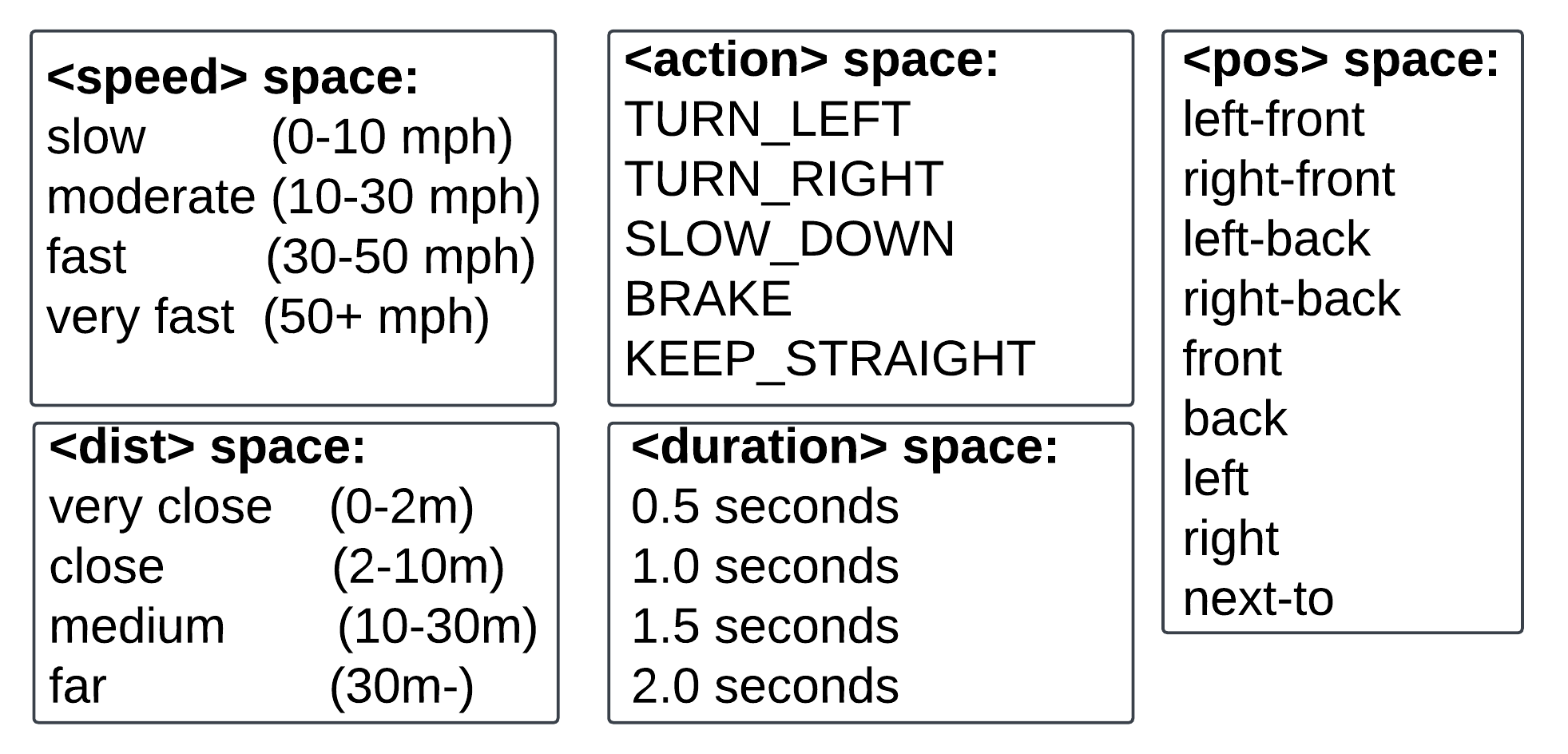}
    \caption{\textbf{Parameter space summary}. Note that space of \texttt{<id*>} is scenario-dependent, namely, all valid labels.}
    \label{fig:param_space}
    \vspace{-5pt}
\end{figure}

\begin{figure*}[!h]
    \centering
    \includegraphics[width=\textwidth]{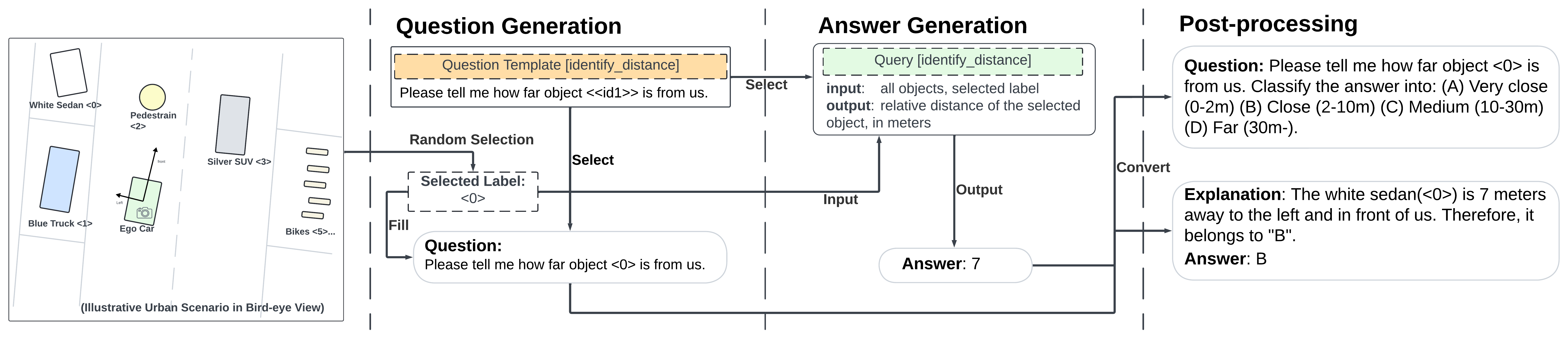}
    \caption{\textbf{Question-Answer generation of parameterized questions with only one type of parameter}.}
    \label{fig:qagen_param1}
\end{figure*}

\begin{figure*}[!h]
    \centering
    \includegraphics[width=\textwidth]{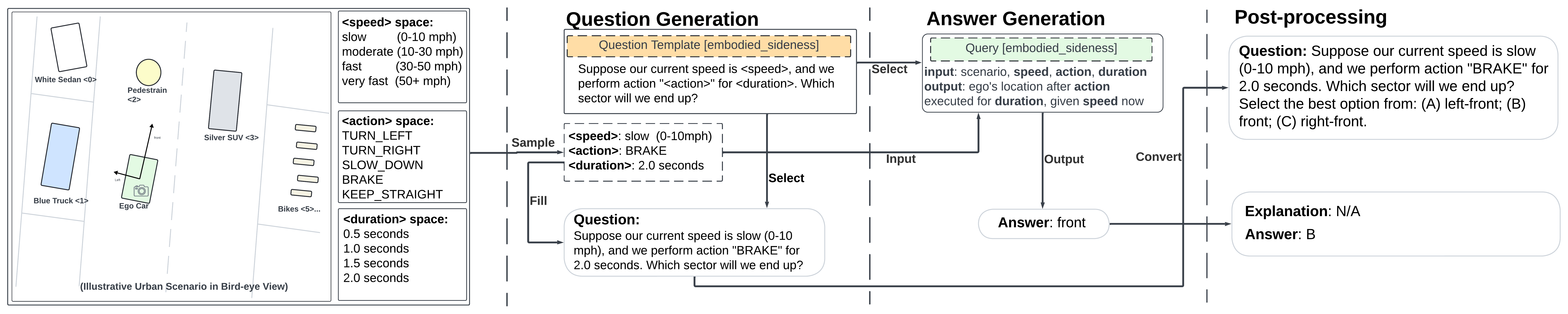}
    \caption{\textbf{Question-Answer generation of parameterized questions with distinct parameters}.}
    \label{fig:qagen_param2}
\end{figure*}

\paragraph{Parameterized question generation.}
Parameters are present for the templates of these questions. These parameters will be replaced upon question generation with concrete values selected from corresponding parameter spaces, the summary of which is provided in ~\cref{fig:param_space}. The generation process for a parameterized question is illustrated in ~\cref{fig:qagen_param1}: the template of \texttt{identify\_distance} contains a single \texttt{<id1>} parameter, the parameter space of which is all valid labels generated from the Set-of-Mark prompting. In this example, \texttt{<id1>} is replaced by the randomly selected label \texttt{<0>}. Additionally, multiple parameters belonging to different types can co-exist in a single-question template. Refers to ~\cref{fig:qagen_param2} for an illustration. Observe how concrete values for parameters are sampled from the parameter spaces.

\begin{figure*}[!h]
    \centering
    \includegraphics[width=\textwidth]{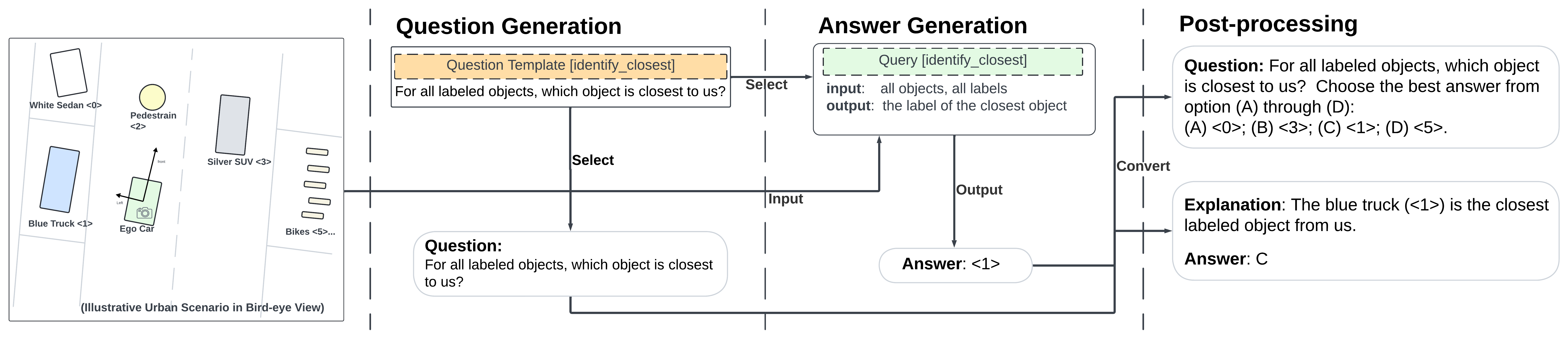}
    \caption{\textbf{Question-Answer generation of non-parameterized questions}.}
    \label{fig:qagen_nonparam}
\end{figure*}

\paragraph{Non-parameterized question generation.}
These questions don't have any parameters in their templates, as they demand the VLMs to examine all present objects in observations before answering. An example can be found in ~\cref{fig:qagen_nonparam}. Therefore, no computation is done in the question generation phase. 


\subsubsection{Answer Generation}
A unique query program is selected to generate answers for each type of question. Refer to ~\cref{fig:qagen_param1}, ~\cref{fig:qagen_param2}, and ~\cref{fig:qagen_nonparam} for examples. Upon the execution of these query programs, the concrete answers are extracted utilizing scenario information for simulated dynamics. Note that both the question-answer pairs at this stage are not formulated in the multiple-choice setting, and the next stage will reformat the pairs. 

\subsubsection{Post-processing}
At this point, question-answer pairs are already generated. The remaining works are (1) the generation of non-answer candidates for multiple-choice setup (2) the creation of the multiple-choice description strings which map choices with concrete answer candidates (3) the creation of optional ``explanation" strings to elevate VLMs' learning. Each question has different search spaces for non-answer candidates. As shown in ~\cref{fig:qagen_param1}, \texttt{identify\_distance}'s candidate space is the \texttt{<dist>} space listed in ~\cref{fig:param_space}, while that of \texttt{embodied\_sideness} is a subset of \texttt{<pos>} space, shown in ~\cref{fig:qagen_param2}. When applicable, non-answer candidates are selected to challenge the evaluated VLMs maximally. For example, candidate generation in question \texttt{identify\_type} prioritizes ones present in the scenarios on which the question is constructed. After the candidates' generation, they are put into multiple-choice format as suffixes to the original question, and the answer is replaced by the answer choice. The optional ``explanation" strings (used interchangeably with ``reasoning") are also programmatically created, depending on the choice-candidate mapping. Complete implementation will be included in the released codebase.

\section{MetaVQA Dataset}
\subsection{Dataset Composition}
\begin{figure*}[!h]
    \centering
    \includegraphics[width=\textwidth]{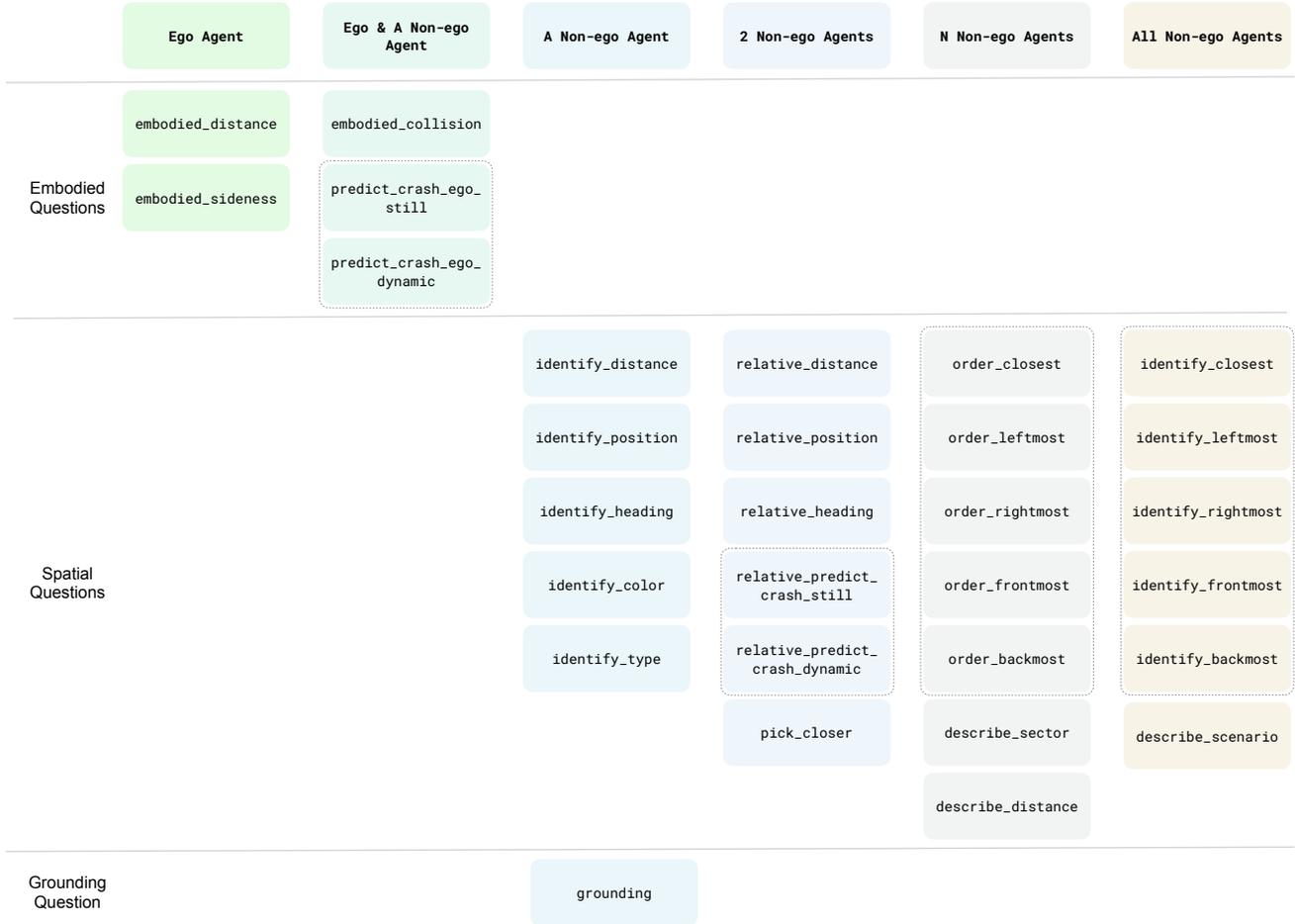}
    \caption{\textbf{Question taxonomy} of MetaVQA Dataset. Notice that questions are further blocked by black dotted contours to denote similarity in the formulation (illustrated collectively in ~\cref{fig:types}).}
\label{fig:q_taxonomy}
\end{figure*}

\cref{fig:q_taxonomy} list all question types divided along two dimensions. The horizontal dimension indicates the objects that need to be analyzed to answer the question successfully, and the vertical dimension indicates which facet of embodied scene understanding is evaluated.
Detailed descriptions--along with two examples using both simulated and real observations--for each question type can be found at the end of this document in ~\cref{fig:types}.

\subsection{Zero-shot Answerability with Set-of-Mark Prompting}
\subsubsection{Human Evaluation}
\begin{figure}[!h]
    \centering
    \includegraphics[width=\linewidth]{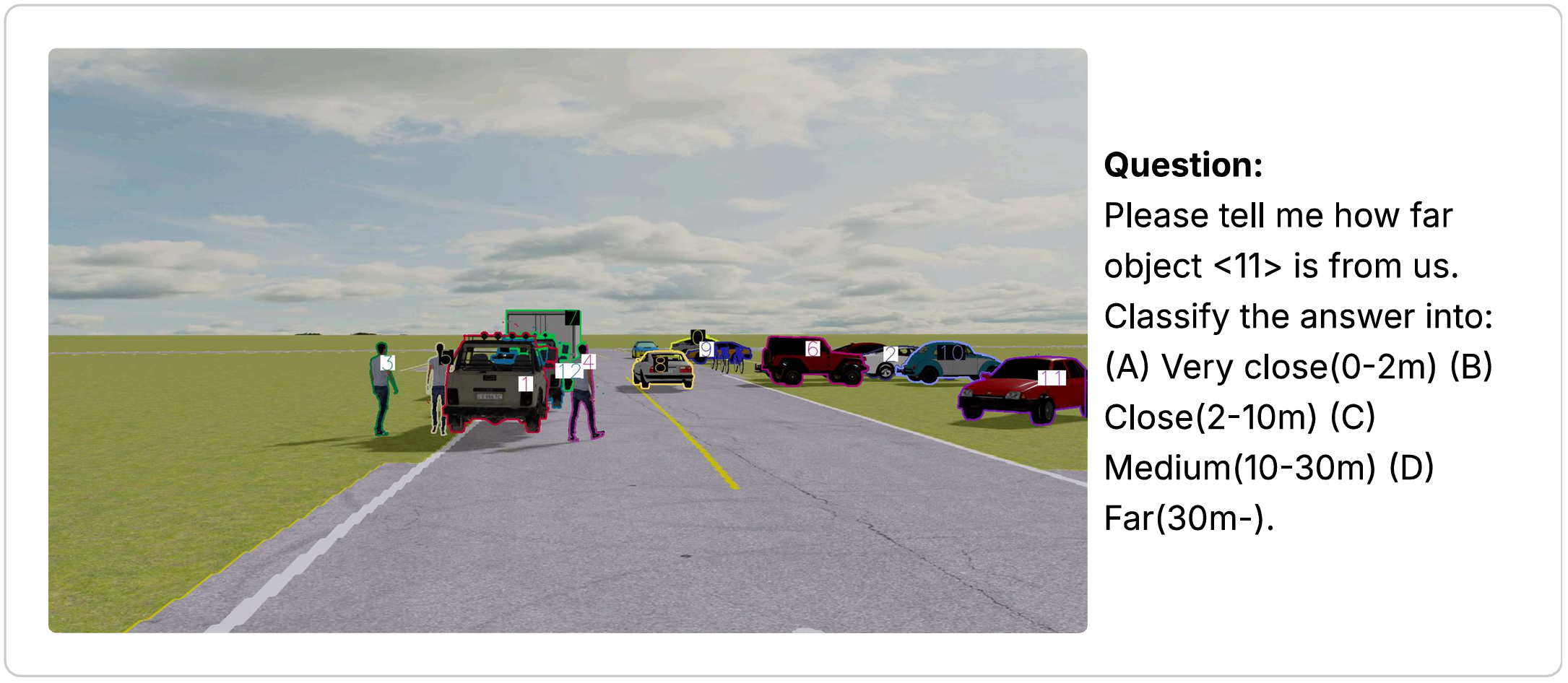}
    \caption{\textbf{Sample question from the questionnaire}. The answer is (C).}
    \label{fig:questionnaire}
    %\vspace{-15pt}
\end{figure}

Before large-scale dataset generation, we first prepare a small questionnaire to examine the answerability and the quality of the MetaVQA Dataset. Since this is a pilot study, we utilize a Set-of-Mark prompting scheme slightly different from the final MetaVQA Dataset: contours are drawn around objects, and the background color of each label is determined by the corresponding text color following the original paper ~\cite{yang2023setofmark}. We sampled 35 questions with distinct types generated from a single keyframe to speed up the evaluation process. Six participants report an average accuracy of 88.05\% on the 35 questions with a standard deviation of 7.54\%. The best-performing participant achieves a 94.2\& accuracy, while the worst-performing participant reports a 74.2\% accuracy. An example question from the questionnaire is illustrated in ~\cref{fig:questionnaire}. 

\begin{figure}[!h]
    \centering
    \includegraphics[width=\linewidth]{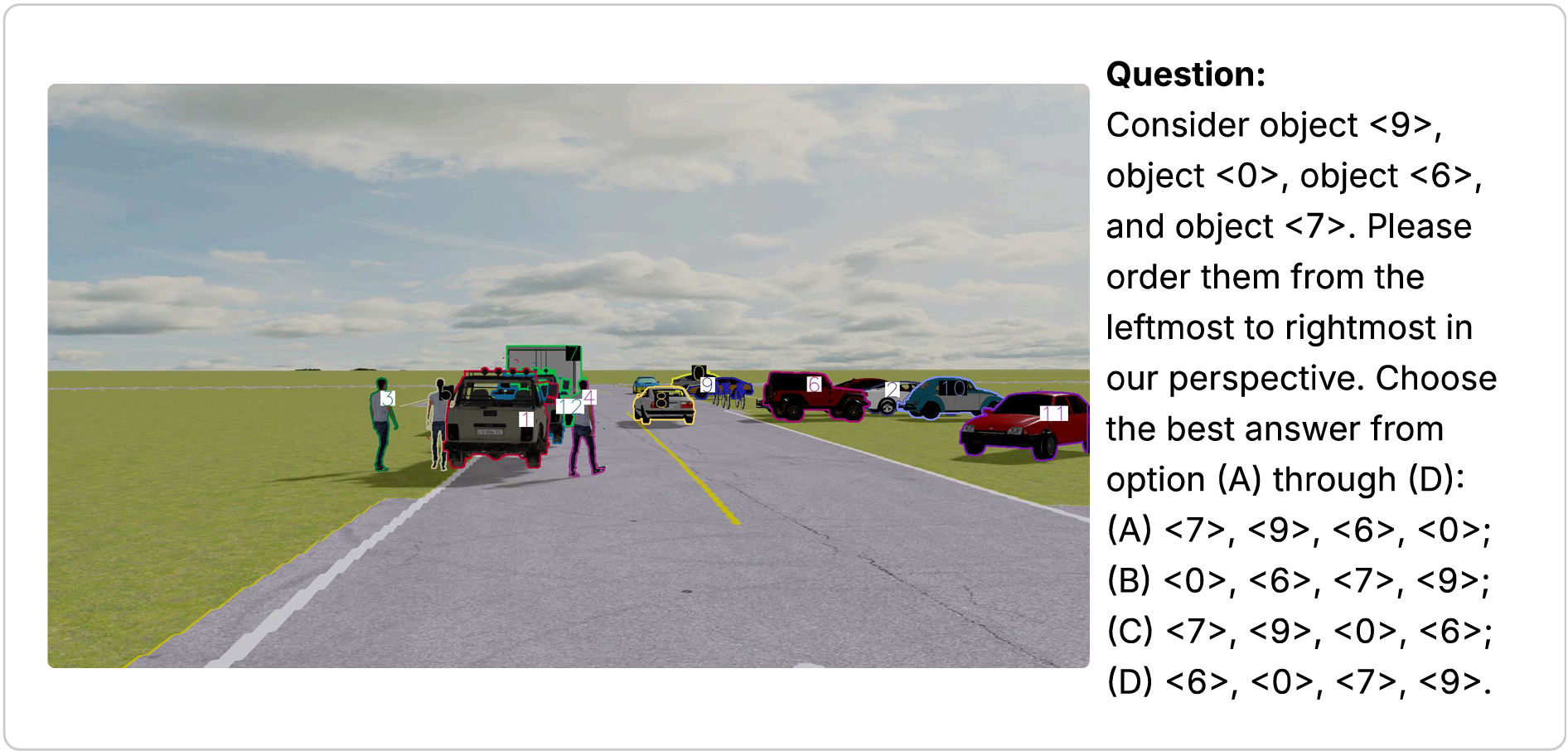}
    \caption{\textbf{Question 19 from the questionnaire}. The answer is (A). Ambiguous wording and distant objects lead to common mistakes by participants.}
    \label{fig:qa_19}
    \vspace{-10pt}
\end{figure}

\begin{figure}[!h]
    \centering
    \includegraphics[width=\linewidth]{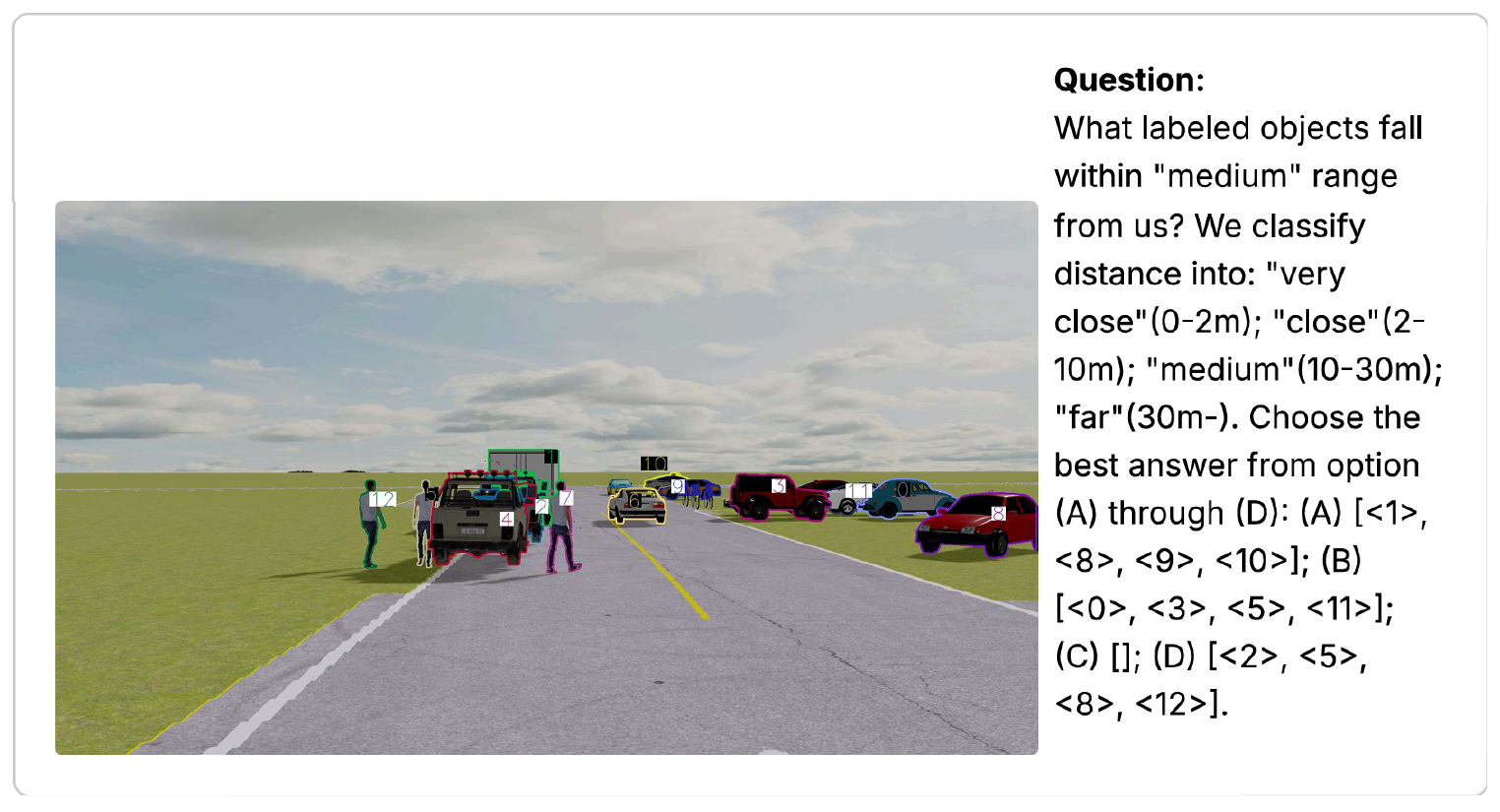}
    \caption{\textbf{Question 29 from the questionnaire}. The answer is (D). Some referred objects show limited visibility, leading to common errors.}
    \label{fig:qa_29}
    \vspace{-10pt}
\end{figure}

Noticeably, participants struggle with question 19 (5 out of 6 wrong) and question 29 (4 out of 6 wrong), zero-indexed. The former is of type ``order\_leftmost", while the latter is of type "describe\_distance." For question 19 illustrated in ~\cref{fig:qa_19}, the participants report--after questionnaire submission--confusion on whether the answer should be deduced using pixel-position ordering of the labels or the world-position ordering of objects. We speculate this confusion leads to the participants' overwhelming mistakes on this question. In addition, since this question involves objects very distant from the ego vehicle, the question is challenging due to the linear perspective. This might also cause conflicted participants' responses to question 29 shown in ~\cref{fig:qa_29}. Accounting for these factors, we refine the generation process for the final version of MetaVQA Dataset by choosing clearer phrasing and enforcing better visibility constraints on objects (for example, increasing the minimum observable pixels). Despite these issues, novice participants still report high test accuracies, and we conclude that the MetaVQA Dataset is intuitive to answer and clear in answering guideline. Therefore, we argue that the MetaVQA Dataset is suitable for zero-shot plug-in-and-play evaluating the embodied scene understanding entertained by general-purpose vision langauge models.

\subsection{Effect of Set-of-Marks Prompting Scheme}
\label{sec:som}
The Set-of-Mark ~\cite{yang2023setofmark} paper proposes numerous prompting schemes, from using instance-segmentation masks to bounding boxes. In addition, the text size and background colors are also varied. 
We perform a grid search with observation generated using different prompting schemes while keeping the base images and object-to-label mapping identical across sets, and we use Qwen2~\cite{Qwen2VL}--the VLM with the best grounding capability as discussed in the main paper.
% --as the arbiter to select the optimal prompting scheme.
Referring to~\cref{tab:labelvariant}, Qwen2 achieves the best overall and grounding performance on images annotated with bounding boxes with labels of text size 1.25 and black for background colors. In addition, we observe that text size seems to have a trivial impact on the final performance. Based on these observations, we fixed the annotation style of MetaVQA with bounding-box annotations, black text background color, and a text size of 1.00 (to reduce label occlusions).
\begin{table}[!h]
 \centering
\begin{scriptsize}
 \begin{tabular}{@{}cccccc@{}}
   \toprule
   Text Size & Form     & Background & Overall & Grounding \\
   \midrule
   0.75      & box      & white      & 0.440   & 0.867     \\
   0.75      & box      & black      & 0.457   & \textbf{0.933}     \\
   0.75      & mask     & white      & 0.422   & 0.467     \\
   0.75      & mask     & black      & 0.420   & 0.533     \\
   0.75      & contour  & white      & 0.430   & 0.467     \\
   0.75      & contour  & black      & 0.420   & 0.733     \\
   1.25      & box      & white      & 0.437   & 0.800     \\
   1.25      & box      & black      & \textbf{0.472}   & \textbf{0.933}     \\
   1.25      & mask     & white      & 0.440   & 0.333     \\
   1.25      & mask     & black      & 0.437   & 0.333     \\
   1.25      & contour  & white      & 0.437   & 0.400     \\
   1.25      & contour  & black      & 0.422   & 0.600     \\
   \bottomrule
 \end{tabular}
 \end{scriptsize}
  \caption{\textbf{Effect of Set-of-Marks Annotations.} We tested different annotation styles, text sizes, and background colors while fixing the model (Qwen2) and the numerical labeling, base images, and grounding questions. 
  %\bz{more like ablation study, put it in appendix.}
 }
 \label{tab:labelvariant}
 \vspace{-10pt}
\end{table}

\section{Benchmark Results}
\subsection{Definitions}
We present the naming conventions used in this work in this subsection.
\begin{table}[!h]
\centering
\resizebox{1.0\linewidth}{!}{
\begin{scriptsize}
  \begin{tabular}{ll}
    \toprule
    Abbreviation                  & Checkpoint  \\
    \midrule
    \textbf{LLaVA}-NeXT        & llava-1.6-vicuna-7b~\cite{liu2023improved}          \\ 
    \textbf{LLaVA}-OneVision    & llava-onevision-7b-ov ~\cite{li2024llava}          \\
    \textbf{GPT}-4o     & GPT-4o ~\cite{OpenAI_ChatGPT4_2024}         \\
    \textbf{Qwen2}      & qwen2-vl-7b-instruct ~\cite{Qwen2VL}          \\ %\textbf{qwen2}-vl-7b-instruct
    \textbf{Llama3.2}   & llama-3.2-11B-Vision-Instruct ~\cite{dubey2024llama3herdmodels}         \\ 
    \textbf{InternVL2}-4B   & InternVL2-4B ~\cite{chen2024internvl} \\
    \textbf{InternVL2}-8B   & InternVL2-8B ~\cite{chen2024internvl}          \\
    \bottomrule
  \end{tabular}
    \end{scriptsize}
    }
    \caption{\textbf{Model Abbreviations}. These mappings are used consistently throughout the main paper and the supplementary materials.}
    \vspace{-1.5em}
    \label{tab:abbreviations}
\end{table}

\subsection{Visual Question Answering}
We benchmark the performance of various baselines ~\cite{liu2023improved, li2024llava, Qwen2VL, OpenAI_ChatGPT4_2024,dubey2024llama3herdmodels, chen2023internvl} on the withheld test set (``overall") mentioned in the main paper. Furthermore, We provide detailed performances of baselines on (1) test questions with simulated observations (``sim" split) (2) test questions with real observations (``real" split). To save spaces, we used abbreviations illustrated in ~\cref{tab:abbreviations}
for baselines in these benchmark tables. 

\subsubsection{Response Parsing}
\label{sec:parser}
We establish a unified parsing standard using regular expression (regex) matching for the token sequences generated by all VLMs. If only a singular token is generated, we use this character as the option. If this is not the case, we search for option keywords provided in the multiple-choice questions. In cases of multiple matches, We select the last matched string as the model's output upon empirical examinations of the VLMs' raw outputs. If still no match, the parser will look for single characters enclosed by parentheses. If all searches returns ill-composed results (empty match or illegal character), we consider the parsing to be a failed case. In the closed-loop evaluations, if a parse failure happens, a randomized action is taken. Code implementation will be available in the Github repository.

\subsubsection{Benchmarks on Test Set}
~\cref{tab:overall-spatial} presents the performance in ``spatial reasoning" of the baseline VLMs on the withheld test set. ~\cref{tab:overall-embodied} presents the performance in ``embodied understanding" on the withheld test set. ~\cref{tab:Grounding}
presents the grounding performance of baseline VLMs, categorized according to the test set compositions.

\subsubsection{Benchmarks on Real Test Split}
~\cref{tab:real-spatial} presents the performance in ``spatial reasoning" of the baseline VLMs on the ``real" split of withheld test set. ~\cref{tab:real-embodied} presents the performance in ``embodied understanding" on the ``real" split.

\subsubsection{Benchmarks on Simulated Test Split}
~\cref{tab:sim-spatial} presents the performance in ``spatial reasoning" of the baseline VLMs on the ``sim" split of withheld test set. ~\cref{tab:sim-embodied} presents the performance in ``embodied understanding" on the ``sim" split.

%\subsubsection{Learned Questions after Fine-tuning}
%~\cref{fig:learned} showcases a few test questions to which the fine-tuned InternVL2-8B~\cite{chen2024internvl} model learns to answer correctly. The fine-tuned VLM enjoys improved spatial reasoning and embodied understanding. The numbers in parenthesis are the question IDs in the test set.

%overall-spatial
\begin{table*}[!htbp] %
\centering % Center the table
\scriptsize % Use smaller font size for better fit
\setlength{\tabcolsep}{2pt} % Adjust column spacing
\renewcommand{\arraystretch}{1.0} % Adjust row spacing
\resizebox{1.0\linewidth}{!}{
\begin{tabular}{lcccccccccccccccccccccccc}
\toprule
& \multicolumn{24}{c}{Spatial Questions (Overall)} \\
\cmidrule(r){2-25}
Model & \rotatebox{90}{\makecell[l]{Overall}} & \rotatebox{90}{\makecell[l]{relative\_distance}} & \rotatebox{90}{\makecell[l]{order\_rightmost}} & \rotatebox{90}{\makecell[l]{describe\_distance}} & \rotatebox{90}{\makecell[l]{identify\_closest}} & \rotatebox{90}{\makecell[l]{relative\_predict\\\_crash\_still}} & \rotatebox{90}{\makecell[l]{order\_closest}} & \rotatebox{90}{\makecell[l]{identify\_heading}} & \rotatebox{90}{\makecell[l]{identify\_rightmost}} & \rotatebox{90}{\makecell[l]{relative\_heading}} & \rotatebox{90}{\makecell[l]{relative\_predict\\\_crash\_dynamic}} & \rotatebox{90}{\makecell[l]{identify\_distance}} & \rotatebox{90}{\makecell[l]{order\_leftmost}} & \rotatebox{90}{\makecell[l]{identify\_type}} & \rotatebox{90}{\makecell[l]{order\_backmost}} & \rotatebox{90}{\makecell[l]{identify\_backmost}} & \rotatebox{90}{\makecell[l]{order\_frontmost}} & \rotatebox{90}{\makecell[l]{relative\_position}} & \rotatebox{90}{\makecell[l]{describe\_sector}} & \rotatebox{90}{\makecell[l]{identify\_frontmost}} & \rotatebox{90}{\makecell[l]{pick\_closer}} & \rotatebox{90}{\makecell[l]{identify\_position}} & \rotatebox{90}{\makecell[l]{identify\_leftmost}} & \rotatebox{90}{\makecell[l]{identify\_color}} \\
\midrule
random & 0.287 & 0.267 & 0.218 & 0.245 & 0.254 & 0.510 & 0.240 & 0.308 & 0.241 & 0.535 & 0.467 & 0.289 & 0.215 & 0.250 & 0.250 & 0.234 & 0.213 & 0.272 & 0.265 & 0.253 & 0.281 & 0.221 & 0.246 & 0.315 \\
LLaVA-NeXT ~\cite{liu2023improved} & 0.190 & 0.183 & 0.239 & 0.149 & 0.060 & 0.206 & 0.201 & 0.147 & 0.139 & 0.450 & 0.467 & 0.054 & 0.267 & 0.000 & 0.297 & 0.092 & 0.312 & 0.203 & 0.272 & 0.049 & 0.127 & 0.295 & 0.113 & 0.000 \\
LLaVA-OneVision ~\cite{li2024llava} & 0.422 & 0.233 & 0.401 & 0.226 & 0.590 & 0.779 & 0.338 & 0.171 & 0.646 & 0.599 & 0.948 & 0.126 & 0.600 & 0.674 & 0.324 & 0.454 & 0.255 & 0.319 & 0.398 & 0.444 & 0.382 & 0.326 & 0.669 & 0.460 \\
GPT-4o ~\cite{OpenAI_ChatGPT4_2024} & 0.489 & 0.254 & 0.585 & 0.234 & 0.575 & 0.740 & 0.403 & 0.360 & 0.639 & 0.609 & 0.887 & 0.329 & 0.541 & 0.663 & 0.385 & 0.560 & 0.355 & 0.440 & 0.626 & 0.340 & 0.342 & 0.593 & 0.599 & 0.516 \\
\midrule
Qwen2 ~\cite{Qwen2VL} & 0.411 & 0.221 & 0.408 & 0.381 & 0.687 & 0.186 & 0.390 & 0.199 & 0.658 & 0.530 & 0.396 & 0.220 & 0.644 & 0.808 & 0.385 & 0.539 & 0.340 & 0.319 & 0.386 & 0.173 & 0.351 & 0.474 & 0.669 & 0.492 \\
Qwen2-finetuned & 0.740 & 0.550 & \textbf{0.761} & 0.891 & 0.843 & 0.627 & \textbf{0.766} & 0.404 & 0.899 & 0.540 & 0.821 & 0.487 & \textbf{0.904} & \textbf{0.902} & \textbf{0.770} & \textbf{0.922} & \textbf{0.801} & 0.431 & \textbf{0.985} & 0.759 & 0.325 & \textbf{0.979} & \textbf{0.873} & 0.831 \\
\midrule
Llama3.2 ~\cite{dubey2024llama3herdmodels} & 0.442 & 0.308 & 0.577 & 0.207 & 0.507 & 0.446 & 0.351 & 0.226 & 0.633 & 0.490 & 0.915 & 0.484 & 0.511 & 0.699 & 0.365 & 0.461 & 0.355 & 0.310 & 0.519 & 0.302 & 0.382 & 0.330 & 0.570 & 0.677 \\
Llama3.2-finetuned & 0.610 & \textbf{0.667} & 0.465 & 0.544 & 0.821 & \textbf{0.873} & 0.630 & 0.435 & \textbf{0.905} & \textbf{0.688} & \textbf{0.953} & 0.787 & 0.126 & 0.804 & 0.514 & 0.858 & 0.099 & 0.207 & 0.658 & \textbf{0.772} & 0.215 & 0.512 & 0.817 & 0.758 \\
\midrule
InternVL2-8B ~\cite{chen2023internvl} & 0.476 & 0.421 & 0.317 & 0.241 & 0.664 & 0.858 & 0.370 & 0.363 & 0.601 & 0.569 & \textbf{0.953} & 0.415 & 0.504 & 0.652 & 0.372 & 0.482 & 0.227 & 0.349 & 0.568 & 0.364 & 0.338 & 0.418 & 0.648 & 0.492 \\
InternVL2-8B-finetuned & \textbf{0.813} & 0.600 & 0.669 & \textbf{0.904} & \textbf{0.866} & 0.868 & 0.734 & \textbf{0.647} & 0.804 & 0.678 & \textbf{0.953} & \textbf{0.834} & 0.741 & 0.899 & 0.696 & 0.865 & 0.745 & \textbf{0.776} & 0.927 & 0.759 & \textbf{0.794} & 0.940 & 0.824 & \textbf{0.863} \\
\bottomrule
\end{tabular}
}
\vspace{-10pt}
\caption{\textbf{VQA benchmarks (Overall-Spatial)}. Per-question accuracies are evaluated on the withheld test set.}
\vspace{-10pt}
\label{tab:overall-spatial}
\end{table*}

%real-spatial
\begin{table*}[!htbp]
\centering % Center the table
\scriptsize % Use smaller font size for better fit
\setlength{\tabcolsep}{2pt} % Adjust column spacing
\renewcommand{\arraystretch}{1.0} % Adjust row spacing
\resizebox{1.0\linewidth}{!}{
\begin{tabular}{lccccccccccccccccccccccc}
\toprule
& \multicolumn{23}{c}{Spatial Questions (Real)} \\
\cmidrule(r){2-24}
Model & \rotatebox{90}{\makecell[l]{Overall}} & \rotatebox{90}{\makecell[l]{relative\_distance}} & \rotatebox{90}{\makecell[l]{order\_rightmost}} & \rotatebox{90}{\makecell[l]{describe\_distance}} & \rotatebox{90}{\makecell[l]{identify\_closest}} & \rotatebox{90}{\makecell[l]{relative\_predict\\\_crash\_still}} & \rotatebox{90}{\makecell[l]{order\_closest}} & \rotatebox{90}{\makecell[l]{identify\_heading}} & \rotatebox{90}{\makecell[l]{identify\_rightmost}} & \rotatebox{90}{\makecell[l]{relative\_heading}} & \rotatebox{90}{\makecell[l]{relative\_predict\\\_crash\_dynamic}} & \rotatebox{90}{\makecell[l]{identify\_distance}} & \rotatebox{90}{\makecell[l]{order\_leftmost}} & \rotatebox{90}{\makecell[l]{identify\_type}} & \rotatebox{90}{\makecell[l]{order\_backmost}} & \rotatebox{90}{\makecell[l]{identify\_backmost}} & \rotatebox{90}{\makecell[l]{order\_frontmost}} & \rotatebox{90}{\makecell[l]{relative\_position}} & \rotatebox{90}{\makecell[l]{describe\_sector}} & \rotatebox{90}{\makecell[l]{identify\_frontmost}} & \rotatebox{90}{\makecell[l]{pick\_closer}} & \rotatebox{90}{\makecell[l]{identify\_position}} & \rotatebox{90}{\makecell[l]{identify\_leftmost}} \\
\midrule
random & 0.296 & 0.237 & 0.188 & 0.250 & 0.203 & 0.500 & 0.362 & 0.317 & 0.194 & 0.602 & 0.490 & 0.279 & 0.210 & 0.280 & 0.246 & 0.236 & 0.185 & 0.287 & 0.256 & 0.310 & 0.286 & 0.245 & 0.244 \\
LLaVA-NeXT & 0.187 & 0.211 & 0.219 & 0.167 & 0.017 & 0.209 & 0.241 & 0.106 & 0.129 & 0.470 & 0.529 & 0.024 & 0.290 & 0,000 & 0.279 & 0.036 & 0.370 & 0.218 & 0.250 & 0.069 & 0.114 & 0.224 & 0.178 \\
LLaVA-OneVision & 0.452 & 0.228 & 0.406 & 0.294 & 0.542 & 0.764 & 0.466 & 0.174 & 0.694 & 0.614 & \textbf{0.941} & 0.091 & 0.629 & 0.826 & 0.246 & 0.364 & 0.185 & 0.366 & 0.494 & 0.362 & 0.381 & 0.476 & 0.778 \\
GPT-4o & 0.509 & 0.246 & 0.703 & 0.256 & 0.508 & 0.836 & 0.328 & 0.323 & 0.677 & 0.566 & 0.873 & 0.321 & 0.645 & 0.770 & 0.377 & 0.545 & 0.315 & 0.396 & 0.685 & 0.276 & 0.314 & 0.633 & 0.733 \\
\midrule
Qwen2 & 0.405 & 0.158 & 0.469 & 0.439 & 0.610 & 0.218 & 0.431 & 0.518 & 0.710 & 0.518 & 0.373 & 0.176 & 0.710 & 0.907 & 0.361 & 0.527 & 0.278 & 0.366 & 0.315 & 0.155 & 0.286 & 0.531 & 0.667 \\
%qwen2
Qwen2-finetuned & 0.723 & 0.649 & \textbf{0.781} & \textbf{0.894} & 0.780 & 0.591 & 0.759 & 0.342 & \textbf{0.871} & 0.651 & 0.745 & 0.515 & \textbf{0.919} & \textbf{0.963} & 0.770 & 0.891 & \textbf{0.833} & 0.406 & \textbf{0.976} & 0.621 & 0.295 & \textbf{0.986} & 0.844 \\
\midrule
%qwen2
Llama3.2 & 0.464 & 0.368 & 0.594 & 0.189 & 0.475 & 0.436 & 0.362 & 0.211 & 0.629 & 0.578 & 0.931 & 0.539 & 0.548 & 0.832 & 0.410 & 0.400 & 0.296 & 0.317 & 0.583 & 0.207 & 0.352 & 0.374 & 0.622 \\
Llama3.2-finetuned & 0.627 & \textbf{0.728} & 0.578 & 0.522 & 0.729 & 0.864 & 0.500 & 0.422 & \textbf{0.871} & 0.590 & \textbf{0.941} & \textbf{0.824} & 0.048 & 0.938 & 0.475 & 0.818 & 0.019 & 0.208 & 0.667 & \textbf{0.810} & 0.219 & 0.735 & 0.867 \\
\midrule
InternVL2-8B & 0.516 & 0.283 & 0.328 & 0.283 & 0.644 & \textbf{0.873} & 0.379 & 0.675 & 0.694 & 0.675 & \textbf{0.941} & 0.424 & 0.548 & 0.795 & 0.410 & 0.400 & 0.278 & 0.376 & 0.708 & 0.259 & 0.305 & 0.503 & 0.711 \\
InternVL2-8B-finetuned & \textbf{0.838} & 0.640 & 0.672 & 0.878 & \textbf{0.847} & 0.864 & \textbf{0.793} & \textbf{0.696} & 0.790 & \textbf{0.735} & \textbf{0.941} & 0.640 & 0.823 & 0.950 & \textbf{0.803} & \textbf{0.909} & 0.759 & \textbf{0.822} & 0.958 & \textbf{0.810} & \textbf{0.781} & 0.952 & \textbf{0.911} \\
\bottomrule
\end{tabular}
}
\vspace{-10pt}
\caption{\textbf{VQA benchmarks (Real-Spatial)}. Per-question accuracies are evaluated on the ``real" split of the withheld test set.}
\vspace{-10pt}
\label{tab:real-spatial}
\end{table*}

%sim-spatial
\begin{table*}[!htbp]
\centering % Center the table
\scriptsize % Use smaller font size for better fit
\setlength{\tabcolsep}{2pt} % Adjust column spacing
\renewcommand{\arraystretch}{1.0} % Adjust row spacing
\resizebox{1.0\linewidth}{!}{
\begin{tabular}{lcccccccccccccccccccccccc}
\toprule
& \multicolumn{24}{c}{Spatial Questions (Sim)} \\
\cmidrule(r){2-25}
Model & \rotatebox{90}{\makecell[l]{Overall}} & \rotatebox{90}{\makecell[l]{relative\_distance}} & \rotatebox{90}{\makecell[l]{order\_rightmost}} & \rotatebox{90}{\makecell[l]{describe\_distance}} & \rotatebox{90}{\makecell[l]{identify\_closest}} & \rotatebox{90}{\makecell[l]{relative\_predict\\\_crash\_still}} & \rotatebox{90}{\makecell[l]{order\_closest}} & \rotatebox{90}{\makecell[l]{identify\_heading}} & \rotatebox{90}{\makecell[l]{identify\_rightmost}} & \rotatebox{90}{\makecell[l]{relative\_heading}} & \rotatebox{90}{\makecell[l]{relative\_predict\\\_crash\_dynamic}} & \rotatebox{90}{\makecell[l]{identify\_distance}} & \rotatebox{90}{\makecell[l]{order\_leftmost}} & \rotatebox{90}{\makecell[l]{identify\_type}} & \rotatebox{90}{\makecell[l]{order\_backmost}} & \rotatebox{90}{\makecell[l]{identify\_backmost}} & \rotatebox{90}{\makecell[l]{order\_frontmost}} & \rotatebox{90}{\makecell[l]{relative\_position}} & \rotatebox{90}{\makecell[l]{describe\_sector}} & \rotatebox{90}{\makecell[l]{identify\_frontmost}} & \rotatebox{90}{\makecell[l]{pick\_closer}} & \rotatebox{90}{\makecell[l]{identify\_position}} & \rotatebox{90}{\makecell[l]{identify\_leftmost}} & \rotatebox{90}{\makecell[l]{identify\_color}} \\
\midrule
random & 0.281 & 0.294 & 0.244 & 0.242 & 0.293 & 0.521 & 0.167 & 0.298 & 0.271 & 0.487 & 0.445 & 0.304 & 0.219 & 0.209 & 0.253 & 0.233 & 0.230 & 0.260 & 0.270 & 0.221 & 0.276 & 0.196 & 0.247 & 0.315 \\
LLaVA-NeXT & 0.192 & 0.159 & 0.256 & 0.138 & 0.093 & 0.202 & 0.177 & 0.198 & 0.146 & 0.437 & 0.409 & 0.098 & 0.247 & 0.000 & 0.310 & 0.128 & 0.276 & 0.191 & 0.287 & 0.038 & 0.138 & 0.370 & 0.082 & 0.000 \\
LLaVA-OneVision & 0.398 & 0.238 & 0.397 & 0.185 & 0.627 & 0.798 & 0.260 & 0.168 & 0.615 & 0.588 & 0.955 & 0.179 & 0.575 & 0.461 & 0.379 & 0.512 & 0.299 & 0.282 & 0.332 & 0.490 & 0.382 & 0.167 & 0.619 & 0.460 \\
GPT-4o & 0.474 & 0.262 & 0.487 & 0.221 & 0.627 & 0.628 & 0.448 & 0.405 & 0.615 & 0.639 & 0.900 & 0.339 & 0.452 & 0.513 & 0.391 & 0.570 & 0.379 & 0.473 & 0.586 & 0.375 & 0.366 & 0.551 & 0.536 & 0.516 \\
\midrule
Qwen2 & 0.415 & 0.278 & 0.359 & 0.346 & 0.747 & 0.149 & 0.365 & 0.282 & 0.625 & 0.538 & 0.418 & 0.286 & 0.589 & 0.670 & 0.402 & 0.547 & 0.379 & 0.282 & 0.434 & 0.183 & 0.407 & 0.413 & 0.670 & 0.492 \\
Qwen2-finetuned & 0.754 & 0.460 & \textbf{0.744} & 0.889 & \textbf{0.893} & 0.670 & \textbf{0.771} & 0.481 & 0.917 & 0.462 & 0.891 & 0.446 & \textbf{0.890} & 0.817 & \textbf{0.770} & \textbf{0.942} & \textbf{0.782} & 0.450 & \textbf{0.992} & \textbf{0.837} & 0.350 & \textbf{0.971} & \textbf{0.887} & 0.831 \\
\midrule
Llama3.2 & 0.424 & 0.254 & 0.564 & 0.218 & 0.533 & 0.457 & 0.344 & 0.244 & 0.635 & 0.429 & 0.900 & 0.402 & 0.479 & 0.513 & 0.333 & 0.500 & 0.391 & 0.305 & 0.475 & 0.356 & 0.407 & 0.283 & 0.546 & 0.677 \\
Llama3.2-finetuned & 0.596 & \textbf{0.611} & 0.372 & 0.557 & \textbf{0.893} & \textbf{0.883} & 0.708 & 0.450 & \textbf{0.927} & \textbf{0.756} & \textbf{0.964} & 0.732 & 0.192 & 0.617 & 0.540 & 0.884 & 0.149 & 0.206 & 0.652 & 0.750 & 0.211 & 0.275 & 0.794 & 0.758 \\
\midrule
InternVL2-8B & 0.444 & 0.413 & 0.308 & 0.215 & 0.680 & 0.840 & 0.365 & 0.450 & 0.542 & 0.496 & \textbf{0.964} & 0.402 & 0.466 & 0.452 & 0.345 & 0.535 & 0.195 & 0.328 & 0.471 & 0.423 & 0.366 & 0.326 & 0.619 & 0.492 \\
InternVL2-8B-finetuned & \textbf{0.793} & 0.563 & 0.667 & \textbf{0.919} & 0.880 & 0.872 & 0.698 & \textbf{0.588} & 0.813 & 0.639 & \textbf{0.964} & \textbf{0.795} & 0.671 & \textbf{0.826} & 0.621 & 0.837 & 0.736 & \textbf{0.740} & 0.906 & 0.731 & \textbf{0.805} & 0.928 & 0.784 & \textbf{0.863} \\
\bottomrule
\end{tabular}
}
\caption{\textbf{VQA benchmarks (Sim-Spatial)}. Per-question accuracies are evaluated on the ``sim" split of the withheld test set.}
\label{tab:sim-spatial}
\end{table*}

\begin{figure*}[!t]
\centering
\begin{minipage}[t]{0.45\textwidth} % First table
    \centering
    \scriptsize % Smaller font size
    \setlength{\tabcolsep}{2pt} % Adjust column spacing
    \renewcommand{\arraystretch}{1.0} % Adjust row spacing
    \begin{tabular}{lcccccc}
        \toprule
        & \multicolumn{6}{c}{Embodied Questions (Overall)} \\
        \cmidrule(r){2-7}
        Model & \rotatebox{90}{\makecell[l]{Overall}} & \rotatebox{90}{\makecell[l]{embodied\_distance}} & \rotatebox{90}{\makecell[l]{embodied\_collision}} & \rotatebox{90}{\makecell[l]{predict\_crash\\ego\_still}} & \rotatebox{90}{\makecell[l]{embodied\_sideness}} & \rotatebox{90}{\makecell[l]{predict\_crash\\ego\_dynamic}} \\
        \midrule
        random & 0.382 & 0.255 & 0.498 & 0.521 & 0.348 & 0.504 \\
        LLaVA-NeXT & 0.419 & 0.159 & 0.489 & 0.303 & 0.652 & 0.384 \\
        LLaVA-OneVision & 0.746 & 0.442 & 0.923 & 0.976 & 0.794 & \textbf{0.961} \\
        GPT-4o & 0.764 & 0.785 & 0.719 & 0.893 & 0.732 & 0.873 \\
        \midrule
        Qwen2 & 0.649 & 0.451 & 0.836 & 0.259 & 0.804 & 0.482 \\
        Qwen2-finetuned & \textbf{0.948} & \textbf{0.998} & 0.879 & 0.817 & \textbf{1.000} & 0.894 \\
        \midrule
        Llama3.2 & 0.536 & 0.332 & 0.650 & 0.517 & 0.574 & 0.849 \\
        Llama3.2-finetuned & 0.944 & 0.962 & 0.846 & \textbf{0.997} & 0.999 & \textbf{0.961} \\
        \midrule
        InternVL2-8B & 0.711 & 0.620 & 0.923 & 0.914 & 0.509 & \textbf{0.961} \\
        InternVL2-8B-finetuned & 0.926 & 0.807 & \textbf{0.953} & \textbf{0.997} & \textbf{1.000} & \textbf{0.961} \\
        \bottomrule
    \end{tabular}
    \captionof{table}{\textbf{VQA benchmarks (Overall-Embodied)}. Per-question accuracies are evaluated on the withheld test set.}
    \label{tab:overall-embodied}
\end{minipage}
\hfill
\begin{minipage}[t]{0.45\textwidth} % Second table
    \centering
    % \scriptsize % Smaller font size
    % \setlength{\tabcolsep}{2pt} % Adjust column spacing
    % \renewcommand{\arraystretch}{1.0} % Adjust row spacing
    \footnotesize % Larger font size for the table
    \setlength{\tabcolsep}{4pt} % Increase column spacing
    \renewcommand{\arraystretch}{1.2} % Increase row spacing

    \begin{tabular}{lccc}
        \toprule
        & \multicolumn{3}{c}{Grounding Questions} \\
        \cmidrule(r){2-4}
        Model & {\makecell[l]{Overall}} & {\makecell[l]{Real}} & {\makecell[l]{Sim}} \\
        \midrule
        random & 0.268 & 0.257 & 0.280 \\
        LLaVA-NeXT & 0.248 & 0.229 & 0.271 \\
        LLaVA-OneVision & 0.728 & 0.827 & 0.615 \\
        GPT4-o & 0.831 & 0.888 & 0.766 \\
        \midrule
        Qwen2 & 0.874 & 0.859 & 0.890 \\
        Qwen2-finetuned & \textbf{0.972} & \textbf{0.992} & \textbf{0.950}\\
        \midrule
        Llama3.2 & 0.790 & 0.855 & 0.716 \\
        Llama3.2-finetuned & 0.923 & 0.944 & 0.899 \\
        \midrule
        InternVL2-8B & 0.702 & 0.783 & 0.610 \\
        InternVL2-8B-finetuned & 0.916 & 0.948 & 0.881 \\
        \bottomrule
    \end{tabular}
    \captionof{table}{\textbf{VQA benchmarks (Grounding)}. Per-question accuracies are evaluated on the withheld whole test set, ``real'' split of the test set, and ``sim'' split of the test set.}
    \label{tab:Grounding}
\end{minipage}
\vspace{-10pt} % Adjust vertical spacing if needed
\end{figure*}

%sim-embodied + real-embodied
\begin{figure*}[!t]
\centering
\begin{minipage}[t]{0.45\textwidth} % Right table
    \vspace{0pt}
    \centering
    \tiny % Reduce font size further
    \setlength{\tabcolsep}{1pt} % Reduce space between columns

    \begin{tabular}{l*{6}{>{\centering\arraybackslash}p{0.8cm}}}
    \toprule
    & \multicolumn{6}{c}{Embodied Questions (Real)} \\
    \cmidrule(r){2-7}
    Model & \rotatebox{90}{Overall} & \rotatebox{90}{Emb\_Dist} & \rotatebox{90}{Emb\_Coll} & \rotatebox{90}{\makecell[l]{PredCrash\\EgoStill}} & \rotatebox{90}{Emb\_Side} & \rotatebox{90}{\makecell[l]{PredCrash\\EgoDyn}} \\
    \midrule
    random & 0.372 & 0.248 & 0.498 & 0.487 & 0.345 & 0.467 \\
    LLaVA-NeXT & 0.414 & 0.189 & 0.445 & 0.342 & 0.647 & 0.327 \\
    LLaVA-OneVision & 0.735 & 0.430 & 0.905 & 0.980 & 0.795 & \textbf{0.980} \\
    GPT-4o & 0.762 & 0.784 & 0.706 & 0.895 & 0.739 & 0.873 \\
    \midrule
    Qwen2 & 0.653 & 0.446 & 0.852 & 0.322 & 0.796 & 0.453 \\
    Qwen2-finetuned & 0.946 & \textbf{0.999} & 0.875 & 0.789 & \textbf{1.000} & 0.887 \\
    \midrule
    Llama3.2 & 0.542 & 0.339 & 0.654 & 0.566 & 0.580 & 0.867 \\
    Llama3.2-finetuned & \textbf{0.947} & 0.969 & 0.844 & \textbf{0.993} & 0.999 & \textbf{0.980} \\
    \midrule
    InternVL2-8B & 0.720 & 0.615 & 0.905 & 0.895 & 0.576 & \textbf{0.980} \\
    InternVL2-8B-finetuned & 0.919 & 0.780 & \textbf{0.956} & \textbf{0.993} & \textbf{1.000} & \textbf{0.980} \\
    \bottomrule
    \end{tabular}
    \captionof{table}{\textbf{VQA benchmarks (Real-Embodied)}. Per-question accuracies are evaluated on the ``real" split of the withheld test set. Question types are shortened for formatting.}
    \label{tab:real-embodied}
\end{minipage}
\hfill
\begin{minipage}[t]{0.45\textwidth} % Left table
    \vspace{0pt}
    \centering
    \tiny % Reduce font size further
    \setlength{\tabcolsep}{1pt} % Reduce space between columns

    \begin{tabular}{l*{6}{>{\centering\arraybackslash}p{0.8cm}}}
    \toprule
    & \multicolumn{6}{c}{Embodied Questions (Sim)} \\
    \cmidrule(r){2-7}
    Model & \rotatebox{90}{Overall} & \rotatebox{90}{Emb\_Dist} & \rotatebox{90}{Emb\_Coll} & \rotatebox{90}{\makecell[l]{PredCrash\\EgoStill}} & \rotatebox{90}{Emb\_Side} & \rotatebox{90}{\makecell[l]{PredCrash\\EgoDyn}} \\
    \midrule
    random & 0.395 & 0.264 & 0.497 & 0.558 & 0.353 & 0.545 \\
    LLaVA-NeXT & 0.426 & 0.12 & 0.542 & 0.261 & 0.660 & 0.448 \\
    LLaVA-OneVision & 0.760 & 0.457 & 0.945 & 0.971 & 0.793 & \textbf{0.940} \\
    GPT-4o & 0.767 & 0.786 & 0.734 & 0.891 & 0.722 & 0.873 \\
    \midrule
    Qwen2 & 0.645 & 0.457 & 0.817 & 0.188 & 0.815 & 0.515 \\
    Qwen2-finetuned & \textbf{0.949} & \textbf{0.998} & 0.883 & 0.848 & \textbf{1.000} & 0.903 \\
    \midrule
    Llama3.2 & 0.527 & 0.321 & 0.644 & 0.464 & 0.565 & 0.828 \\
    Llama3.2-finetuned & 0.939 & 0.952 & 0.847 & \textbf{1.000} & \textbf{1.000} & \textbf{0.940} \\
    \midrule
    InternVL-8B & 0.699 & 0.627 & 0.945 & 0.935 & 0.418 & \textbf{0.940} \\
    InternVL-8B-finetuned & 0.936 & 0.843 & \textbf{0.949} & \textbf{1.000} & \textbf{1.000} & \textbf{0.940} \\
    \bottomrule
    \end{tabular}
    \captionof{table}{\textbf{VQA benchmarks (Sim-Embodied)}. Per-question accuracies are evaluated on the ``sim" split of the withheld test set. Question types shortened for formatting.}
    \label{tab:sim-embodied}
\end{minipage}
\vspace{-10pt} % Adjust vertical spacing if needed
\end{figure*}

\begin{figure*}[!t]
    \centering
    \includegraphics[width=\textwidth]{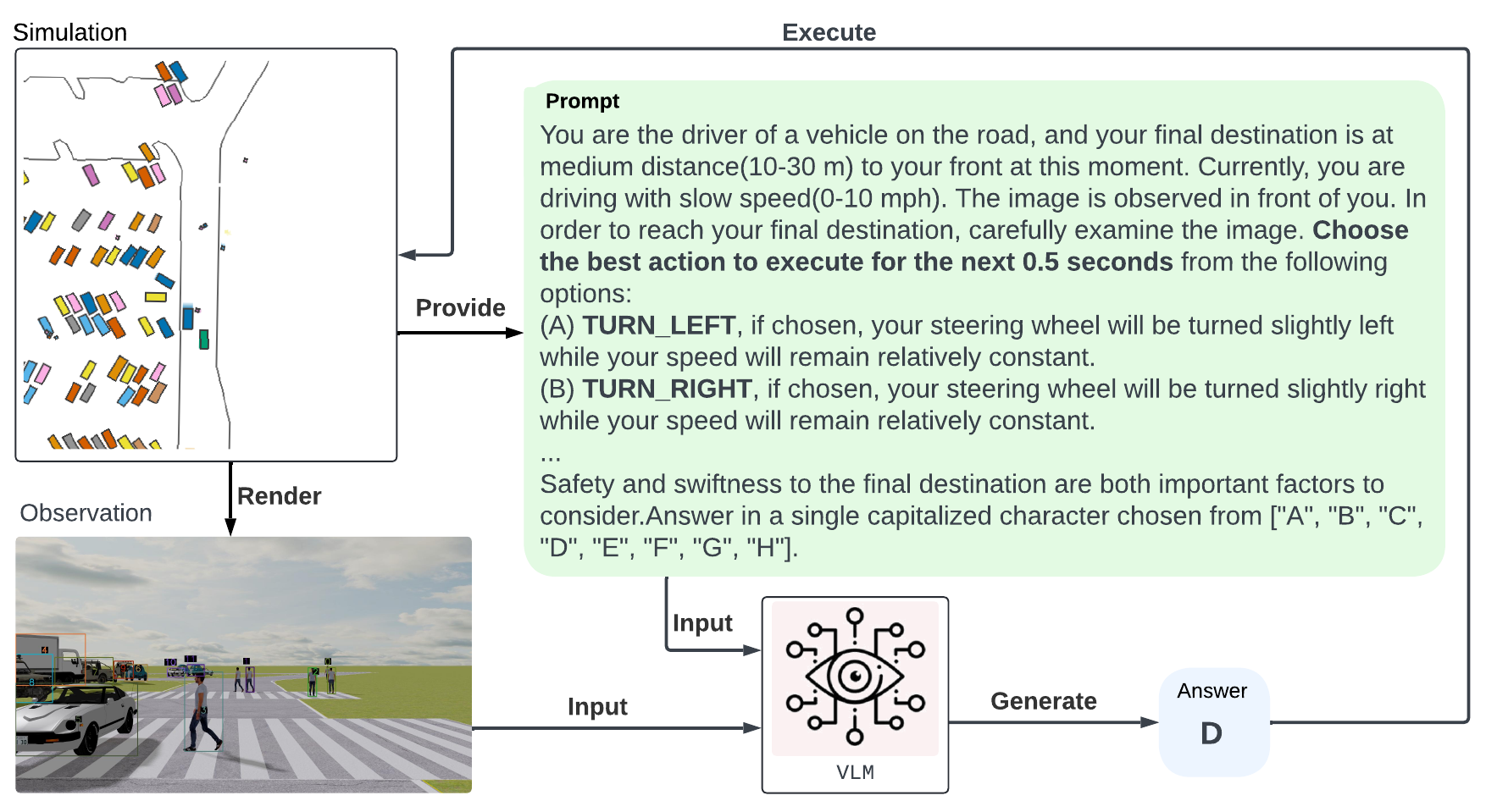}
    \caption{
\textbf{Closed-loop evaluation paradigm.}
%\pzh{Is this figure repeated?}
}
    \label{fig:closed_loop}
\end{figure*}

\subsection{Closed-loop Evaluation}
\subsubsection{Task Formulation}
\paragraph{Interaction Paradigm.} We use the MetaDrive ~\cite{li2022metadrive} simulator, which provides accurate vehicle dynamics simulation for closed-loop evaluations. VLMs are deployed as driving agents in 
imported scenarios using ~\cite{li2023scenarionet}. At every five simulation steps (0.5 seconds wall time), the tested VLM is provided with (1) a Set-of-Mark annotated observation captured from the ego's front camera in $1600 \times 900$ resolution and (2) a driving prompt containing current navigation command and allowed discrete action space. The model will analyze the combined input and select the best action from available options. The chosen action will be fed into the simulation, and it will be repeated for the next 0.5 seconds (5 steps in simulation time) until the next inference step. \cref{fig:closed_loop} illustrates this process. The simulations terminate when their time horizons are reached or when the ego vehicle wanders off drivable regions. 

Very rarely, the tested VLM will generate an invalid response according to the parser mentioned in ~\cref{sec:parser}. In this situation, we fix the chosen action as ``KEEP\_STRAIGHT" such that the speed and the heading of the ego vehicle will remain roughly identical. 

\paragraph{Navigation command.}
At each inference step, the navigation command is recomputed to adjust for the current position of the ego vehicle. The command follows the following form:

\texttt{your final destination is at <distance> to <position> at this moment.}

Here, the \texttt{<distance>} and \texttt{<position>} parameters will be replaced with concrete values chosen from the discrete vocabulary for spatial information mentioned in ~\cref{sec:qa_generation}.  

\paragraph{Action space.}
The actions in the driving prompts are statically mapped to low-level control signals to MetaDrive. MetaDrive receives normalized action as input to control the ego vehicle: $\mathbf a = [a_1, a_2]^T \in [-1, 1]^2$. At each simulation time step, MetaDrive converts the normalized action into the steering \(u_s\) (degree), acceleration \(u_a\) (hp) and brake signal \(u_b\) (hp) in the following ways: (i) $u_s = S_{max} a_1 $, (ii) $ u_a = F_{max} \max(0, a_2)$, (iii) $u_b = -B_{max} \min(0, a_2)$, wherein $S_{max}$ (degree) is the maximal steering angle, $F_{max}$ (hp) is the maximal engine force, and $B_{max}$ (hp) is the maximal brake force. For fair and replicable experiments, we use identical vehicle configurations(for example,  maximum engine force) across different trials.

We conducted grid searches to fix the suitable set of actions. For each candidate, we reconstruct real-world driving trajectories as action sequences with only allowed action provided by the candidate. These sequences are computed greedily (and repeated) at every five simulation steps, following the same inference frequency as the closed-loop evaluation. The optimal action at a particular step is decided according to the resulting deviation from the original trajectories if the action is executed. This sequence-building is autoregressive, meaning that previous optimal actions(and their generated trajectories) affect the decision on later optimal actions. We fix the current action space as it leads to the best reconstruction quality.

\paragraph{Test scenarios.}
We tailor 120 diverse scenarios to evaluate VLMs' embodied scene understanding holistically. These scenarios include 60 from the nuScenes dataset and the other 60 selected from a corpus of safety-critical situations generated using CAT ~\cite{zhang2023cat}. For each of the 60 safety-critical scenarios, an adversarial agent will attempt to run into the ego vehicle, and we ensure the observability of adversarial agents.

\subsubsection{Metrics}
\paragraph{Route Completion}
The ratio of the traveled distance against the length of the complete route averaged across scenarios.

\paragraph{Collision Rate}
The ratio of scenarios where the ego vehicle collides with any other object.

\paragraph{Off-Road Rate}
The ratio of scenarios where the ego vehicle leaves drivable regions. 

\paragraph{Final Displacement Error (FDE)}
The L2 distance between the final position of the ego vehicle from the final destination averaged across scenarios. 

\paragraph{Average Displacement Error (ADE)}
The mean per-step L2 distance between the ground-truth trajectories and the VLM-driven trajectories averaged across scenarios. If a simulation terminates prematurely(due to VLMs driving off-road), the last ego vehicle position is appended to align the length of the ground-truth trajectory and the VLM-driven trajectory.

\clearpage
\newpage
{
\captionsetup{type=table}
\begin{tcolorbox}[colback=white!10,%gray background
                  colframe=black,% black frame color
                  width=\textwidth,
                  arc=1mm, auto outer arc,
                  boxrule=0.5pt,
                 ]
\label{fig:types}

\texttt{\textcolor{blue}{\textbf{Embodied Questions:}}}

\texttt{\textbf{embodied\_distance}.} This question examines how far the ego will move from the current position, assuming that \texttt{<action>} is executed over the next \texttt{<duration>} period and the ego's current speed is \texttt{<speed>}.

\begin{minipage}[b]{\textwidth}
            \centering
            \includegraphics[width=\textwidth]{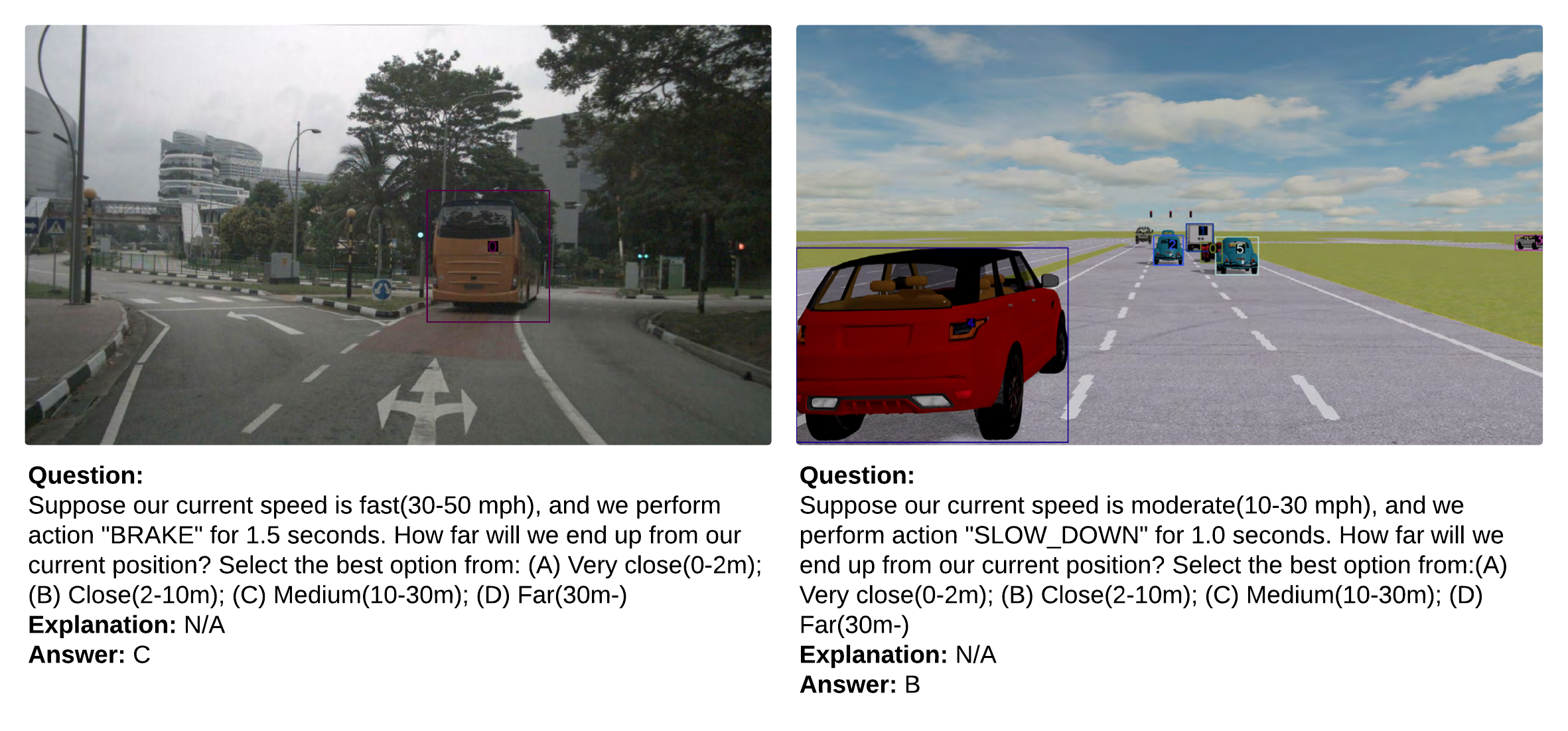}
\end{minipage}
\texttt{\textbf{embodied\_sideness}.} This question examines how whether the ego will move to its left or its right(in the current frame), assuming that \texttt{<action>} is executed over the next\texttt{<duration>} period and the ego's current speed is \texttt{<speed>}.

\begin{minipage}[b]{\textwidth}
            \centering
            \includegraphics[width=\textwidth]{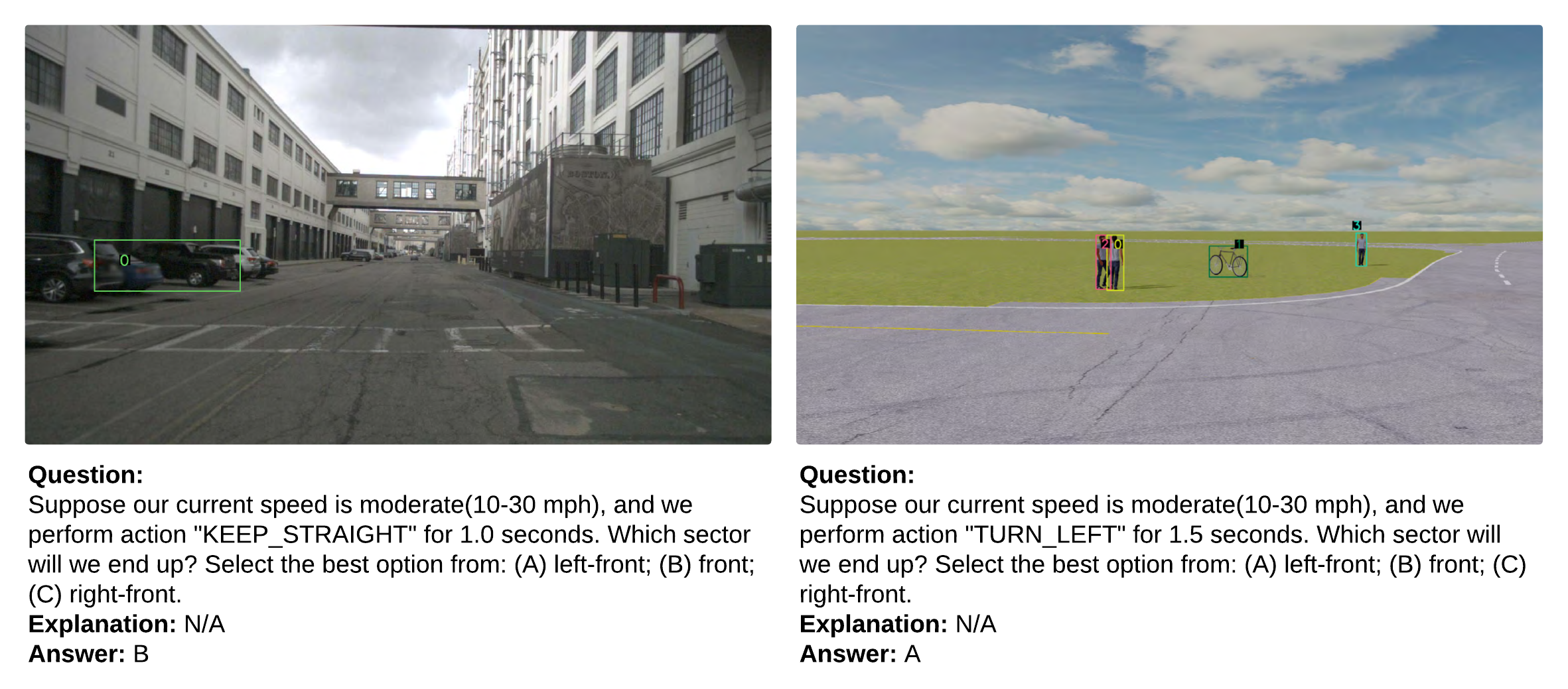}
\end{minipage}
\end{tcolorbox}
}

\clearpage
\newpage
{
\captionsetup{type=table}
\begin{tcolorbox}[colback=white!10,%gray background
                  colframe=black,% black frame color
                  width=\textwidth,
                  arc=1mm, auto outer arc,
                  boxrule=0.5pt,
                 ]
\texttt{\textcolor{blue}{\textbf{Embodied Questions:}}}

\texttt{\textbf{embodied\_collision}.} This question examines whether the ego will collide into selected object \texttt{<id1>}, assuming that \texttt{<action>} is executed over the next \texttt{<duration>} period and the ego's current speed is \texttt{<speed>}.

\begin{minipage}[b]{\textwidth}
            \centering
            \includegraphics[width=\textwidth]{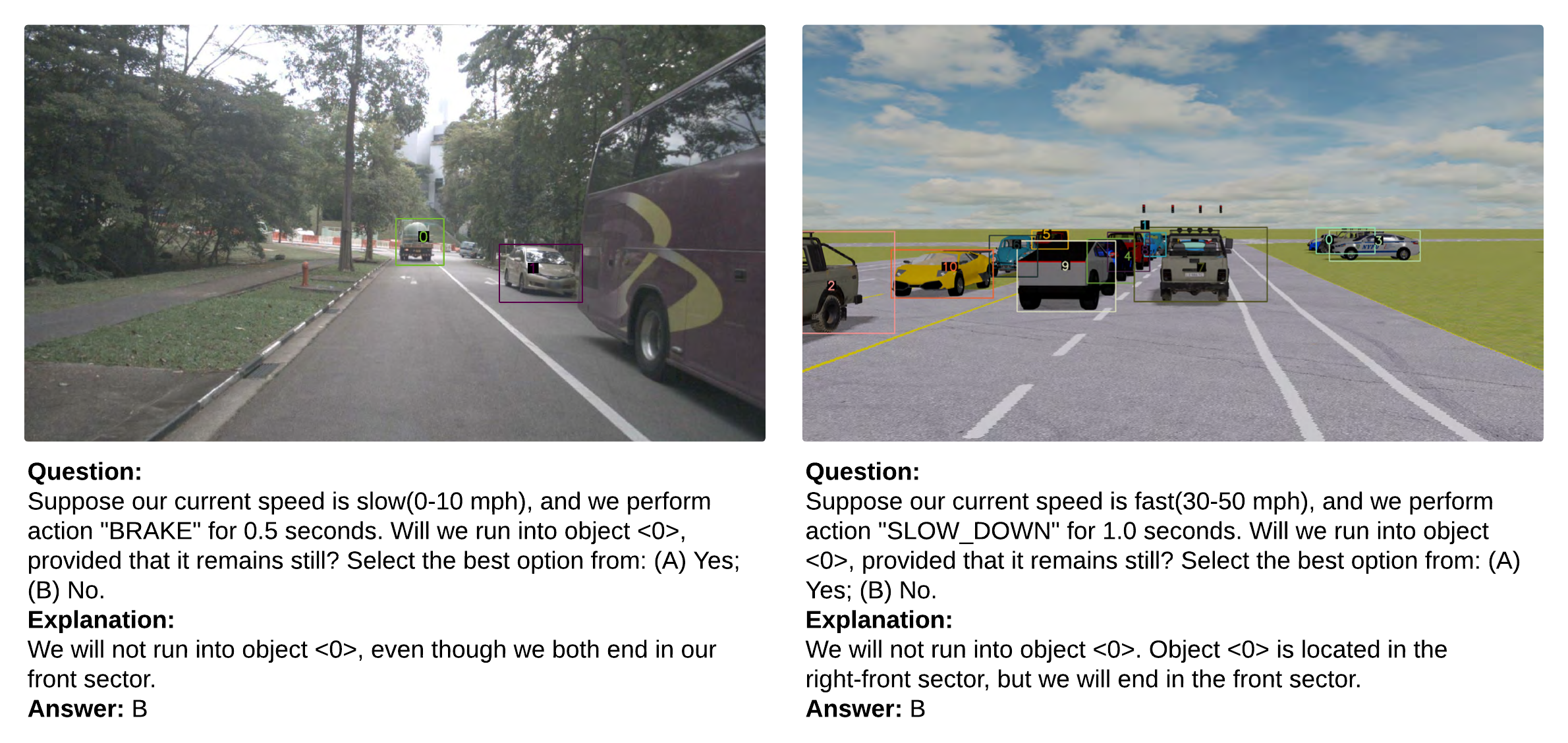}
\end{minipage}
\texttt{\textbf{predict\_crash\_ego\_*}.} This family of questions examines how whether the selected object \texttt{<id1>} will collide with the ego under various conditions.

\begin{minipage}[b]{\textwidth}
            \centering
            \includegraphics[width=\textwidth]{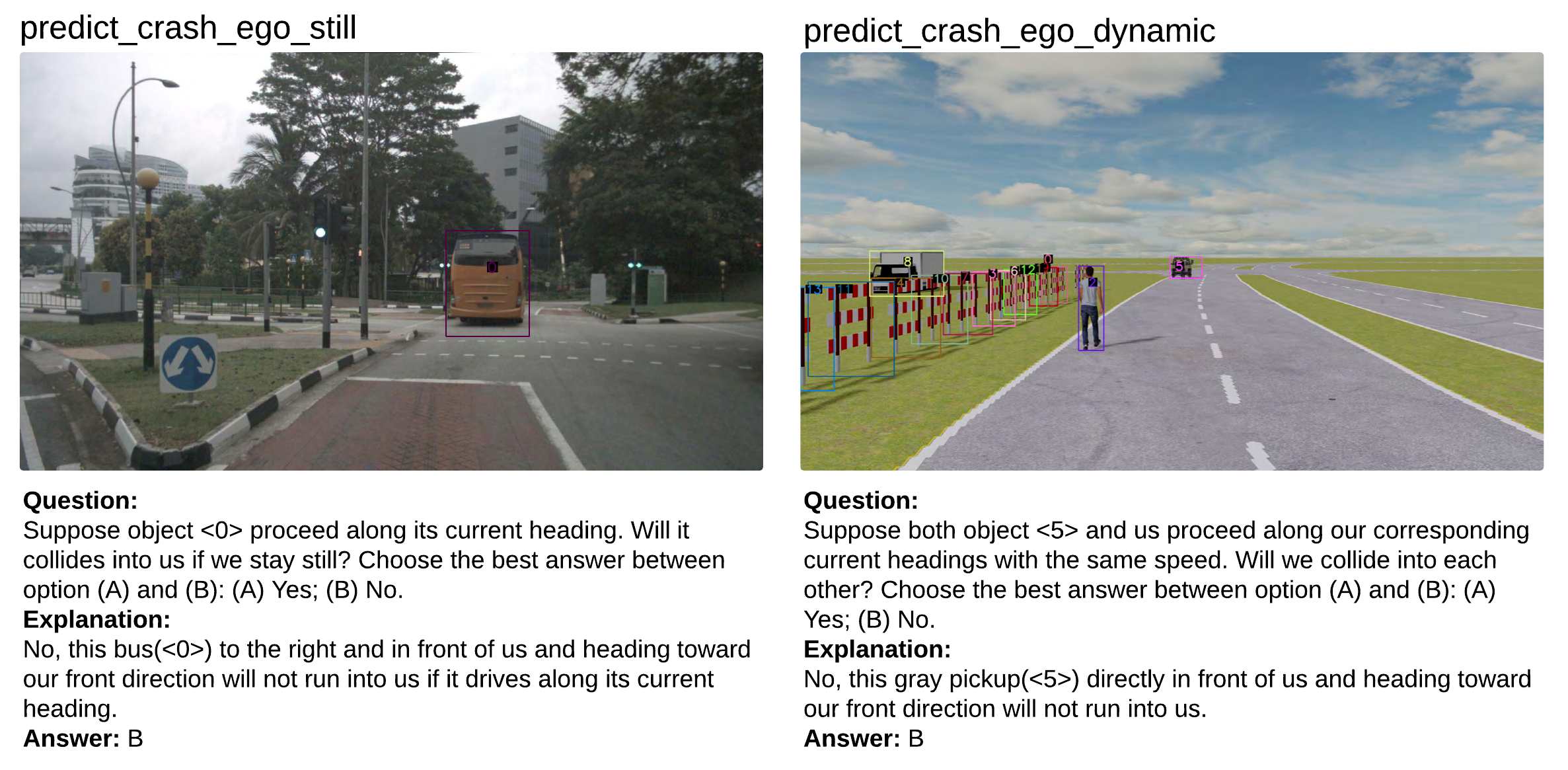}
\end{minipage}
\end{tcolorbox}
}
\clearpage
\newpage
{
\captionsetup{type=table}
\begin{tcolorbox}[colback=white!10,%gray background
                  colframe=black,% black frame color
                  width=\textwidth,
                  arc=1mm, auto outer arc,
                  boxrule=0.5pt,
                 ]

\texttt{\textcolor{blue}{\textbf{Spatial Questions:}}}

\texttt{\textbf{identify\_distance}.} This question prompts VLMs to estimate the distance of the selected object \texttt{<id1>} from the ego.

\begin{minipage}[b]{\textwidth}
            \centering
            \includegraphics[width=\textwidth]{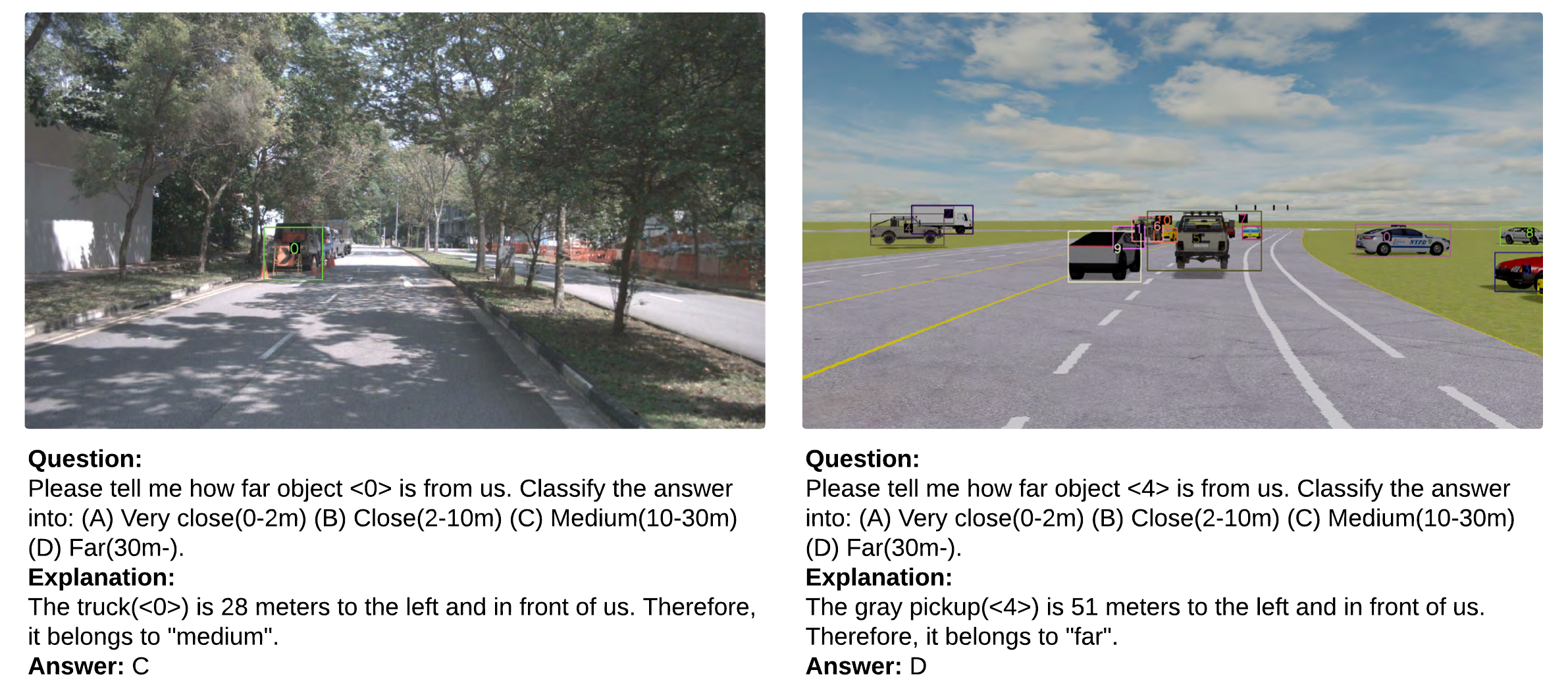}
\end{minipage}
\texttt{\textbf{identify\_position}.} This question prompts VLMs to estimate the direction of the selected object \texttt{<id1>} from the ego.

\begin{minipage}[b]{\textwidth}
            \centering
            \includegraphics[width=\textwidth]{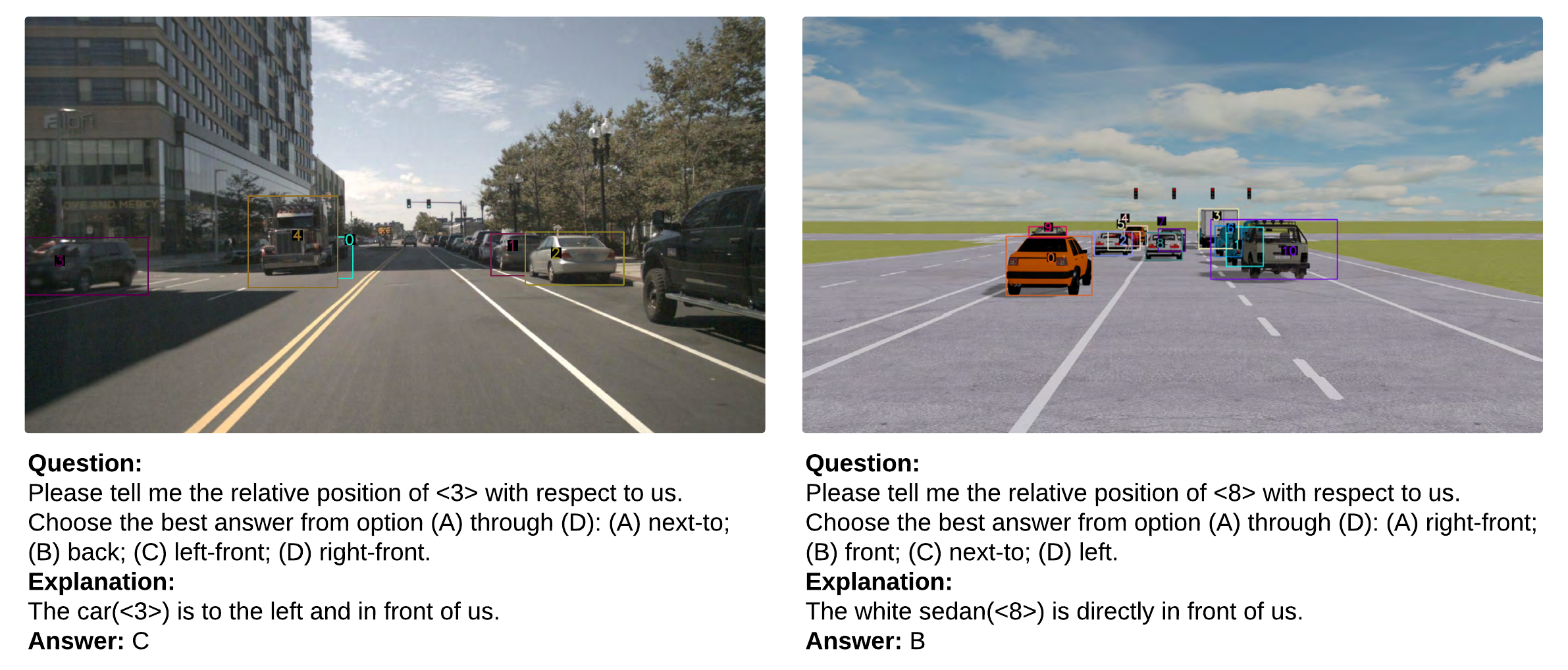}
\end{minipage}
\end{tcolorbox}
}
\clearpage
\newpage
{
\captionsetup{type=table}
\begin{tcolorbox}[colback=white!10,%gray background
                  colframe=black,% black frame color
                  width=\textwidth,
                  arc=1mm, auto outer arc,
                  boxrule=0.5pt,
                 ]

\texttt{\textcolor{blue}{\textbf{Spatial Questions:}}}

\texttt{\textbf{identify\_heading}.} This question prompts models to estimate the heading angle of the selected object \texttt{<id1>}, expressed relative to the ego's front direction. The provided options are sufficiently distinct to avoid ambiguity. 

\begin{minipage}[b]{\textwidth}
            \centering
            \includegraphics[width=\textwidth]{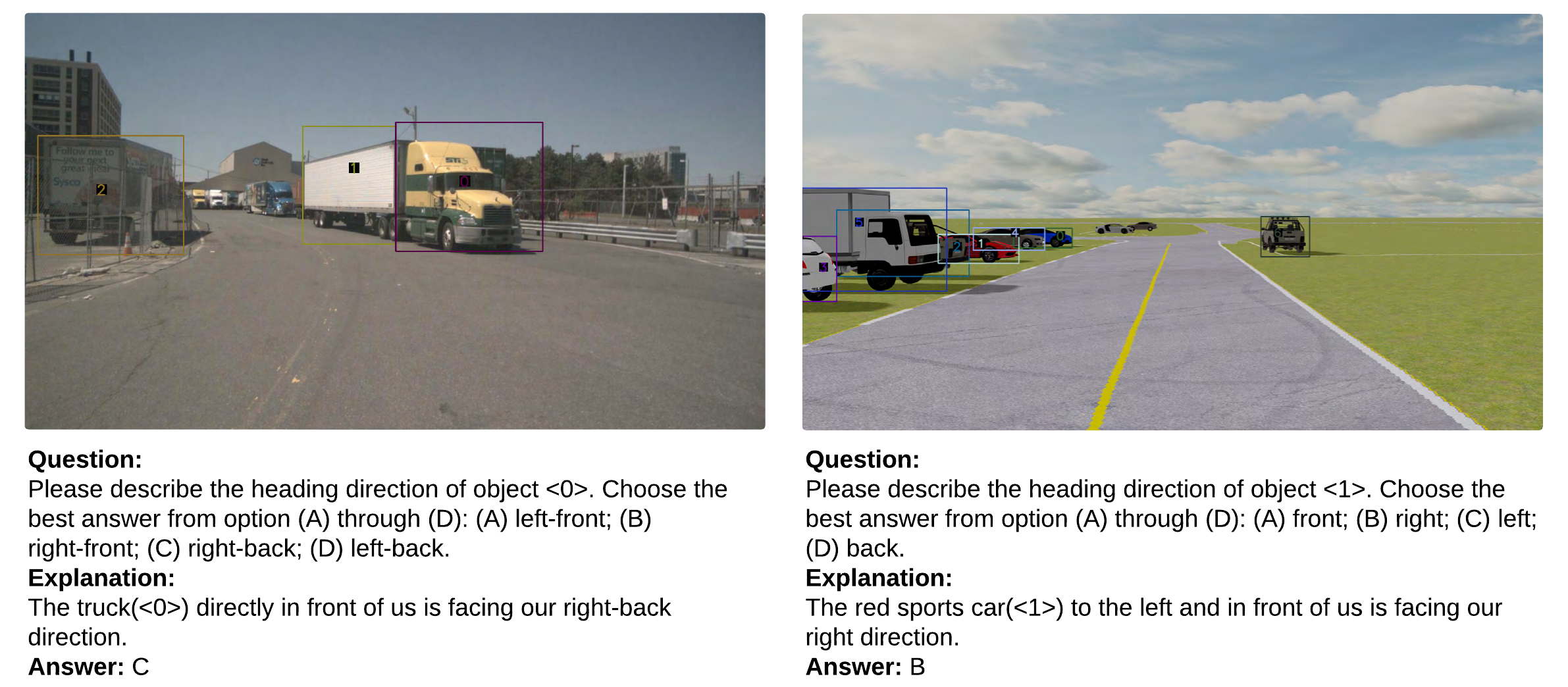}
\end{minipage}
\texttt{\textbf{identify\_color}} This question prompts models to select the color of object \texttt{<id1>}. Note that it is generated only with simulated observations, as ``color" is not annotated in the nuScenes dataset.

\begin{minipage}[b]{\textwidth}
            \centering
            \includegraphics[width=\textwidth]{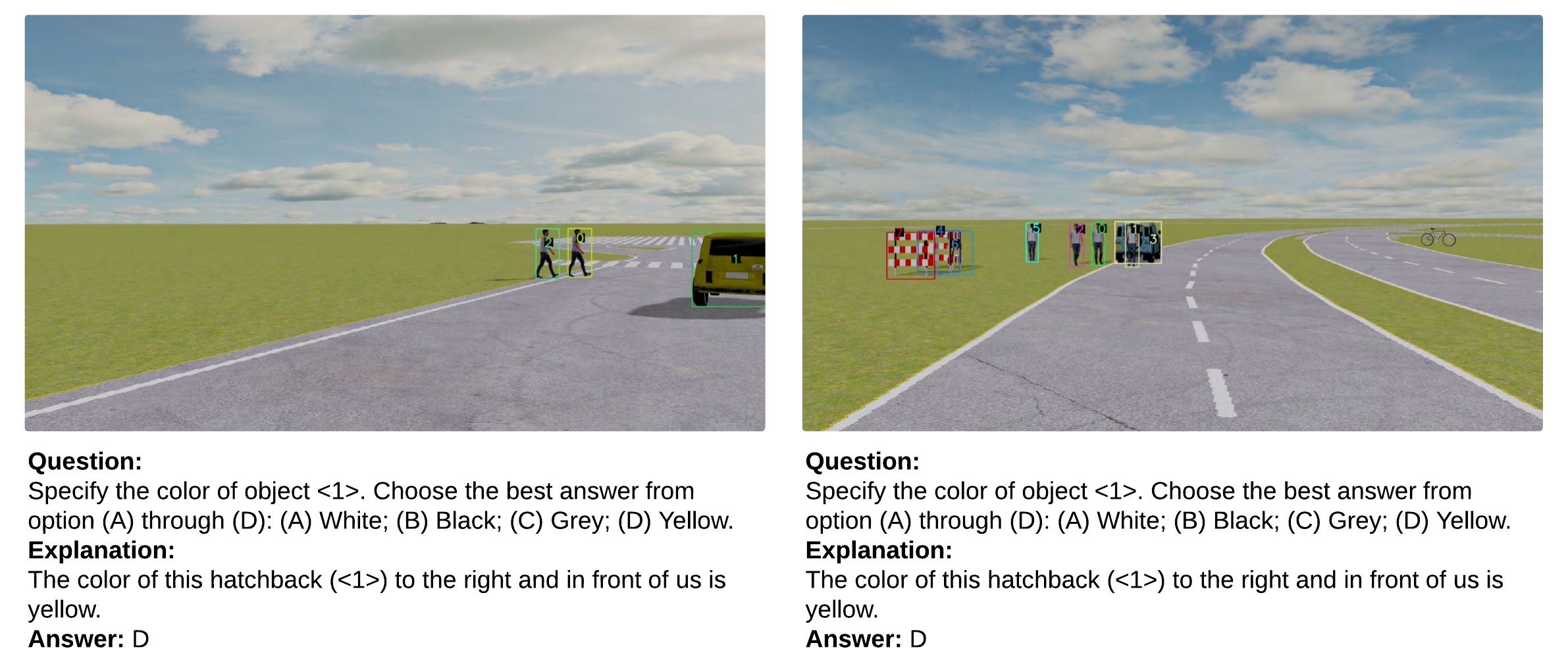}
\end{minipage}
\end{tcolorbox}
}
\clearpage
\newpage
{
\captionsetup{type=table}
\begin{tcolorbox}[colback=white!10,%gray background
                  colframe=black,% black frame color
                  width=\textwidth,
                  arc=1mm, auto outer arc,
                  boxrule=0.5pt,
                 ]

\texttt{\textcolor{blue}{\textbf{Spatial Questions:}}}

\texttt{\textbf{identify\_type}.} This question prompts VLMs to select the most descriptive type of the selected object \texttt{<id1>}.

\begin{minipage}[b]{\textwidth}
            \centering
            \includegraphics[width=\textwidth]{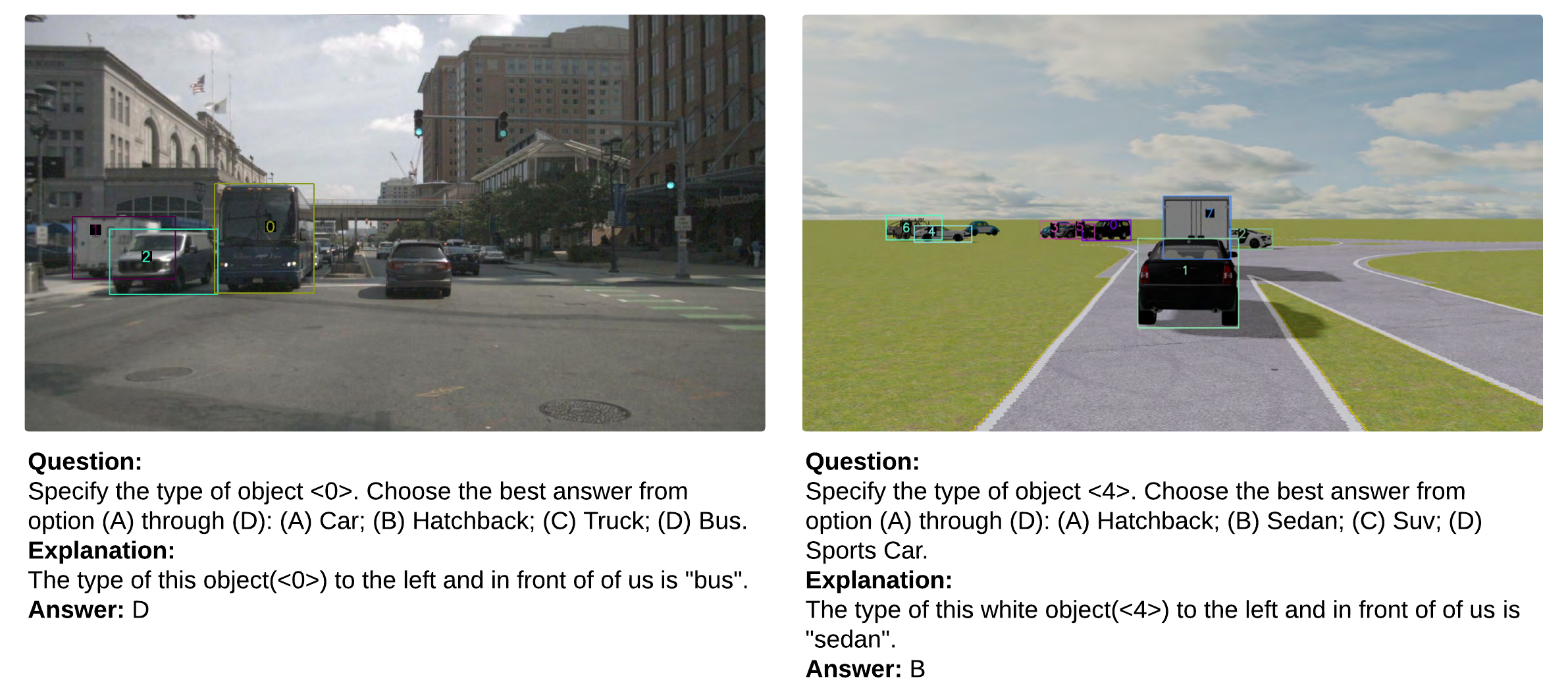}
\end{minipage}
\texttt{\textbf{relative\_distance}.} This question prompts VLMs to select the relative distance between two objects \texttt{<id1>} and \texttt{<id2>}.

\begin{minipage}[b]{\textwidth}
            \centering
            \includegraphics[width=\textwidth]{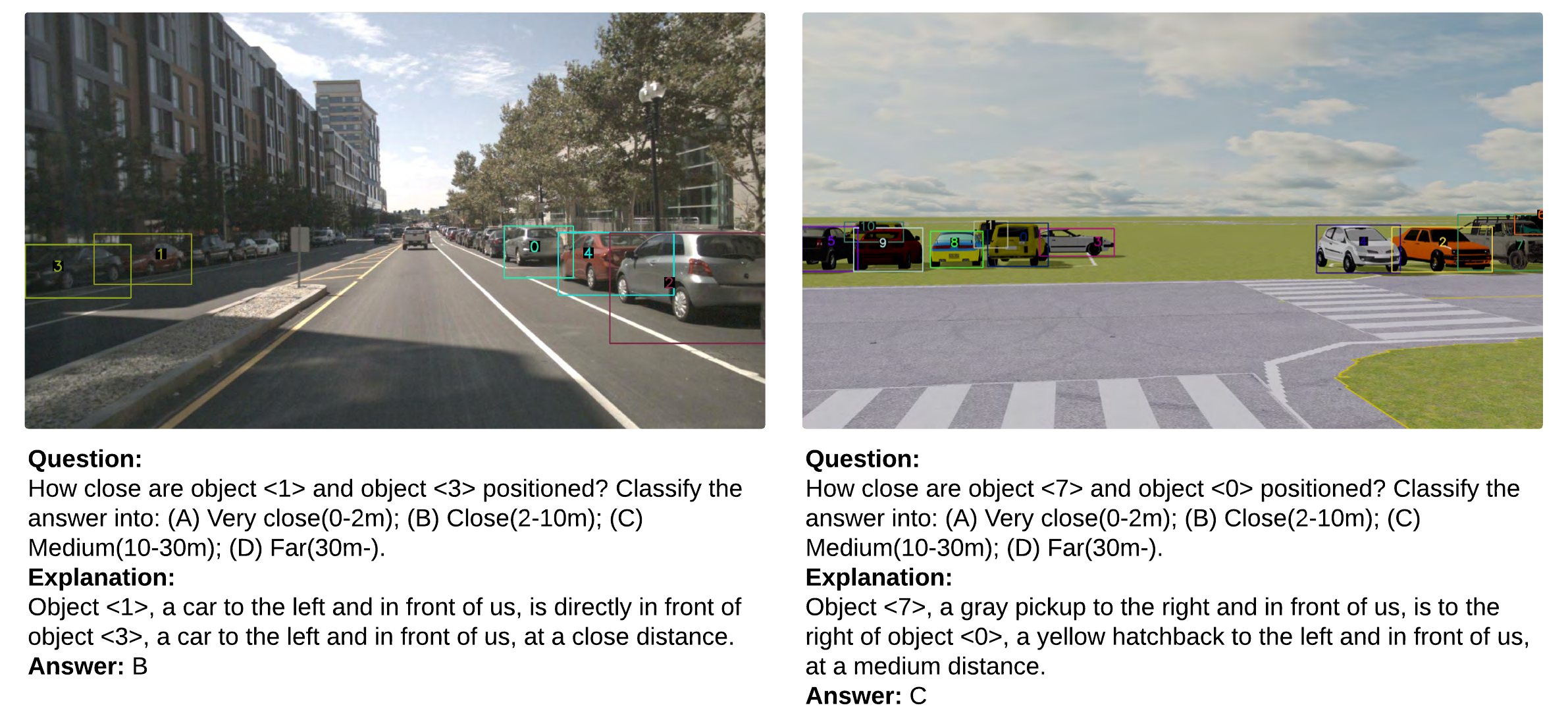}
\end{minipage}
\end{tcolorbox}
}
\clearpage
\newpage
{
\captionsetup{type=table}
\begin{tcolorbox}[colback=white!10,%gray background
                  colframe=black,% black frame color
                  width=\textwidth,
                  arc=1mm, auto outer arc,
                  boxrule=0.5pt,
                 ]

\texttt{\textcolor{blue}{\textbf{Spatial Questions:}}}

\texttt{\textbf{relative\_position}.} This question prompts VLMs to evaluate how is object \texttt{<id1>} related spatially with object \texttt{<id12>}, expressed in the ego perspective.

\begin{minipage}[b]{\textwidth}
            \centering
            \includegraphics[width=\textwidth]{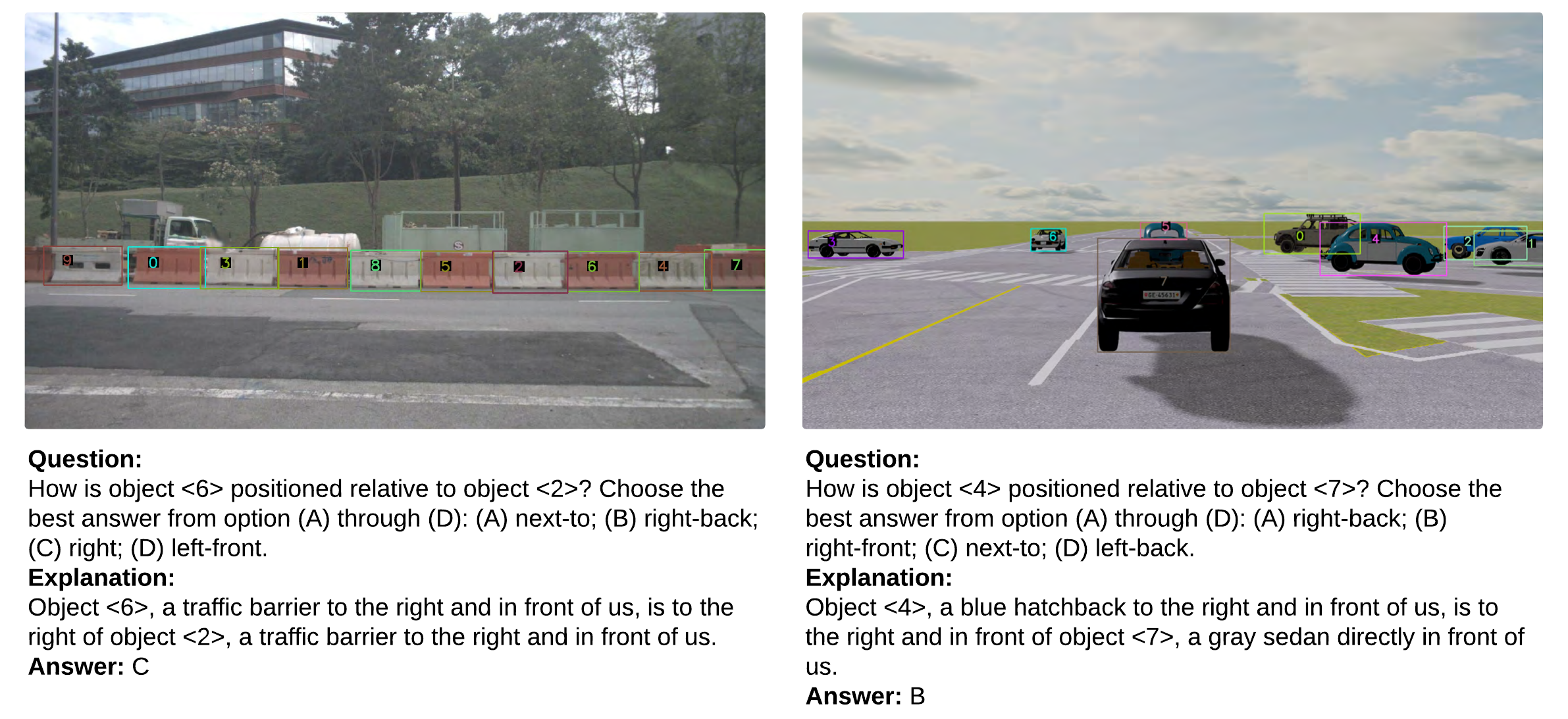}
\end{minipage}
\texttt{\textbf{relative\_heading}.} This question prompts VLMs to determine if object \texttt{<id1>} and \texttt{<id12>} are heading towards roughly the same direction.

\begin{minipage}[b]{\textwidth}
            \centering
            \includegraphics[width=\textwidth]{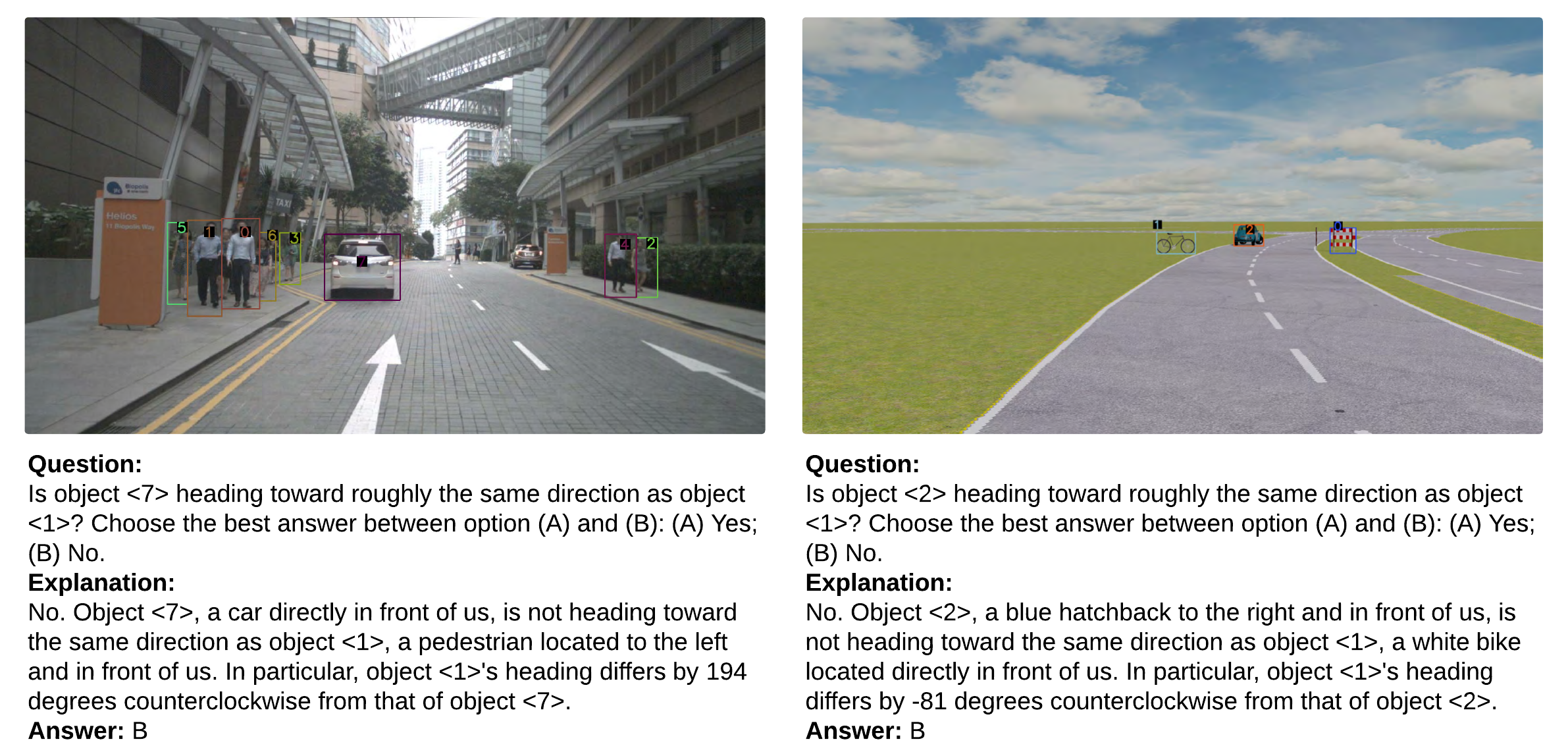}
\end{minipage}
\end{tcolorbox}
}
\clearpage
\newpage
{
\captionsetup{type=table}
\begin{tcolorbox}[colback=white!10,%gray background
                  colframe=black,% black frame color
                  width=\textwidth,
                  arc=1mm, auto outer arc,
                  boxrule=0.5pt,
                 ]

\texttt{\textcolor{blue}{\textbf{Spatial Questions:}}}

\texttt{\textbf{relative\_predict\_crash\_*}.} This family of questions prompts VLMs to infer whether two objects \texttt{<id1>} and \texttt{<id12>} will collect under varying assumptions.

\begin{minipage}[b]{\textwidth}
            \centering
            \includegraphics[width=\textwidth]{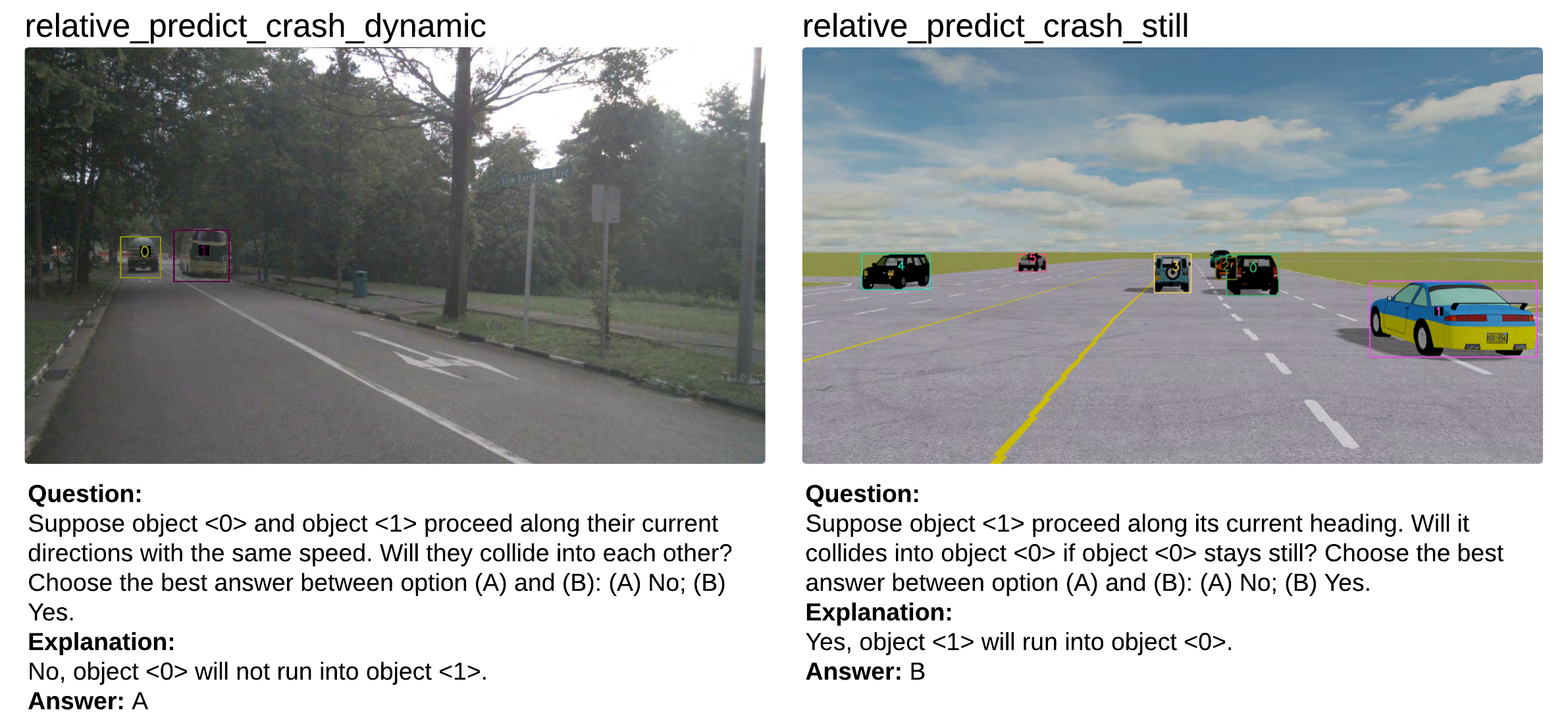}
\end{minipage}
\texttt{\textbf{pick\_closer}.} This question asks the VLM to select the closer object from two candidates.

\begin{minipage}[b]{\textwidth}
            \centering
            \includegraphics[width=\textwidth]{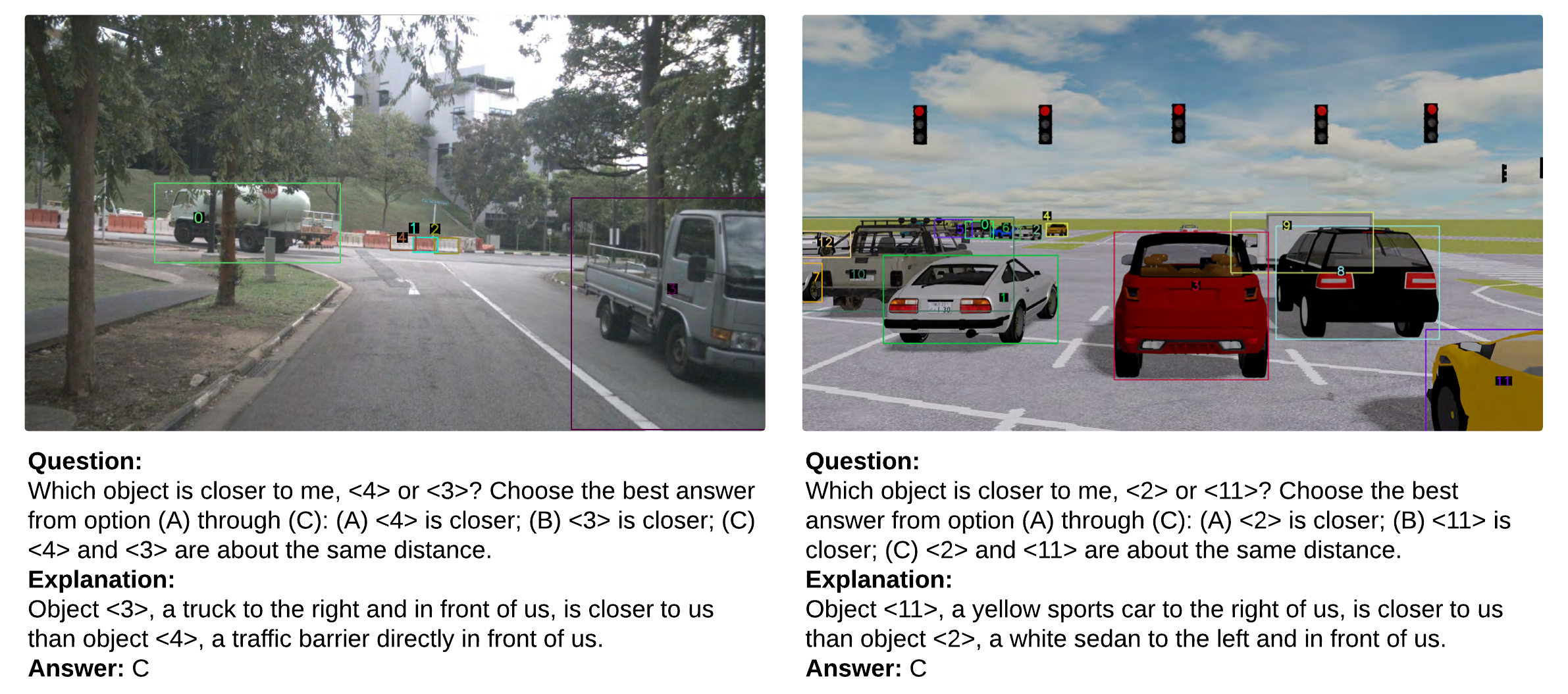}
\end{minipage}
\end{tcolorbox}
}
\clearpage
\newpage
{
\captionsetup{type=table}
\begin{tcolorbox}[colback=white!10,%gray background
                  colframe=black,% black frame color
                  width=\textwidth,
                  arc=1mm, auto outer arc,
                  boxrule=0.5pt,
                 ]

\texttt{\textcolor{blue}{\textbf{Spatial Questions:}}}

\texttt{\textbf{order\_*st}.} This family of questions asks the VLM to attend to multiple objects and sort their relevance by some spatial ordering in top-down world coordinates.

\begin{minipage}[b]{\textwidth}
            \centering
            \includegraphics[width=\textwidth]{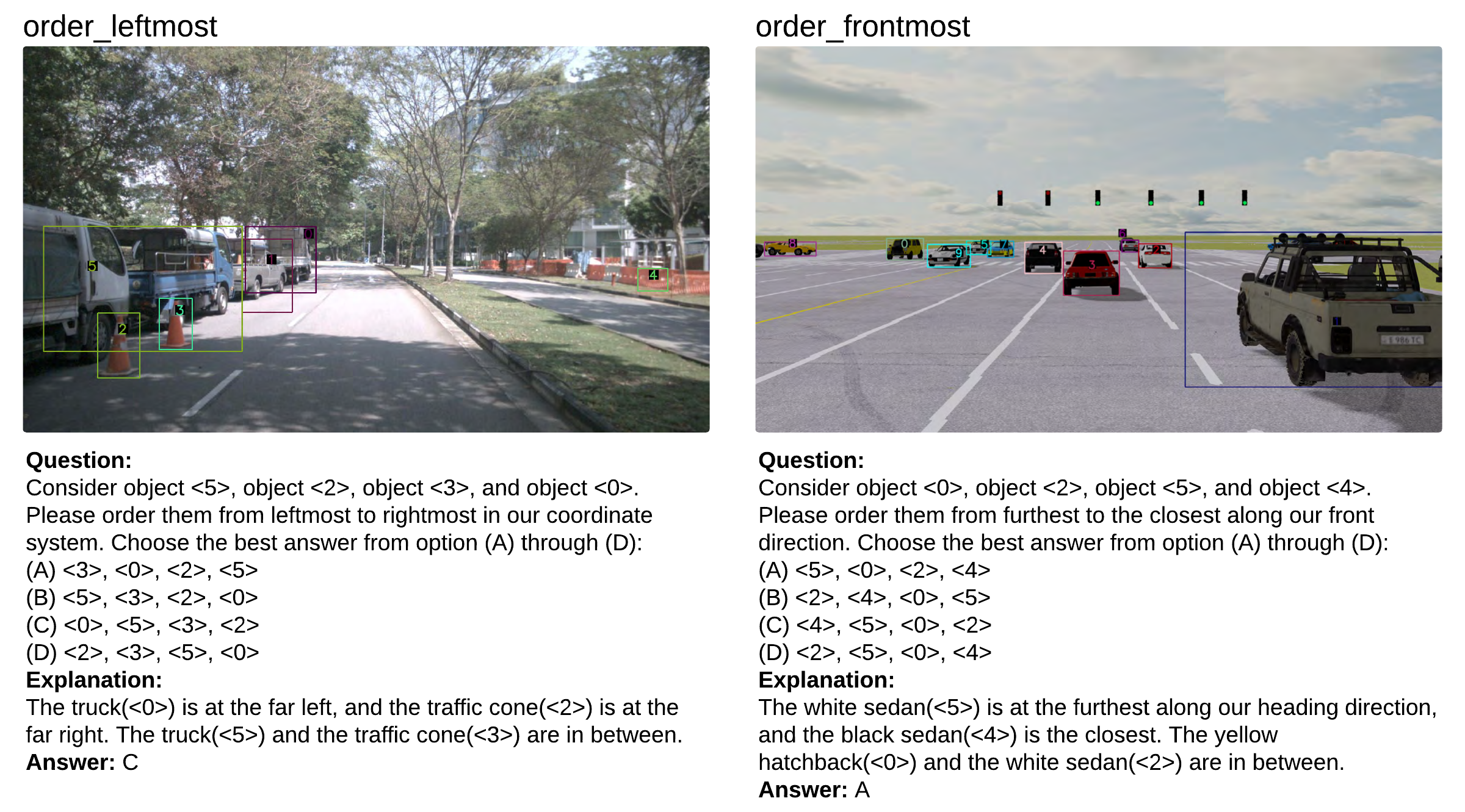}
\end{minipage}
\texttt{\textbf{describe\_sector}.} This question asks the VLM to attend to all observable objects and select the maximal object set such that all of its members are in the specified ego's direction from the question body.

\begin{minipage}[b]{\textwidth}
            \centering
            \includegraphics[width=\textwidth]{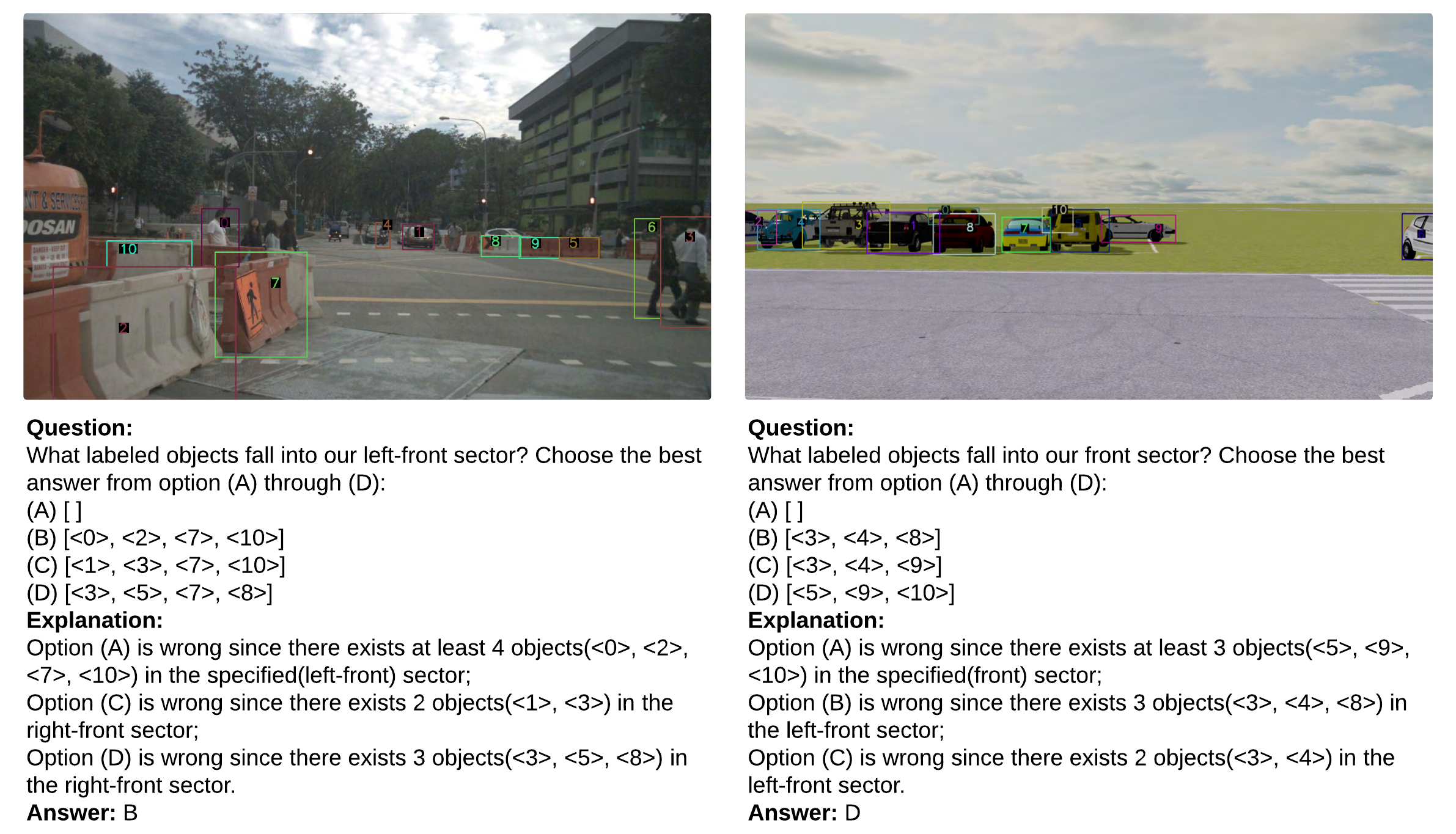}
            \vspace{-10pt}
\end{minipage}
\end{tcolorbox}
}
\clearpage
\newpage
{
\captionsetup{type=table}
\begin{tcolorbox}[colback=white!10,%gray background
                  colframe=black,% black frame color
                  width=\textwidth,
                  arc=1mm, auto outer arc,
                  boxrule=0.5pt,
                 ]

\texttt{\textcolor{blue}{\textbf{Spatial Questions:}}}

\texttt{\textbf{describe\_distance}.} This question asks VLMs to attend to all observable objects and select the maximal object set such that all of its members are located away from the ego by the specified distance from the question body.

\begin{minipage}[b]{\textwidth}
            \centering
            \includegraphics[width=\textwidth]{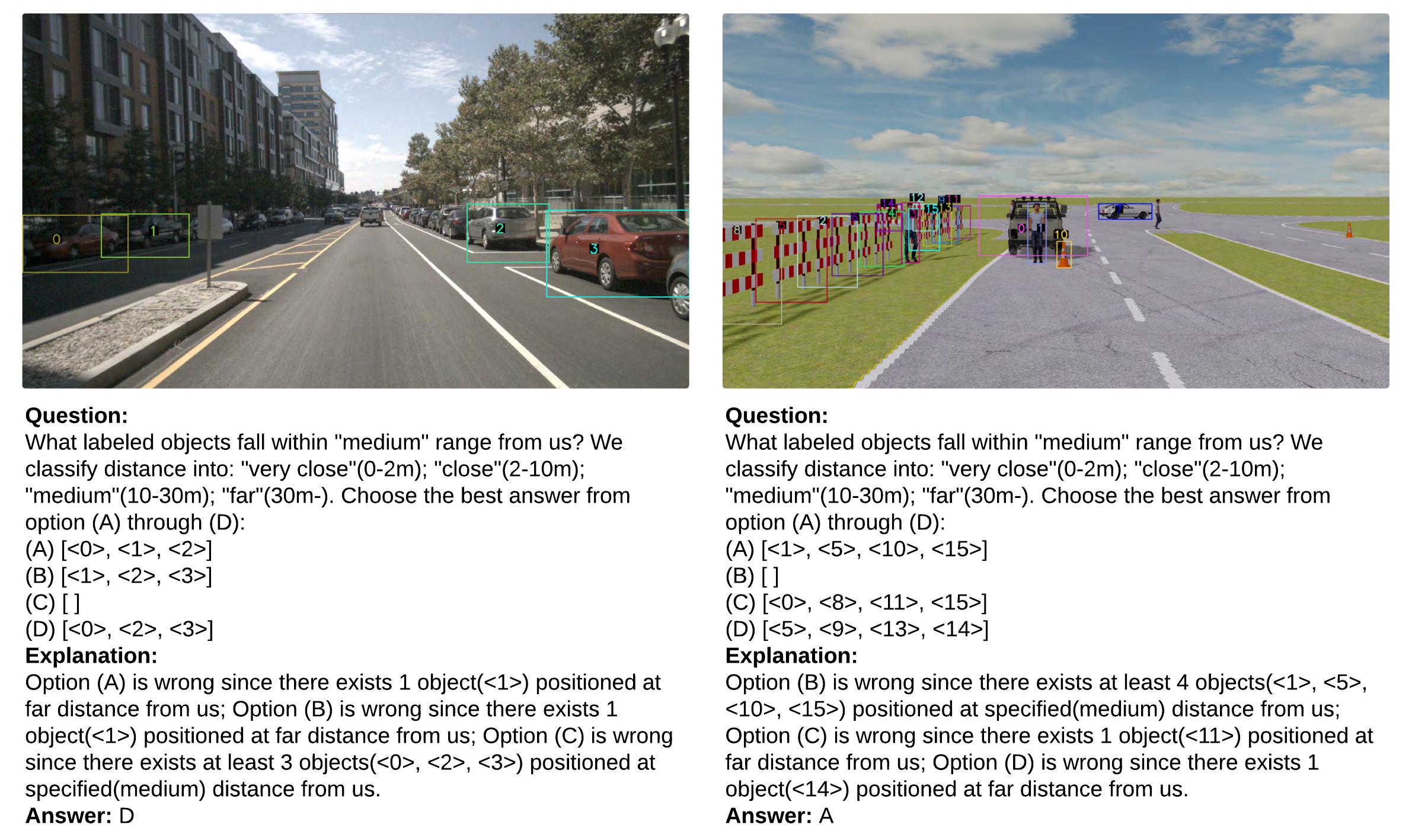}
\end{minipage}
\texttt{\textbf{identify\_*st}.} This question asks the VLM to attend to all observable objects and select the leading object according to some ordering specified in the question body.

\begin{minipage}[b]{\textwidth}
            \centering
            \includegraphics[width=\textwidth]{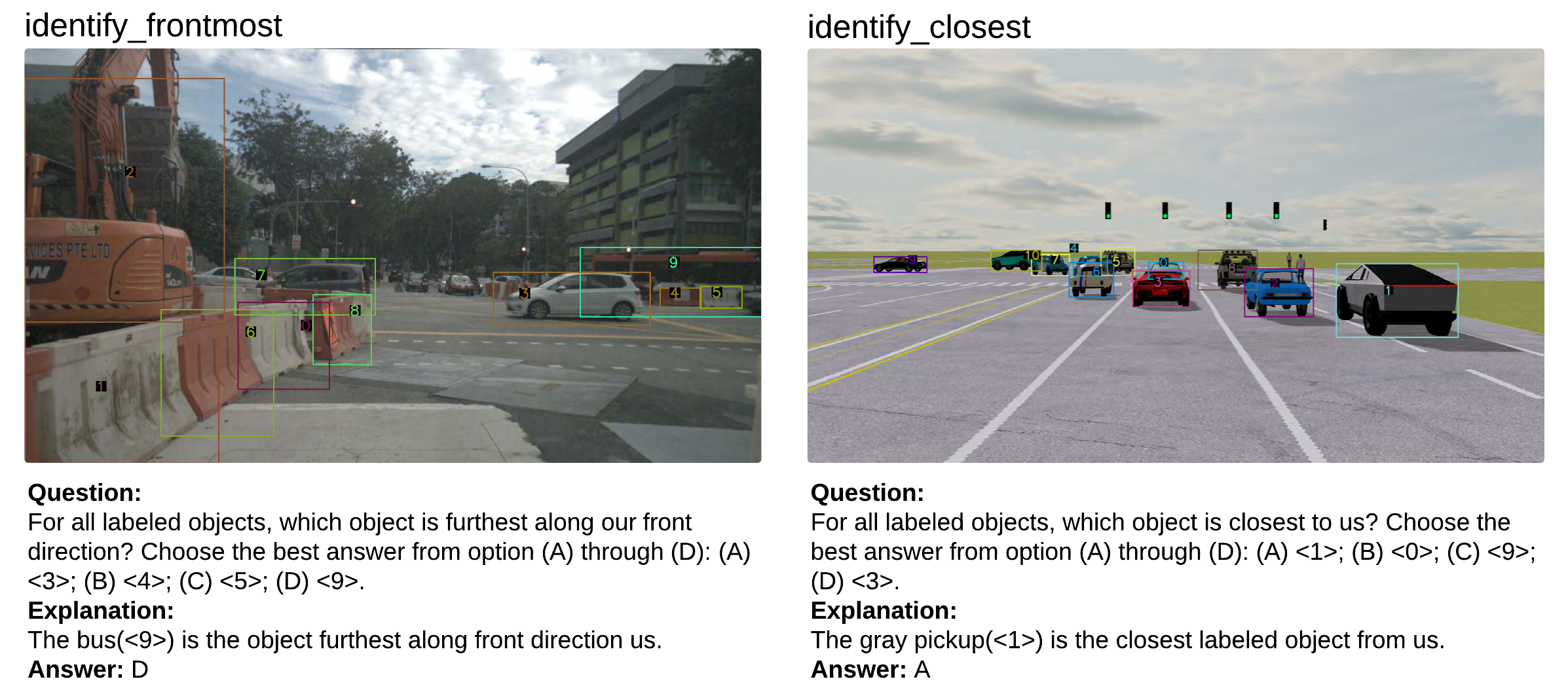}
            \vspace{-10pt}
\end{minipage}
\end{tcolorbox}
}
\clearpage
\newpage
{
\captionsetup{type=table}
\begin{tcolorbox}[colback=white!10,%gray background
                  colframe=black,% black frame color
                  width=\textwidth,
                  arc=1mm, auto outer arc,
                  boxrule=0.5pt,
                 ]

\texttt{\textcolor{blue}{\textbf{Spatial Questions:}}}

\texttt{\textbf{describe\_scenario}.} This question prompts the VLM to examine all labeled objects in the scenario. It is a train-only question designed to boost learning performance and avoid VLM collapse.

\begin{minipage}[b]{\textwidth}
            \centering
            \includegraphics[width=\textwidth]{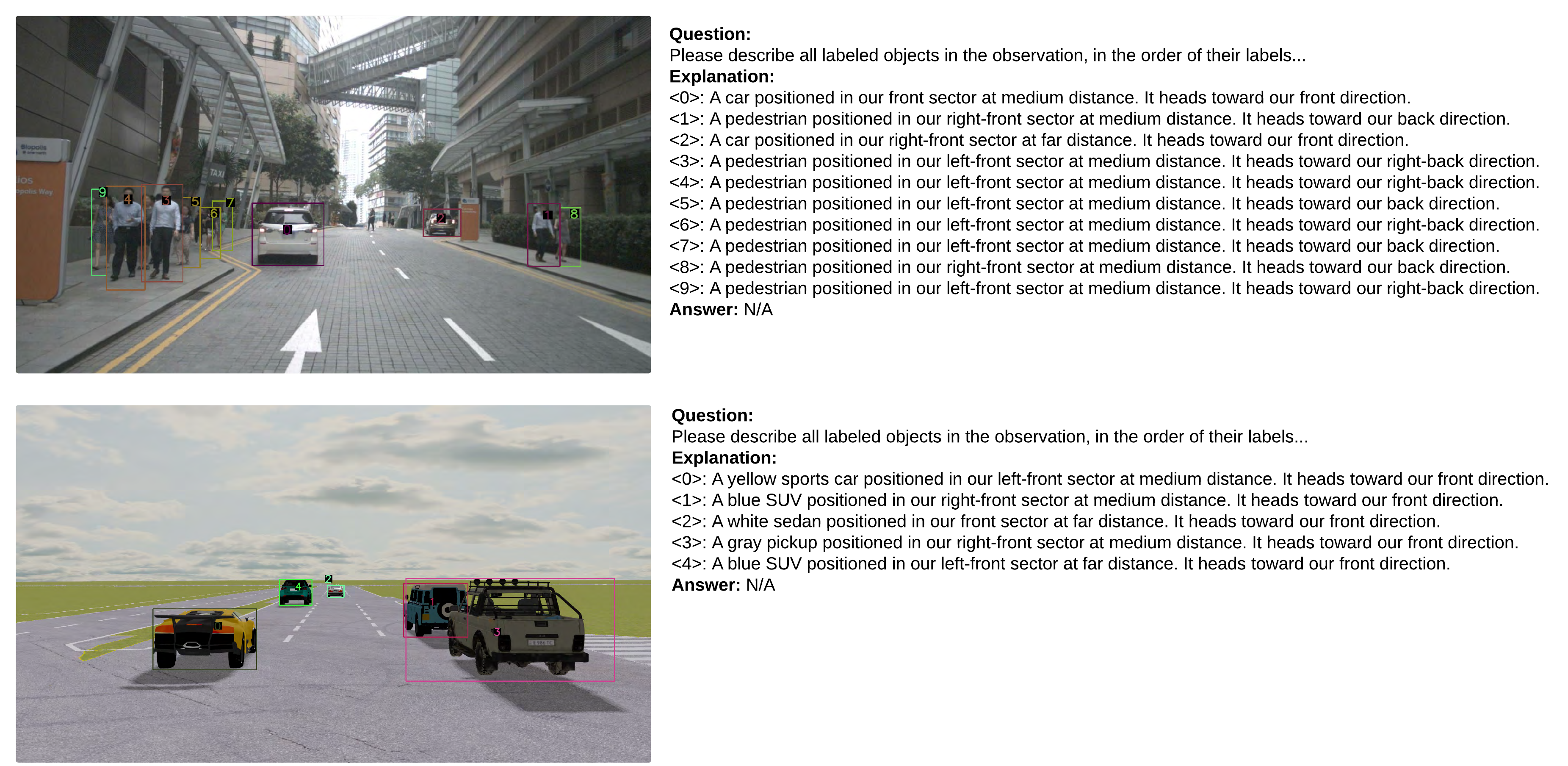}
\end{minipage}
\texttt{\textcolor{blue}{\textbf{Grounding Questions:}}}

\texttt{\textbf{grounding}.} This question examines the visual grounding ability of the tested VLM. All non-answer options are selected from valid labels to challenge to model maximally.

\begin{minipage}[b]{\textwidth}
            \centering
            \includegraphics[width=\textwidth]{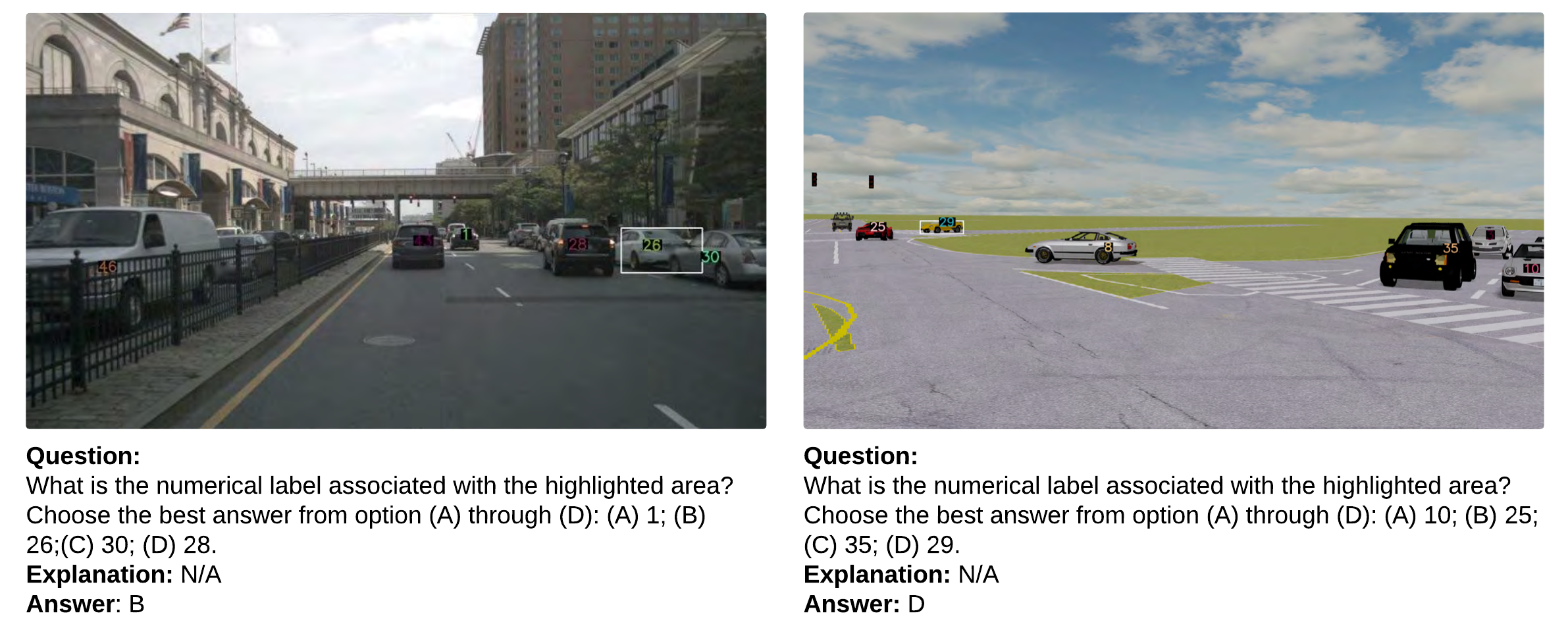}
\end{minipage}
\end{tcolorbox}
}

\clearpage
\newpage
%\begin{figure*}[!t]
%    \centering
%    \vspace{-25pt}
%    \includegraphics[width=\textwidth]{examples/learned.pdf}
%    \vspace{-25pt}
%    \caption{\textbf{Successfully-answered test questions after fine-tuning}. Answered by InternVL-8B v.s. InternVL-8B fine-tuned on the withheld training dataset. The numbers in parenthesis are the question IDs in the test set.}
%    \label{fig:learned}
%\end{figure*}
%\clearpage
%\newpage

%\begin{figure*}[!t]
%    \centering
%    \includegraphics[width=\textwidth]{examples/learned.pdf}
%    \caption{\textbf{Successfully-answered test questions after fine-tuning}. Answered by InternVL-8B v.s. InternVL-8B fine-tuned on the withheld training dataset.}
%    \label{fig:learned}
%\end{figure*}